\documentclass[orms,nonblindrev]{informs3_no_remark}

\usepackage{amsmath,amsfonts,amssymb}
\usepackage[ruled,lined,linesnumbered]{algorithm2e}
\usepackage{hyperref}

\usepackage{natbib}
 \bibpunct[, ]{(}{)}{,}{a}{}{,}%
 \def\bibfont{\small}%

\usepackage{rotating}
\usepackage{fancyvrb}

\def\Prob{{\mathbb P}}
\def\E{{\mathbb E}}
\def\R{{\mathbb R}}

\def\ghat{\hat{g}}

\def\bfP{\mathcal{P}}

\usepackage{appendix}

\TheoremsNumberedThrough     % Preferred (Theorem 1, Lemma 1, Theorem 2)
\ECRepeatTheorems
\EquationsNumberedThrough    % Default: (1), (2), ...
\MANUSCRIPTNO{IJDS-0001-1922.65}

\begin{document}
%\setcounter{page}{1}
%\DOI{10.1287/xxxx.0000.0000}%


\RUNTITLE{Stochastic Gradient Descent with Adaptive Data}

\TITLE{Stochastic Gradient Descent with Adaptive Data}

\ARTICLEAUTHORS{%
\AUTHOR{Ethan Che}
%,\textsuperscript{a} Second Author,\textsuperscript{b} Third Author,\textsuperscript{c} Fourth Author,\textsuperscript{c}

\AFF{Columbia Business School, \EMAIL{eche25@gsb.columbia.edu}}
%\textsuperscript{b}School of Industrial Engineering, Good College, Collegeville, Maine 01234 \EMAIL{secauth@goodcoll.edu}; 
%\textsuperscript{c}Their Common Affiliation \EMAIL{thauth@anywhere.edu, fourauth@anywhere.edu}

%mirko.janc@informs.org
\AUTHOR{Jing Dong}

\AFF{Columbia Business School, \EMAIL{jing.dong@gsb.columbia.edu}}

\AUTHOR{Xin T. Tong}

\AFF{National University of Singapore, \EMAIL{xin.t.tong@nus.edu.sg}}
}

\ABSTRACT{
Stochastic gradient descent (SGD) is a powerful optimization technique that is particularly useful in online learning scenarios. Its convergence analysis is relatively well understood under the assumption that the data samples are independent and identically distributed (iid). However, applying SGD to policy optimization problems in operations research involves a distinct challenge: the policy changes the environment and thereby affects the data used to update the policy. The adaptively generated data stream involves samples that are non-stationary, no longer independent from each other, and affected by previous decisions. The influence of previous decisions on the data generated introduces bias in the gradient estimate, which presents a potential source of instability for online learning not present in the iid case. In this paper, we introduce simple criteria for the adaptively generated data stream to guarantee the convergence of SGD. We show that the convergence speed of SGD with adaptive data is largely similar to the classical iid setting, as long as the mixing time of the policy-induced dynamics is factored in. Our Lyapunov-function analysis allows one to translate existing stability analysis of stochastic systems studied in operations research into convergence rates for SGD, and we demonstrate this for queueing and inventory management problems. We also showcase how our result can be applied to study the sample complexity of an actor-critic policy gradient algorithm.
}

\maketitle

\section{Introduction}
We consider the following stochastic optimization problem
\begin{equation}\label{eq:main}
\min_{\theta \in \Theta} \ell(\theta)=\mathbb{E}_{\mu_\theta}[\mathcal{L}(\theta, z)].
\end{equation}
In \eqref{eq:main}, $\theta$ is a $m$-dimensional policy parameter that parametrizes a Markov chain with transition kernel $P_\theta$. The probability measure $\mu_\theta$ is the unique invariant distribution of $P_\theta$, and $z\sim\mu_{\theta}$ is a random outcome. Here, the set $\Theta$ could either be a convex constraint set or $\mathbb{R}^m$ in the unconstrained setting. 
%(Later we do not need $\Theta$ to be compact? Also, it is $C$ or $\Theta$?)

% where $\theta$ is a $d$-dimensional parameter, $\Theta$ could either be a compact and convex constraint set or $\mathbb{R}^d$ (i.e., the unconstrained setting, $z\sim\mu_{\theta}$, and $\mu_{\theta}$ is the invariance distribution of a Markov chain with transition kernel $P_{\theta}$, i.e., the transition kernel of the Markov chain depends on the parameter $\theta$.
For example, when designing service systems, we are interested in finding the optimal pricing and/or capacity sizing policies to strike a balance between revenue and service quality \citep{kim2018value, chen2023online}. 
Here the policy parameter $\theta$ may consist of the price, which affects the demand, and the capacity, which affects the service speed. Jointly, they control a Markov chain $P_\theta$ that describes the queueing dynamics, and the service quality is often measured by the steady-state average waiting time.
Similarly, in inventory management, we are interested in finding the optimal inventory ordering policy (e.g., base-stock level) to minimize the long-run average holding and backlog costs \citep{huh2009adaptive,zhang2020closing}. Here, the replenishment policy, indexed by $\theta$, controls the dynamics of the inventory level, whose stationary distribution in turn determines the long-run average costs.
The stochastic optimization problem \eqref{eq:main} also arises in machine-learning applications, such as policy gradient-based reinforcement learning \citep{sutton2018reinforcement, agarwal2021theory}, strategic classification with adaptive best response \citep{mendler2020stochastic,li2022state}, and adaptive experimental design with temporal carryovers \citep{glynn2020adaptive, hu2022switchback}.

In many problems of practical interests, direct access to the distribution $\mu_\theta$ may not be available. Instead, at each time point $t$, we can apply a new candidate policy $\theta_t$ on the concurrent system state $z_{t-1}$ and obtain a new data point following 
\begin{equation}
\label{eqn:adaptivedata}
z_t\sim P_{\theta_t}(\,\cdot\,|z_{t-1}).    
\end{equation}
We will refer to this as the \emph{adaptive} data setting. Instead of minimizing~\eqref{eq:main} using the true gradient $\nabla \ell(\theta)$, which relies on the unknown distribution $\mu_{\theta}$, one only has access to a gradient estimator $g(\theta_{t}, z_{t})$ based on the adaptive data stream, which satisfies
\begin{equation}
\label{eq:adaptive-gradient}
\E_{z\sim \mu_{\theta}}[g(\theta,z)] = \nabla \ell(\theta).
\end{equation}
This resembles the problem setup for reinforcement learning (RL)~\citep{sutton2018reinforcement}, and includes policy-gradient algorithms as a special case. However, the RL literature generally focuses on a specific class of gradient estimators derived from the $\mathsf{REINFORCE}$ estimator~\citep{williams1992simple} or the $Q$-function. In contrast, the adaptive data setting we consider is applicable for any gradient estimator satisfying~\eqref{eq:adaptive-gradient} and thereby covers a much larger range of gradient estimation strategies~\citep{mohamed2020monte}, including infinitesimal perturbation analysis (IPA)~\citep{heidelberger1988convergence, glasserman1992stationary} and general likelihood-ratio gradient estimation~\citep{glynn1990likelihood}. These gradient estimation strategies are outside the scope of the existing RL convergence analysis, but are of particular relevance to operations research applications, such as queuing and inventory management (see, e.g., \cite{chen2023online,huh2009adaptive}).

The core challenge of applying stochastic gradient descent (SGD) to the adaptive data setting is that $z_{t}$ is not only determined by the current action $\theta_{t}$ but also depends on previous actions through $z_{t-1}$. As a result, using $z_{t}$ to form a stochastic gradient estimator $g(\theta_{t},z_{t})$ leads to \emph{biased} estimation of the true gradient $\nabla \ell (\theta_{t})$, as  $\E[g(\theta_{t},z_{t})|z_{t-1}] \neq \nabla \ell(\theta)$ in general. If the effects of previous actions persist in the system for a long time, the bias can be significant.
SGD with biased gradients may not be able to converge to the desired minimum (or even a small enough neighborhood of the minimum).
%even after infinitely many steps.

Consider, for example, the optimal pricing problem in a GI/GI/1 queue, where the arrival rate is determined by the price charged according to a demand function $\lambda(p)$. Let $T_{t+1}$ denote the baseline interarrival time between the $t$-th and $(t+1)$-th arrivals, $S_t$ denote the service time, and $W_{t}$ denote the waiting time of customer $t$, i.e., the $t$-th arrival. For a fixed price $p$, by Lindley's recursion, $\{W_t: t\geq 0\}$ is a Markov chain satisfying
\[
W_{t+1}=\left(W_t + S_t - \frac{T_{t+1}}{\lambda(p)}\right)^+.
\]
Our goal is to choose the optimal price to maximize the revenue minus the long-run average cost of waiting, or equivalently,  
\[
\min_p ~ \ell(p)=-\left(p\lambda(p)-h\lambda(p)\E_{\pi_p}[W]\right),
\]
where $\pi_p$ denotes the steady-state distribution of the Markov chain $\{W_t: t\geq 0\}$ with arrival rate $\lambda(p)$ and $h$ denote the holding/waiting cost per unit time per customer. If we are to update the price after each arrival and let $p_t$ denote the price charged for the $t$-th arrival, then the waiting times satisfy
\[
\tilde W_{t+1}=\left(\tilde W_t + S_t - \frac{T_{t+1}}{\lambda(p_{t+1})}\right)^+,
\]
which is a non-stationary Markov process due to varying $p_t$'s.

%The challenge lies in the fact that 
Note that the waiting time $\tilde W_t$ is not only affected by $p_t$, but also the previous prices charged, i.e. $p_s$ for $s<t$, as they affect the current congestion level in the queue. As a result, $\E[\tilde W_{t}] \neq \E_{\pi_{p_{t}}}[W_t]$ due to transient behavior of the Markov chain. In this case, using the current waiting time to estimate the steady-state waiting time leads to a biased estimation of $\nabla \ell(p_t)$, which may derail the convergence of the standard SGD updates to the optimal $p^{*}$. For example, if one sets very low prices $p_{s} \ll p_{t}$ for $s < t$, the waiting time $\tilde{W_{t}}$ could be much larger than $\E_{\pi_{p_{t}}}[W_t]$ due to the congestion induced by high demands in previous periods (as result of the low prices $p_{s}$'s). Since $\tilde{W}_{t}$ will be an over-estimate of the stationary waiting time $\E_{\pi_{p_{t}}}[W_t]$, using this estimate to update the price could cause the new price $p_{t+1}$ to be too high. 
%out of concern for congestion, 
%even if $p_{t}$ were the optimal price in steady-state. 
The oscillations due to delayed feedback can lead to instability as demonstrated in control theory.
%could prevent the SGD updates from converging to the optimal price.
%Our question is 
%In this paper, we study the stochastic gradient descent-based update of $\theta_t$ using the adaptive data sequence $z_{1},\ldots,z_t$ and its convergence and finite-time performance. 

Optimization algorithms that use data streams to sequentially update the solution are often referred to as \emph{online} algorithms. 
%Among them, stochastic gradient descent (SGD)\citep{robbins1951stochastic}, is the most popular one due to its simplicity. 
The convergence behavior of online algorithms
%Online algorithms for solving \eqref{eq:main}, such as stochastic gradient descent, under an 
using independent and identically distributed (iid) data stream has been well studied in the literature \citep{robbins1951stochastic,shapiro2021lectures, moulines2011non}.
More recent results have shown that the iid requirement can be relaxed.
\cite{duchi2012ergodic,agarwal2012generalization, sun2018markov} study the performance of online learning algorithms on a dependent data stream. In particular, they assume $z_k$'s are generated from a fixed suitably ergodic Markov chain. 
But these results cannot be applied to our problem directly, since they would require that the invariant distribution does not depend on the policy $\theta$. In contrast, in the service system design example discussed above, the price changes the transition dynamics of the queue, which changes the steady-state waiting time distribution.
%The problems discussed above clearly do not fit these requirements.

 % One indirect way to apply the existing results with iid data to adaptive data is to only update $\theta$ periodically, where the period (or batch size) $b_k\in \mathbb{Z}^+$, is set to be long enough such that the data distribution is close to stationarity towards the end of the period. In other words, we would fix $\theta_t\equiv\zeta_k$ for $\sum_{j\leq k-1}{b_j}<t\leq \sum_{j\leq k}{b_j}$, so that an accurate estimation of $\nabla_{\theta}\E_{\mu_{\zeta_k}}[l(\zeta_k,z)]$ is available using the batch updates
 
One indirect way to apply the existing results with iid data to the adaptive data setting is to only update $\theta$ periodically, where the period (or batch size) $B\in \mathbb{N}^+$ is set to be long enough such that the data distribution is close to stationarity towards the end of the period. In other words, for all $t \in \{0,...,B - 1\}$ we would maintain the same policy  $\theta_{0} = ... = \theta_{B -1}$ in order for $z_{B-1}$ to resemble a draw from the stationary distribution $\mu_{\theta_{0}}$. This would allow for a good estimate of  $\nabla_{\theta}\E_{\mu_{\theta_{0}}}[\mathcal{L}(\theta_{0},z)]$, which would then be used to update the policy at time $B$. One may even specify a schedule of batch sizes $B_{k}$ to reliably control the non-stationarity induced by the actions (see, e.g., \cite{chen2023online,huh2009adaptive, hu2022switchback}). However, reducing the frequency at which the policy is updated curtails the adaptivity of the algorithm, exposing a potential trade-off between adaptivity and bias from non-stationarity. Balancing this tradeoff would require carefully choosing the batch size $B$. This may require detailed knowledge about the ergodicity property, e.g., the exact rate of convergence, of the underlying Markov chain which is unavailable or hard to get in many applications. Overall, it is a priori unclear how the length of the period/batch size affects the performance of the learning algorithm in the adaptive setting.

The papers \cite{mendler2020stochastic} and \cite{drusvyatskiy2023stochastic} study stochastic optimization when one can draw independent samples from the updated data distribution $\mu_{\theta}$. This is similar to the periodic updating design discussed above, because, in most practical settings, one can only generate independent samples from $\mu_{\theta}$ by applying the same policy for a long enough time, i.e., running the Markov chain under $P_{\theta}$ until it reaches stationarity.

In this paper, we study the convergence of SGD where only one sample or a minibatch of samples, i.e., $B_{k}$'s being a fixed $O(1)$ constant, 
is used at each iteration.
Under certain ergodicity and continuity conditions on the Markov transition kernels, we show that SGD with adaptive data can achieve $O((\log T)^2/\sqrt{T})$ convergence to a stationary point in the nonconvex case. In the convex case, for projected SGD where the projection set is convex (can be $\mathbb{R}^d$, which corresponds to the case without projection), we show that it can achieve $O((\log T)^4/\sqrt{T})$ convergence to the optimal when $l$ is convex, and $O((\log T)^2/T)$ convergence when $l$ is strongly convex. These rates are similar to the iid case \citep{shapiro2021lectures}. We also show how the mixing time of the underlying Markov chains is incorporated into the convergence rate analysis.  It is important to note that knowledge of the mixing time is not necessary for the implementation of the SGD algorithm. Overall, our results show that under the conditions we specify, non-stationarity induced by the policy updates does not impose fundamental limitations on adaptivity.
%These rates are similar to the ones achieved with delicate batch updates \citep{chen2023online,huh2009adaptive, hu2022switchback}.

Our finite-time convergence analysis for SGD with adaptive data can be applied to study a wide range of problems of practical relevance. In particular, we demonstrate how the analysis can be applied to study online learning algorithms for service and inventory systems.  As mentioned before, our setting also covers policy-gradient approaches in RL, and we show how our analysis applies to an actor-critic policy gradient algorithm. These examples demonstrate that the assumptions we impose are easy to verify and widely applicable. 

When applying SGD with adaptive data to service and inventory management problems, we demonstrate how to construct gradient estimators using the sample path derivative, which can be updated recursively. In particular, we augment the Markov chain to include both the original Markov chain and a derivative process. This sample-path derivative construction utilizes developments in the simulation literature on infinitesimal perturbation analysis (IPA) \citep{heidelberger1988convergence, glasserman1992stationary}. However, in our applications of IPA, we need to establish stronger ergodicity results than the standard convergence results in the literature. 
%{\color{green} Do we?}

%{\color{blue} Need to carefully compare our work to \citep{li2022state} and \citep{roy2022constrained}. The main difference will be in the assumptions required.}

% {\color{blue} Finally, although our theoretical results identify conditions under which convergence of SGD in adaptive environments resembles that in the stationary iid case, our numerical experiments show that when these conditions do not hold, there can be a stark difference between the two settings. In particular, we note that small step-size, i.e., $1/\sqrt{t}$ or $1/t$, are important for ``practical" convergence in the adaptive setting, especially when the policy parameters have a large effect on the environment. 
% %Notably, for the joint pricing and capacity sizing problem in the $G/G/1$ queue, while using the theoretically recommended step-size of $\eta_{t} = \eta_{0}t^{-1}$ leads to predictably smooth convergence, setting a higher step-size of $\eta_{t} = \eta_{0}t^{-1/3}$ leads to instabilities that completely impede convergence. 
% This is in contrast to the iid setting where even constant step sizes are used in practice.} %These results show that setting \emph{small} step-sizes is crucial for convergence in the adaptive setting, especially when the policy parameters have a large effect on the environment.}

The rest of the paper is organized as follows. We conclude this section with a review of the literature to highlight our contribution. In Section \ref{sec:main}, we present our main results -- the finite-time performance bounds for SGD with adaptive data in various settings, i.e., nonconvex, convex with/without projection, and strongly convex with/without projection. In Sections \ref{sec:inventory} -- \ref{sec:RL}, we demonstrate how to apply the main results to study various online learning problems. We also discuss how to apply the sample path derivative to construct gradient estimators in Sections \ref{sec:inventory} and \ref{sec:queue}. Lastly, we conduct numerical experiments to demonstrate the performance of the SGD algorithm with adaptive data in various applications in Section \ref{sec:num} and conclude in Section \ref{sec:conclude}. All the proofs are provided in the appendices.

\subsection*{Literature Review} In this paper, we study the convergence rate of the SGD updates with adaptive data:
\[
\theta_{t+1}=\theta_t-\eta_t g(\theta_t,z_t)
\]
where $z_t\sim P_{\theta_t}(\cdot|z_{t-1})$, $g(\theta_t,z_t)$ is a gradient estimator, and we assume $\nabla l(\theta)=\mathbb{E}_{\mu_\theta}[g(\theta, z)]$. 
As discussed above, motivated by different applications, most previous literature on SGD either assumes iid data or dependent but non-adaptive data (see, e.g., \citep{moulines2011non,ghadimi2013stochastic, agarwal2012generalization, chen2023lyapunov}).
Extending stochastic gradient/ stochastic approximation algorithms to adaptive data is an active area of research \citep{benveniste2012adaptive}. When the data is drawn adaptively from a policy-dependent Markov chain, some simple gradient estimators can suffer from estimation biases even when evaluated according to the corresponding stationary distribution. \cite{doucet2017asymptotic} studies the asymptotic behavior of SGD with a biased gradient estimator. \cite{karimi2019non, li2022state, roy2022constrained} further establishes non-asymptotic convergence results of SGD with adaptive data. 
They demonstrate how to apply the results to analyze the regularized online expectation maximization algorithm, policy gradient method, and strategic classification.
Recently, \cite{li2023stochastic} demonstrate how to apply the framework in \citep{benveniste2012adaptive, karimi2019non} to optimize the design of GI/GI/1 queues. 
Similar to \citep{karimi2019non, li2022state, roy2022constrained}, our work also establishes non-asymptotic performance bounds. However, we require a different set of assumptions that are easier to verify and can be applied to many online learning problems in operations research. We next discuss some key differences.

First, most of the previous works require the stochastic gradient estimator, $g(\theta,z)$, to be Lipschitz continuous in $\theta$ (see, e.g., Assumptions 1 and 2 in \cite{li2022state} and Assumption 2.4 in \cite{roy2022constrained}). 
However, this assumption does not hold in many applications, e.g., when $g(\theta,z)$ involves indicators. 
%we studied (see Section \ref{sec:OR}), since the stochastic gradients contain indicators involving $\theta$ and $z$. 
To handle this challenge, our results only require the population gradient $\nabla l(\theta)$ to be Lipschitz. This is a much weaker assumption since $\nabla l$ is the weighted average of $g(\theta,z)$'s, and its smoothness is in general satisfied 
(see, e.g., Proposition 2.1 of \cite{roy2022constrained}). Moreover, our analysis leverages the recent developments in the perturbation theory for Markov chains \citep{rudolf2018perturbation} to explain why $\nabla l$ is Lipschitz in general.

Second, most of the existing works require finding the solution of a Poisson equation associated with $P_{\theta}$ and $g(\theta,\cdot)$ and verifying certain continuity properties of the Poisson equation solution (see, e.g., Assumptions A5 and A6 in  \citep{karimi2019non} and the development in \citep{li2023stochastic}). In many applications, it is hard to find the solution to the Poisson equation. In particular, solving the Poisson equation is as hard as solving the original problem. 
%and verifying the continuity property of the Poisson equation solution can be even more challenging. 
Our results do not require these. The assumptions we impose are rather standard Markov chain mixing properties which are fairly well known for many operations research applications. 

Third, our work also provides a more complete picture of the convergence of SGD with adaptive data (nonconvex, convex with/without projection, and strongly convex with/without projection), while previous literature often analyzes convergence under only one or two settings. In addition, with non-convex loss, the estimate (Corollary 1) in  \cite{li2022state} has a nonvanishing bias, while \cite{roy2022constrained} gives only $\tilde O(T^{-2/5})$ convergence rate, which is slower than the standard $\tilde O(T^{-1/2})$ rate. Our analysis combines Markov chain perturbation theory with IPA to achieve a faster rate of convergence.

From the application perspective, our work is related to the online learning literature in operations research, especially in queueing and inventory systems. This has been an active area of research. For example, \cite{chen2023online} study learning pricing and capacity sizing in single server queues, \cite{krishnasamy2021learning, zhong2022learning} study learning the scheduling policy in multiclass queues. See \cite{walton2021learning} for a review of recent developments in learning in stochastic networks. 
\cite{huh2009adaptive, zhang2020closing} study learning the replenishment policy in lost sales inventory systems. 
\cite{tang2023online} study learning dual-index policies in dual sourcing systems. \cite{cheung2022inventory} study learning how to allocate limited resources to heterogeneous customers. Our work complements these works by taking a closer look at how many samples need to be collected before a policy update, i.e., how often the policy can be updated. We provide easy-to-verify conditions under which updating the policy after each new sample or a constant batch of samples leads to a near-optimal rate of convergence.

Our work is also related to the literature on policy gradient in reinforcement learning, especially recent developments on sample complexity analysis of policy gradient algorithms \citep{wang2019neural, zhang2020global, xiong2021non, yuan2022general, xu2020improving}. The convergence of policy gradient with exact gradient information --thus eliminating approximation errors -- has been studied in several different settings \citep{mei2020global, agarwal2021theory, xiao2022convergence, bhandari2024global}. 
%Under suitable regularity conditions, these works not only show convergence to a first-order stationary point but actually to a global minimum. 
However, when implementing policy gradient methods, a key challenge is how to estimate the gradient in a sample-efficient manner. For example, if we are to estimate the gradient by sampling the trajectories of the MDP under the current policy $\pi^{\theta}$, what would be the horizon for the trajectory (i.e., where to truncate since we cannot sample an infinite-horizon trajectory) and how many trajectories do we need to sample? Note that in addition to the standard stochastic noise, we also need to handle the bias due to truncation (i.e., the underlying Markov chain has not reached stationarity yet). In this paper, we demonstrate how our finite-time convergence analysis for SGD with adaptive data can be applied to study an actor-critic policy gradient algorithm, where we use temporal difference learning (TD) to estimate the state-action value function/Q-function under policy $\pi^{\theta}$. Our algorithm only requires one TD update in each iteration, which is substantially less than what is required in \cite{wang2019neural, yuan2022general, xiong2021non, xu2020improving}. To achieve this, we study the Markov chain $(s_t,a_t,Q_t)$ induced by the TD sampler with a constant learning rate, where $s_t$ and $a_t$ are the state and action visited and $Q_t$ is the state-action value function (Q-function) at time $t$. We show that even though $Q_t$ does not converge to the desired Q-function, i.e., the Q-function under policy $\pi^{\theta}$, $Q_t$ evaluated under the corresponding stationary measure is equal to the desired Q-function. 

The SGD-based methods/analyses have also been applied to conduct finite-time analysis of TD algorithms \citep{bhandari2018finite,dalal2018finite, srikant2019finite, qu2020finite}. There, because the policy is fixed, i.e., the Markov chain dynamics is fixed, the data is Markovian but not adaptive, which is similar to the setting studied \citep{duchi2012ergodic,agarwal2012generalization, sun2018markov}.

% on the non-asymptotic analysis of temporal difference (TD) learning \citep{bhandari2018finite,dalal2018finite, srikant2019finite, qu2020finite}. These works study a special linear stochastic approximation and focus on specific objective functions that arise in the reinforcement learning context. In contrast, our results can be applied more broadly. For the policy gradient application, we study the Markov chain $(s_t,a_t,Q_t)$ induced by the TD sampler, where $s_t$ and $a_t$ are the state and action visited and $Q_t$ is the state-action value function (Q-function) at time $t$. We show that even though $Q_t$ does not converge to the desired Q-function, i.e., the Q-function under policy $\pi^{\theta}$, $Q_t$ evaluated under the corresponding stationary measure (of the Markov chain $(s_t,a_t,Q_t)$) is equal to the desired Q-function. This result can be of independent interest. {\color{red} Do we need to expand this a bit more?}

\subsection*{Notations} Let $(\Omega, d)$ denote a metric space for the data stream $z$. We denote $\|\cdot\|$ as the $L^2$ norm. 
%and $\|\cdot\|_{\infty}$ as the $L^{\infty}$ norm. 
For a transition kernel $P$, we write 
$Pf(x)=\int P(x,dx')f(x')$,
%$Pf(x)=\sum_{x'} P(x,x')f(x')$, \blue{[This indicates we are doing finite state? Use $Pf(x)=\int P(x,dx')f(x')$? ]} 
and we write the distribution of the Markov chain starting from $x$ after $n$ steps of transition as $\delta_x P^n$. For a nonnegative sequence $\{a_t: t\geq 0\}$ and a nonnegative function $f(t)$, we say $a_t=O(f(t))$ if there is a constant $C$ so that $a_t\leq Cf(t)$ for any $t\geq 0$. We also use the notation $a_t=\tilde O(f(t))$ when we ignore the logrighmic factors, e.g., when $a_t=(f(t)(\log f(t))^k)$, we can write $a_t=\tilde O(f(t))$.

\section{Main Results} \label{sec:main}
To solve the stochastic optimization problem \eqref{eq:main}, we consider the SGD update:
\begin{equation}\label{eq:genSGD}
\theta_{t+1}=\theta_t-\eta_t \ghat_t(\theta_t,z_t).
\end{equation}
We assume the data sequence comes from a Markov chain controlled by $\theta_t$. In particular, we assume there is a transition kernel $P_{\theta_t}$ on $\Omega$ such that
\[
z_{t}\sim P_{\theta_t}(\cdot|z_{t-1}).
\]
For each given $\theta$, $P_{\theta}$ generates an ergodic Markov chain with invariant measure $\mu_{\theta}$. In addition, we assume there is a stochastic gradient estimator $g(\theta,z_t)$ satisfying
\[
\nabla \ell(\theta)=\mathbb{E}_{\mu_\theta}[g(\theta, z)].
\]
We will discuss how to find such a gradient estimator using the sample path derivative in Sections \ref{sec:inventory} and \ref{sec:queue} for some examples. Note that $g(\theta,z_t)$ is unbiased only under the corresponding invariant measure. Due to the transience and non-stationarity of the underlying Markov chain, $\mathbb{E}[g(\theta_t,z_t)|\theta_t,z_{t-1}]\neq \nabla \ell(\theta_t)$ in general. 
%This is because $z_t$ is not generated from the stationary distribution $\mu_{\theta_t}$ due to the transience of the underlying Markov chain. Thus, we have to take the transient behavior of the Markov chain into account. 
Meanwhile, to accommodate more general settings, we assume $\ghat_t(\theta,z_t)$ in \eqref{eq:genSGD} is only an approximation of $g(\theta,z_t)$ with diminishing errors. 
%But first-time readers can assume $\ghat=g$, since it holds in most applications.  

We make the following assumptions.
\begin{assumption}\label{ass:ergodicity0}
$P_{\theta}$'s have a common Lyapunov function $V\geq 1$ where
\[
P_\theta V(z)\leq \rho V(z)+K,
\]
for some $\rho \in(0,1)$ and $K\in(1,\infty)$.
In addition, $P_{\theta}$'s are Wasserstein contractive with respect to the metric $d$ on $\Omega$, i.e., for any $x,y\in \Omega$,
\[
d(\delta_x P^n_{\theta},\delta_y P^n_{\theta} )\leq K\rho^n d(x,y).
\] 
\end{assumption}

\begin{assumption} \label{ass:lip0}
There exists $L>0$ such that
\[
%\|\E[g(\theta, z)-g(\theta^{\prime}, z)]\| \leq L\|\theta-\theta^{\prime}\|, ~~~
\|\nabla \ell(\theta)-\nabla \ell(\theta^{\prime})\| \leq L\|\theta-\theta^{\prime}\|,
\]
\[
\|g(\theta, z)-g(\theta, z^{\prime})\|\leq Ld(z,z^{\prime}),\]
and
\[
d(\delta_x P_{\theta},\delta_x P_{\theta^{\prime}} )\leq L\|\theta-\theta^{\prime}\|V(x).
\]
\end{assumption}
 
\begin{assumption} \label{ass:bound0}
There exists  $M\in(0,\infty)$, such that 
\[
\|g(\theta, z)\|\leq MV(z), \|\ghat_t(\theta, z)\| \leq MV(z), \mbox{ and } \|\nabla l(\theta)\| \leq M.
\]
\end{assumption}
%\red{Do we assume dim$(\theta)=O(1)$? Else we assume $\|g\|\leq M$?}

\begin{assumption} \label{ass:bound1}
There exits $C>0$, such that for any $t\geq 0$,
\[
%\E [V(z_t)^4]\leq C(V(z_0)^4+1).
P_{\theta}^t V(z) \leq C(V(z)^4+1).
\]
\end{assumption}
Note that a sufficient condition for Assumption \ref{ass:bound1} is that $V(z)^4$ is also a Lyapunov function.

\begin{assumption} \label{ass:error0}
There exists a stochastic sequence $e_t$ with $\E[e_t^2]<\infty$, such that
\[
\|g(\theta,z_t)-\ghat_t(\theta,z_t)\|\leq e_t.
\]
\end{assumption}

For the ease of exposition, we also introduce the concept of mixing time.
\begin{definition} The Markov chain $P_{\theta}$ has a mixing time $\tau<\infty$ if for any $z\in \Omega$,
\[
d(\delta_{z} P_{\theta}^{\tau},\mu_{\theta}(\cdot))\leq \frac{1}{4}V(z).
\]
\end{definition}

Assumption \ref{ass:ergodicity0} requires that the underlying Markov chains are suitably ergodic.
This assumption ensures that the stationary distribution $\mu_{\theta}$ is well defined and the Markov chain converges to stationarity sufficiently fast. In particular, the existence of the Lyapunov function implies that the dynamics return to the “center” of the state space regularly and the length of the excursions from the center can be properly controlled. Note that by Harris' ergodic theorem, the existence of the Lyapunov function together with a uniform “minorization” condition localized to the
interior of a level set implies geometric ergodicity and thus Wasserstein contraction under an appropriate metric \citep{roberts1997geometric}. This Markov chain convergence framework has been well-studied in the literature \citep{meyn2012markov}. For many existing models, e.g., queueing models, inventory models, etc, Assumption \ref{ass:ergodicity0} has already been verified. 

The first condition in Assumption \ref{ass:lip0} requires $\nabla l$ to be Lipschitz continuous, which is satisfied in many applications. This condition is weaker and more applicable than requiring $g(\theta, z)$ to be Lipschitz continuous in $\theta$ (see Proposition 2.1 of \cite{roy2022constrained}). For example, $g(\theta, z)$ is not Lipschitz in $\theta$ if $g(\theta, z)=1\{z\leq \theta\}$. We also require $g(\theta,z)$ to be Lipschitz continuous in $z$, but have some flexibility in choosing the metric $d$.
In particular, by choosing an appropriate $d$, this condition is easily satisfied in many applications. For example, if we consider the total variation distance, $d(z,z')=2\cdot1\{z\neq z'\}$, then this condition will be a consequence of Assumption \ref{ass:bound0}. Lastly, we require $\delta_x P_{\theta}$ to be Lipschitz continuous in $\theta$. Note that this condition is for the one-step transition kernel, which is easy to verify in practice. By Markov chain perturbation theory, the Lipschitz continuity of the one-step transition kernel together with the ergodicity condition, i.e, Assumption \ref{ass:ergodicity0}, implies the Lipschitz continuity of the corresponding stationary measure \citep{rudolf2018perturbation}.

Assumptions \ref{ass:bound0} and \ref{ass:error0} are boundedness assumptions, which are weaker than those for gradient algorithms with Markovian or adaptive data streams in the literature \citep{sun2018markov,karimi2019non}. In particular, we only require $g(\theta,z)$ to be bounded by $MV(z)$, instead of $M$.

\textbf{Nonconvex case:}
We have the following convergence result for a general loss function $l$.
\begin{theorem}
\label{thm:main1}
Suppose Assumptions \ref{ass:ergodicity0} -- \ref{ass:error0} hold.
%\red{assume the gradient from the beginning.}
%we have
%\[
%\sum_{t=0}^{T-1}\eta_t\E \langle \nabla l(\theta_t),\bar{g}(\theta_t)\rangle=O\left(\tau\log T + \tau\log T\sum_{t=0}^{T-%1}\eta_t^2+\frac{1}{\sqrt{T}}\sum_{t=0}^{T-1}\eta_t+\sum_{t=0}^{T-1}\eta_t\E e_t \right).
%\]
%If $\bar g(\theta)=\nabla l(\theta)$,  \eqref{eq:genSGD} is the SGD update with adaptive Markov data. 
The iterates according to \eqref{eq:genSGD} satsifies
%will find a stationary point. Precisely,
\[
\min_{0\leq t< T}\E \| \nabla \ell(\theta_t)\|^2=O\left(\frac{1}{\sum_{t=0}^{T-1}\eta_t}\left(\tau\log T + \tau\log T\sum_{t=0}^{T-1}\eta_t^2+\frac{1}{\sqrt{T}}\sum_{t=0}^{T-1}\eta_t+\sum_{t=0}^{T-1}\eta_t\E e_t \right)\right),
\]
where $O$ hides a polynomial of $M$ and $L$. 
%\item[2.] If $l$ is convex, the weighted average of iterates satisfies
% \[
% \E l(\bar{\theta}_T)-l(\theta^*)=O\left(\frac{1}{\sum_{t=0}^{T-1}\eta_t}\left(\tau\log T + \tau\log T\sum_{t=0}^{T-1}\eta_t^2+\frac{1}{\sqrt{T}}\sum_{t=0}^{T-1}\eta_t+\sum_{t=0}^{T-1}\eta_t\E e_t \right)\right).
% \]
If we fix $\eta_t=\eta_0 t^{-1/2}$ for some $\eta_0>0$, and assume $\E e_t =O(1/\sqrt{t})$, we can further simplify the bound to 
%\[\E l(\bar{\theta}_T)-l(\theta^*)=O\left(\tau (\log T)^2/\sqrt{T}\right),\]
%where $\theta^*$ is the converging stationary point.
\[
\min_{0\leq t< T}\E \| \nabla \ell(\theta_t)\|^2=O\left(\tau (\log T)^2/\sqrt{T}\right).
\]
\end{theorem}

Theorem \ref{thm:main1} implies that the SGD updates can achieve an $O(\tau(\log T)^2/\sqrt{T})$ convergence to a stationary point, which is similar to the $O(\log(T)/\sqrt{T})$ convergence in Theorem 2 of \citep{karimi2019non}, for which they also establish a matching lower bound. Here, we explicitly characterize how the mixing time affects the convergence rate. The convergence results in \citep{karimi2019non} have an extra asymptotic bias term which does not arise in our setting since our result works under a different set of assumptions and a more refined gradient estimator. 

% %$\min_{0\leq t< T}\E \| \nabla l(\theta_t)\|^2=O\left(\tau (\log T)^2/\sqrt{T}\right)$.
% \item[3.] If $l$ is strongly convex with convexity constant $c$, i.e., $\langle \nabla l(\theta)-\nabla l(\theta'), \theta-\theta'\rangle \geq c\|\theta-\theta'\|^2$
% and we set the step size as
% $\eta_{t}=2\eta_0/(ct)$ for $t\geq 1$ with $\eta_0>2$,
% the iterates according to \eqref{eq:genSGD} satisfy
% \[
% \E\|\theta_{T}-\theta^*\|^2=O\left(\frac{\tau(\log T)^2}{T} + \frac{1}{T}\sum_{t=1}^{T}\E e_t^2\right),
% \]
% where $O$ hides a polynomial of $C$, $M$, and $L$. 
% If $\E e_t^2=O(\frac 1t)$, we can further simplify the bound to
% $\E\|\theta_{T}-\theta^*\|^2=O\left(\tau(\log T)^2/T\right)$.
% \end{itemize}
% \end{theorem}

\textbf{Convex case:}
For general (constrained) convex  optimization, we consider a projected SGD update: 
\begin{equation}\label{eq:genSGD2}
\theta_{t+1}=\bfP_{\Theta}(\theta_t-\eta_t \ghat_t(\theta_t,z_t)),
\end{equation}
where $\Theta$ is a properly defined convex set, and $\bfP_{\Theta}$ is the associated projection. The set $\Theta$ can be the constrained set for constrained optimization, or $\mathbb{R}^m$ in the unconstrained setting.
The projection allows us to relax Assumptions \ref{ass:ergodicity0} -- \ref{ass:error0} such that they only need to hold for $\theta\in\Theta$.

% \begin{assumption} \label{ass:projection}
% %There exists $C>0$ such that $\mathcal{C} \subset\{\theta:\|\theta\|\leq C\}$.
% Assumption \ref{ass:bound0} -- \ref{ass:lip0} hold under the restriction that $\theta\in\Theta$.  
% In addition, for any $t\geq 0$
% \[
% \E [V(z_t)^2]\leq C^2(V(z_0)^2+1),
% \]
% for some $C>0$.
% \end{assumption}

%Note that a sufficient condition for $\E V(z_t)^2\leq C^2(V(z_0)^2+1)$ is that $V(z)^2$ is also a Lyapunov function.

%In the case of projected SGD updates, we can establish the following convergence result.
Define the weighted average of the iterates as
%\begin{equation}
%\label{eq:avg_iterate}
\[
\bar{\theta}_T =\frac{1}{\sum_{t=0}^{T-1}\eta_t}\sum_{t=0}^{T-1}\eta_t\theta_t.
\]
%\end{equation}

\begin{theorem}
\label{thm:main2}
Suppose Assumptions \ref{ass:ergodicity0} -- \ref{ass:error0} hold under the restriction that $\theta\in\Theta$.
%Assumptions \ref{ass:projection} holds.
\begin{itemize}
\item[1.] If $\ell$ is convex and $\E [e_t] =O(1/\sqrt{t})$, by setting the step size as
$\eta_{t}=\eta_0/\sqrt{t}$, the weighted average of iterates according to \eqref{eq:genSGD2} satisfy
%\[
%  \E l(\bar{\theta}_T)-l(\theta^*)
%  =O\left(\frac{\tau(\log T)^2}{\sqrt{T}}
%  +\frac{1}{\sqrt{T}}\sum_{t=0}^{T-1}\frac{1}{\sqrt{t}}\E e_t\right)
% \]
% \[
% \E l(\bar{\theta}_T)-l(\theta^*)=O\left(\frac{1}{\sum_{t=0}^{T-1}\eta_t}\left(\tau\log T + \tau\log T\sum_{t=0}^{T-1}\eta_t^2+\frac{1}{\sqrt{T}}\sum_{t=0}^{T-1}\eta_t+\sum_{t=0}^{T-1}\eta_t\E e_t \right)\right),
% \]
\[\E \ell(\bar{\theta}_T)-\ell(\theta^*)=O\left(\tau^2 (\log T)^4/\sqrt{T}\right),\]
where $O$ hides a polynomial of $M$ and $L$. 
%If we fix $\eta_t=\eta_0 t^{-1/2}$ for some $\eta_0>0$, and assume 
% If $\E e_t =O(1/\sqrt{t})$, 
% we can further simplify the bound to 
% \[\E l(\bar{\theta}_T)-l(\theta^*)=O\left(\tau (\log T)^2/\sqrt{T}\right).\]
\item[2.] If $\ell$ is strongly convex with a convexity constant $c$ and $\E [e_t^2]=O(1/t)$, by setting the step size as
$\eta_{t}=2\eta_0/(ct)$ for $t\geq 1$ with $\eta_0>2$,
the iterates according to \eqref{eq:genSGD2} satisfy
% \[
% \E\|\theta_{T}-\theta^*\|^2=O\left(\frac{\tau(\log T)^2}{T} + \frac{1}{T}\sum_{t=1}^{T}\E e_t^2\right),
% \]
\[\E\|\theta_{T}-\theta^*\|^2=O\left(\tau(\log T)^2/T\right),\]
where $O$ hides a polynomial of $M$ and $L$. 
% If $\E e_t^2=O(\frac 1t)$, we can further simplify the bound to
% \[\E\|\theta_{T}-\theta^*\|^2=O\left(\tau(\log T)^2/T\right).\]
\end{itemize}
\end{theorem}

 Theorem \ref{thm:main2} indicates that when $\ell$ is strongly convex, we achieve $O(\tau(\log T)^2/T)$ convergence rate, which is similar to the $O(1/T)$ convergence rate established in Thereom 1 of \citep{li2022state}. %However, when we relax the assumption of strong convexity, the convergence results in \citep{li2022state} has an asymptotic bias term, which does not arise in our result. This is due to a different set of assumptions and a more refined analysis based on the Markov chain perturbation theory.

In what follows, we will demonstrate how to apply Theorems \ref{thm:main1} and \ref{thm:main2} to various applications. In particular, we will show that the assumptions required in the theorems are satisfied and easy to verify in many online learning problems in operations research. We will also discuss an extension to an actor-critic policy gradient algorithm.

%\subsection{Simple based stock and demand with memory?}
\section{Inventory Control with Stock-Out Damping}\label{sec:inventory}

We consider the problem of selecting a base-stock level in a single-product multi-period inventory system with endogenous demand. Motivated by recent empirical findings, demand is
temporarily reduced whenever a stock-out occurs \citep{anderson2006measuring}. We model demand as
a Markovian autoregressive process subject to a dampening
effect when demand exceeds the base-stock level:
\begin{equation}
D_{t+1}=(\alpha\min\{D_{t}+u_{t+1},\theta\}+(1-\alpha)m+\epsilon_{t+1})^{+},\label{eq:newsvendor-demand}
\end{equation}
where $\epsilon_{t}\stackrel{iid}{\sim}N(0,\sigma^{2})$, $u_{t}\stackrel{iid}{\sim}N(0,\sigma^{2})$,
$u_{t}$ is independent of $\epsilon_{t}$, and $(\cdot)^{+}=\max\{\cdot,0\}$. 
The goal is to select the base-stock level $\theta\in\Theta:=[0,\bar{\theta}]$, for any finite upper bound $\bar\theta>0$, to minimize the underage and overage costs under the stationary distribution induced by $\theta$:
\begin{equation}
\min_{\theta\in\Theta}\ell(\theta):= \mathbb{E}_{\mu_{\theta}}\left[h(\theta-D_{t})^{+}+b(D_{t}-\theta)^{+}\right]\label{eq:newsvendor-loss}
\end{equation}

Note that the base-stock level endogenously affects the demand dynamics and we write $D_t(\theta)$ to mark the dependence explicitly when necessary. To obtain an appropriate gradient estimator, we consider taking a pathwise derivative of $D_t(\theta)$:
\begin{align*}
L_{t+1}(\theta):=\frac{dD_{t+1}(\theta)}{d\theta} 
=&\frac{\partial}{\partial D_{t}}
(\alpha\min\{D_{t}(\theta)+u_{t+1},\theta\}+(1-\alpha)m+\epsilon_{t+1})^{+}\frac{dD_{t}(\theta)}{d\theta}\\
&+\frac{\partial}{\partial\theta}(\alpha\min\{D_{t}(\theta)+u_{t+1},\theta\}+(1-\alpha)m+\epsilon_{t+1})^{+}\\
=&\alpha1\{D_{t+1}(\theta)>0\}1\{D_{t}(\theta)+u_{t+1}\leq\theta\}L_t(\theta)\\
 &+\alpha1\{D_{t+1}(\theta)>0\}1\{\theta<D_{t}(\theta)+u_{t+1}\}
%=&1\{D_{t+1}(\theta)>0\}\alpha\left(1\{\theta<D_{t}(\theta)+\eta_{t+1}\}+1\{D_{t}(\theta)+\eta_{t+1}\leq\theta\}L_t(\theta)\right)\label{eq:grad_recursion}
\end{align*}
This gives us a recursive way to update $L_t$, i.e.,
\[
L_{t+1}(\theta)=1\{D_{t+1}(\theta)>0\}\alpha\left(1\{D_{t}(\theta)+u_{t+1}>\theta\}+1\{D_{t}(\theta)+u_{t+1}\leq\theta\}L_{t}(\theta)\right).
\]

%the gradient of $\ell(\theta)$ with respect to $\theta$ includes
%a term $L_{t}:=\frac{dD_{t}}{d\theta}$, which is a pathwise derivative of demand with respect to the base-stock level $\theta$:

Consider the augmented Markov chain $Z_t(\theta)=(D_t(\theta), L_t(\theta))$. Our next result shows that $Z_t(\theta)$ is well defined with a proper stationary
distribution. 
\begin{lemma}
\label{lem:demand-derivative}
For any $\theta\in\Theta$, the pathwise
derivative $L_{t}(\theta)$ exists and the Markov chain $Z_t(\theta)=(D_{t}(\theta),L_{t}(\theta))$
converges in distribution to $Z_{\infty}(\theta)=(D_{\infty}(\theta),L_{\infty}(\theta))$ as $t\to\infty$ with 
%$\mathbb{E}[Z_{\infty}(\theta)]<\infty$ and
%=\mathbb{E}_{\mu_{\infty}}[L_{t}]<\infty$ and
\[
\nabla\ell(\theta)=\mathbb{E}\left[\left(h1\{D_{\infty}(\theta)<\theta\}-b1\{D_{\infty}(\theta)\geq\theta\}\right)\left(1-L_{\infty}(\theta)\right)\right].\label{eq:newsvendor-stat-grad}
\]
%Moreover, $\mathbb{E}[L_{\infty}]=\frac{d}{d\theta}\mathbb{E}[D_{\infty}]$.
\end{lemma}

% \begin{proof}
% See Appendix \ref{sec:demand-derivative-proof}.
% \end{proof}
% The derivative process $L_{t}$ follows its own dynamics, which can
% be used to estimate its stationary value, initializing with $L_{0}=0$.

% \[
% L_{t+1}=1\{D_{t+1}>0\}\alpha\left(1\{\theta<D_{t}+\eta_{t+1}\}+1\{D_{t}+\eta_{t+1}\leq\theta\}L_{t}\right)
% \]
% We consider the augmented augmented Markov chain $Z_{t}=(D_{t},L_{t})$,
% which gives the following gradient estimator:
Based on Lemma \ref{lem:demand-derivative}, we have the following gradient estimator:
\begin{equation}
g(\theta,Z_{t})=\left(h1\{D_{t}<\theta\}-b1\{D_{t}\geq\theta\}\right)\left(1-L_{t}\right).\label{eq:base-stock-grad}
\end{equation}
% Note that this gradient estimator does not require observing the realization of the demand
% when a stock-out occurs, making it compatible with censored demand. 
The SGD update using the gradient estimator
defined in \eqref{eq:base-stock-grad} with step-size $\eta_{t}$ then takes the form
%and starting from some $\theta_{0}\in\Theta=[0,\theta]$.
%\begin{equation}
\[
\theta_{t+1}=\mathcal{P}_{\Theta}\left(\theta_{t}-\eta_{t}g(\theta_t,Z_{t})\right). 
\]
%\label{eq:newsvendor-sgd}
%\end{equation}

% \begin{theorem}
% \label{thm:newsvendor-conv}Let $\theta^{*}$ be a stationary point
% of \eqref{eq:newsvendor-loss}. Under gradient estimator $\tilde{g}(\theta,Z_{t})$
% defined in (\ref{eq:base-stock-grad}) with step-size $\eta_{t}=\eta_{0}t^{-1/2}$,
% the average iterate $\bar{\theta}_{T}$ of (\ref{eq:newsvendor-sgd})
% after observing the demand for $T$ periods satisfies $\mathbb{E}[\ell(\theta)]-\mathbb{E}[\ell(\theta^{*})]=O\left(\frac{\tau(\log T)}{2\sqrt{T}}\right)$
% where $\tau$ is an upper bound of mixing time of the demand over
% $\Theta$.
% \end{theorem}

\begin{theorem}
\label{thm:newsvendor-conv}
Suppose the objective function \eqref{eq:newsvendor-loss} is convex and let $\theta^*$ be the minimizer. Then, using the gradient estimator $g(\theta,Z_{t})$ defined in \eqref{eq:base-stock-grad} with step-size $\eta_{t}=\eta_{0}t^{-1/2}$, the average SGD iterate $\bar{\theta}_{T}$ satisfies
\[\E [\ell(\bar{\theta}_T)-\ell(\theta^*)]=O\left(\tau^2 (\log T)^4/\sqrt{T}\right),\]
where $\tau$ is an upper bound of the mixing time of the augmented Markov chain $Z_t(\theta)$ for $\theta\in\Theta$.\\
Moreover, if the objective function  is strongly convex with a convexity constant $c$, by setting the step size as
$\eta_{t}=2\eta_0/(ct)$ for $t\geq 1$ with $\eta_0>2$, we have
\[\E[\|\theta_{T}-\theta^*\|^2]=O\left(\tau(\log T)^2/T\right).\]
\end{theorem}
%{\color{red} See if we can check the convexity.}

% \begin{proof}
% See Appendix \ref{sec:newsvendor-conv-proof}.
% \end{proof}
Note that the existing approaches, i.e., those in  \citep{li2022state, roy2022constrained}, require Lipschitzness of the gradient
estimator $g$ in both $\theta$ and $z$ under the Euclidean distance, which does not hold in this example.
%the gradient estimator $g$ is not Lipschitz in either variable under the Euclidean distance, and yet we recover similar rates of convergence. \blue{[The previous sentence doens't read. Please rephrase]}This is because 
In contrast, our framework only requires Lipschitzness
of $\nabla\ell(\theta)$ in $\theta$.
While the gradient estimator is highly non-smooth, the averaged gradient is Lipschitz. In fact, the Lipschitzness of $\nabla\ell(\theta)$ is a direct consequence of the Lipschitzness of the transition kernel $P_{\theta}$, as we show in the proof of Theorem \ref{thm:newsvendor-conv}. At a high level, our framework reveals that randomness in the transition
dynamics implicitly smooths the gradient estimator, enabling a greater range of gradient estimators while maintaining convergence guarantees.

\section{Pricing and Capacity Sizing in Single-Server Queue} \label{sec:queue}

We consider the problem of pricing and capacity sizing in a single-server queue, as studied in \cite{chen2023online}. Consider a $GI/GI/1$ queue, i.e., a single-sever queue with generally distributed interarrival times and service times. The interarrival times and service times are scaled by the arrival rate and service rate respectively. The arrival rate is determined by the price charged according to a known demand function $\lambda(p)$ for feasible prices $p\in[\underline{p},\bar{p}]$. The service provider also selects the service rate $\mu\in[\underline{\mu},\bar{\mu}]$, which incurs a service cost $c(\mu)$ per unit of time. Let $T_{t+1}$ denote the baseline interarrival time between the $t$-th and $(t+1)$-th arrivals, $S_{t}$ denote the baseline service time, and $W_{t}$ denote the waiting time of customer $t$. For given $\mu$ and $p$, the system
dynamics follow
\begin{align}
W_{t+1} & =\left(W_{t}+\frac{S_{t}}{\mu}-\frac{T_{t+1}}{\lambda(p)}\right)^{+}.\label{eq:gg1}
\end{align}
Let $\Theta=[\underline{\mu},\bar{\mu}]\times[\underline{p},\bar{p}]$
denote the feasible set of service rates and prices. The goal of the service provider is to select $(\mu,p)\in\Theta$
to maximize expected net profit, which is the long-run average revenue minus the service
cost and holding cost:
\begin{align}
\label{eq:gg1_orig_objective}
\max_{(\mu,p)\in\Theta}V(\mu,p):= p\lambda(p)-c(\mu)-h_{0}\mathbb{E}[Q_{\infty}(\mu,p)],
\end{align}
where $Q_{\infty}(\mu,p)$ is the stationary queue length (number of people waiting in the system) under $(\mu,p)$, and $h_0$ is the per-unit-time per-customer holding cost (cost of waiting).
By Little's law, the maximization problem in \eqref{eq:gg1_orig_objective} is equivalent to the following minimization
problem involving the stationary waiting time $W_{\infty}(\mu,p)$
under $(\mu,p)$:
\begin{align}
\min_{(\mu,p)\in\Theta} \ell(\mu,p) := h_{0}\lambda(p)\left(\mathbb{E}\left[W_{\infty}(\mu,p)\right]+\frac{1}{\mu}\right)+c(\mu)-p\lambda(p)\label{eq:gg1_objective}
\end{align}

To obtain an appropriate gradient estimator, we consider taking a pathwise derivative of $W_t(\mu,p)$. Define 
\[\begin{split}
L_{\mu,t+1}(\mu,p):=\frac{\partial W_{t+1}(\mu,p)}{\partial \mu} &= \left(\frac{\partial W_{t}(\mu,p)}{\partial \mu}-\frac{S_t}{\mu^2}\right)1\{W_{t+1}(\mu,p)>0\}\\
L_{p,t+1}(\mu,p):=\frac{\partial W_{t+1}(\mu,p)}{\partial p} &= \left(\frac{\partial W_{t}(\mu,p)}{\partial p}+\frac{T_t}{\lambda(p)^2}\lambda^{\prime}(p)\right)1\{W_{t+1}(\mu,p)>0\}
\end{split}\]
%Let $D_{\mu,t}(\mu,p)=\frac{\partial W_{t}(\mu,p)}{\partial \mu}$ and $D_{p,t}(\mu,p)=\frac{\partial W_{t}(\mu,p)}{\partial p}$. 
For the augmented Markov chain $(W_t,L_{\mu,t}, L_{p,t})$, by verifying the conditions in \cite{glasserman1992stationary}, we can show that $L_{\mu,t}$ and $L_{p,t}$ are well-defined, the Markov chain $(W_t,L_{\mu,t}, L_{p,t})$ converges to a unique stationary distribution, and  
\[\E[L_{\mu,\infty}(\mu,p)]=\frac{\partial \E[W_{\infty}(\mu,p)]}{\partial \mu},  ~~~ \E[L_{p,\infty}(\mu,p)]=\frac{\partial \E[W_{\infty}(\mu,p)]}{\partial p}.\]
%We will discuss this pathwise construction in more detail in Section \ref{sec:gradient}.
Indeed, this has been studied in \cite{chen2023online}.
For the single-server queue, we can achieve further simplification of the derivative processes by considering a simpler augmented Markov chain $Z_t=(W_t, X_t)$, where $X_t$ denotes the server's busy time seen by the $t$-th arrival. In particular, the dynamics of $Z_t$ are as follows:
\begin{align*}
W_{t+1} & =\left(W_{t}+\frac{S_{t}}{\mu}-\frac{T_{t}}{\lambda(p)}\right)^{+}\\
X_{t+1} & =\left(X_{t}+\frac{T_{t}}{\lambda(p)}\right)1\{W_{t+1}>0\}.
\end{align*}
Lemma 5 in \cite{chen2023online} shows that
\begin{equation}\label{eq:gg1_gradient0}
\begin{split}
\frac{\partial}{\partial p}\ell(\mu,p) & =-\lambda(p)-p\lambda'(p)+h_{0}\lambda'(p)\left(\mathbb{E}[W_{\infty}(\mu,p)]+\mathbb{E}[X_{\infty}(\mu,p)]+\frac{1}{\mu}\right)\\
\frac{\partial}{\partial\mu}\ell(\mu,p) & =c'(\mu)-h_{0}\frac{\lambda'(p)}{\mu}\left(\mathbb{E}[W_{\infty}(\mu,p)]+\mathbb{E}[X_{\infty}(\mu,p)]+\frac{1}{\mu}\right).
\end{split}
\end{equation}
%Since the stationary waiting time and busy time are unknown, we can obtain a gradient estimator by using the waiting time and busy time encountered by the current customer as estimates of the stationary quantities.
Based on \eqref{eq:gg1_gradient0}, we define 
\[g(\mu,p, Z_t)=(g_p(\mu,p,Z_t), g_{\mu}(\mu,p,Z_t)),\] 
where
\begin{equation}\label{eq:gg1_gradient}
\begin{split}
g_{p}(\mu,p,Z_{t}) & =-\lambda(p)-p\lambda'(p)+h_{0}\lambda'(p)\left(W_{t}+X_{t}+\frac{1}{\mu}\right),\\
g_{\mu}(\mu,p,Z_{t}) & =c'(\mu)-h_{0}\frac{\lambda'(p)}{\mu}\left(W_{t}+X_{t}+\frac{1}{\mu}\right).
\end{split}
\end{equation}
%We consider the stochastic gradient descent algorithm using the gradient estimator defined in (\ref{eq:gg1_gradient}) 
Then, the SGD update for $\theta_{t}=(\mu_{t},p_{t})$
with the gradient estimator \eqref{eq:gg1_gradient} and step-size $\eta_{t}$ takes the form
%\begin{equation}
\[
\theta_{t+1}=\bfP_{\Theta}\left(\theta_{t}-\eta_{t}g(\theta_{t},Z_{t})\right).
\]
%\label{eq:gg1-sgd}
%\end{equation}

To prove the convergence of the SGD update, 
%i.e., verify the conditions in Theorem \ref{thm:main2}, 
we impose some regularity conditions on the functions $\lambda$ and $c$ and the distributions of the baseline interarrival time and service time, $T$ and $S$.
%we make mild assumptions on
%the regularity of $\lambda$ and $c$ on $\Theta$,
\begin{assumption}
\label{assu:function-assumption}For the demand and cost
functions we have:
\begin{itemize}
\item[(i)] The constraint set $\Theta=[\underline{p},\overline{p}]\times[\underline{\mu},\overline{\mu}]$
is such that $\lambda(\underline{p})<\underline{\mu}$.
\item[(ii)] $\lambda(p)\in C^{2}$ on $[\underline{p},\overline{p}]$ and
non-increasing in $p$. 
\item[(iii)] $c(\mu)\in C^{2}$ on $[\underline{\mu},\overline{\mu}]$ and
non-decreasing in $\mu$. 
\end{itemize}
\end{assumption}

% The service provider need not know any information about the interarrival
% times and the service times. For the proof of convergence, we assume
% that $S,T$ are sufficiently light tailed and have $\mathcal{C}^{1}$
% density functions.
\begin{assumption}
\label{assu:dist-assumption} The baseline service time $S$ and inter-arrival
time $T$ are iid respectively and
\begin{itemize}
\item[(i)] There exists $\alpha^{*}>0$ such that $\mathbb{E}[e^{4\alpha^{*}S}]<\infty$
and $\mathbb{E}[e^{4\alpha^{*}T}]<\infty$.
\item[(ii)] There exist $0<\alpha_2<\alpha_{1}<\min\{\alpha^{*}/\underline{\mu},\alpha^{*}/\lambda(\underline{p})\}$  such that
\[
\mathbb{E}\left[e^{4\alpha_{1}\frac{T}{\underline{\mu}}}\right]\mathbb{E}\left[e^{-4(\alpha_{1}-\alpha_{2})\frac{S}{\lambda(\underline{p})}}\right]<1.
\]
\item[(iii)] The density functions of $T$ and $S$, which are denoted as $f_T$ and $f_S$ respectively, are continuously differentiable. In addition, there exist $c,D_{1},D_{2}>0$
and $k\in\mathbb{N}_{+}$ such that for all $x\geq \frac{c}{\min\{\underline{\mu}, \underline{\lambda}\}}$, 
$\left|\frac{d}{dx}\log f_T(x)\right|\leq D_{1}+D_{2}|x|^{k}$
and 
$\left|\frac{d}{dx}\log f_S(x)\right|\leq D_{1}+D_{2}|x|^{k}$. Lastly, $f_S(\mu x) \leq C (f_S(\underline{\mu}x)+f_S(\bar{\mu}x))$ for $\mu\in[\underline{\mu},\bar{\mu}]$
and
$f_T(\lambda x) \leq C(f_T(\underline{\lambda}x)+f_T(\bar{\lambda}x))$ for $\lambda\in[\underline{\lambda},\bar{\lambda}]$.
\end{itemize}
\end{assumption}
Assumption \ref{assu:dist-assumption} requires $S$ and $T$ to be sufficiently light-tailed and their density functions are smooth enough. Commonly used service and interarrival time distributions, such as exponential, Erlang, Weibull$(\lambda, k)$ with $k\geq 1$, etc. satisfy Assumption \ref{assu:dist-assumption}. For example, the Weibull$(1,k)$, $k\geq 1$, satisfies that for $x\geq 1$:
\[
\frac{d}{dx}\log f(x)=\frac{d}{dx}\left[ k\log x - x^{k} \right] \leq k + k|x|^{k-1}.
\]
In addition, for $\mu\in[\underline{\mu},\bar{\mu}]$,
\[
f(\mu x)\leq \left(\frac{\bar{\mu}}{\underline{\mu}}\right)^{k-1}(f_T(\underline{\lambda}x)+f_T(\bar{\lambda}x)).
\]

Under Assumptions \ref{assu:function-assumption} and \ref{assu:dist-assumption}, we can verify the conditions of Theorem \ref{thm:main2} (Assumptions \ref{ass:ergodicity0} -- \ref{ass:error0} restricted to $\theta\in \Theta$), which leads to the following theorem.

\begin{theorem}
\label{thm:gg1-convex}Suppose Assumptions \ref{assu:function-assumption}
and \ref{assu:dist-assumption} hold. Suppose the objective
(\ref{eq:gg1_objective}) is convex and let $(\mu^{*},p^{*})$ be
the global minimum. Then, using the gradient estimator $g(\mu,p,z_{t})$
defined in (\ref{eq:gg1_gradient}) with step-size $\eta_{t}=\eta_{0}t^{-1/2}$,
the average SGD iterate $\bar{\theta}_{T}=(\bar{\mu}_{T},\bar{p}_{T})$ satisfies
\[
\mathbb{E}[\ell(\bar{\mu}_{T},\bar{p}_{T})]-\ell(\mu^{*},p^{*})=O\left(\tau^2(\log T)^{4}/\sqrt{T}\right),
\]
where $\tau$ is an upper bound of the mixing time of the augmented Markov chain $Z_t(\theta)$ for $\theta\in\Theta$.\\
Moreover, if the objective function is strongly convex with a convexity constant $c$, by setting the step size as
$\eta_{t}=2\eta_0/(ct)$ for $t\geq 1$ with $\eta_0>2$, we have
\[\E\|\theta_{T}-\theta^*\|^2=O\left(\tau(\log T)^2/T\right).\] 
%then the iterate $\theta_{T}$
%of (\ref{eq:gg1-sgd}) after observing $T$ customers satisfies 
% \[
% ||\theta_{T}-\theta^{*}||=O\left(\frac{\tau(\log T)^{2}}{T}\right)
% \]
\end{theorem}

% \begin{proof}
% See Appendix \ref{sec:proof-cor-gg1}.
% \end{proof}

The online learning algorithm proposed by \cite{chen2023online} achieves a regret that is logarithmic in the number of customers assuming $\ell$ is strongly convex, and Theorem \ref{thm:gg1-convex} yields a similar regret. However, their algorithm requires calibrating the number of customers seen before making a gradient update. The algorithm we consider does not require such calibration, i.e., we can update the parameter after each arrival. For this specific example, our convergence result is similar to the one developed in \citep{li2023stochastic}. However, the development in \citep{li2023stochastic} requires verifying the Lipchitz continuity of the corresponding Poisson equation solution, which is more involved.

%\subsection{Numerical Experiments}

\section{Application to policy gradient in reinforcement learning}
\label{sec:RL}
%\blue{[Is there any sample complexity rate for TD learning?]}
We consider the classic Markov decision process (MDP) with a finite state space $\mathcal{S}$, a finite action space $\mathcal{A}$, a collection of transition probabilities 
$\{P(\cdot|s,a)\}_{(s,a)\in\mathcal{S}\times\mathcal{A}}$, and an initial distribution $\rho(\cdot)$.
The policy is parameterized by $\theta$ where $\pi^{\theta}(a|s)$ denote the probability of taking action $a$ when in state $s$.
We focus on the infinite-horizon discounted cost formulation with the discount factor $\gamma\in (0,1)$:
\begin{equation}\label{eq:rl}
\min_{\theta} \ell(\theta):=\E_\rho^\theta\left[\sum_{t=0}^\infty \gamma^t c(s_t,a_t)\right],
\end{equation}
where $c(s,a)$ is the expected instantaneous cost incurred by taking action $a$ at state $s$. %and $\rho$ is the initialization measure. 

When applying policy gradient to solve \eqref{eq:rl}, the gradient can be expressed as \citep{sutton2018reinforcement}
\begin{equation}\label{eq:policy_gradient}
\nabla \ell(\theta)=\sum_{s,a}\nu^\theta(s,a) Q^\theta(s,a)\nabla_\theta\log \pi^\theta(a|s)
\end{equation}
where $Q^\theta$ is the state-action value function under policy $\pi^{\theta}$,
\[
\nu^{\theta}(s,a)=\frac{1}{1-\gamma}\sum_{t=0}^\infty\E_{\rho}^\theta [\gamma^t 1_{s_{t}=s,a_{t}=a}]
\]
is the state-action occupancy measure, which can also be viewed as the stationary distribution of the Markov chain with transition kernel
\[
\mathcal{P}^{\theta}((s,a), (s',a')):=(1-\gamma) \rho(s')\pi^{\theta}(a'|s') + \gamma P(s'|s, a)\pi^{\theta}(a'|s').
\]
In general, $\ell(\theta)$ is a non-convex function of $\theta$. Recent works have shown that under certain regularity conditions, any stationary point of the policy gradient loss function is globally optimal \citep{agarwal2021theory, bhandari2024global,wang2019neural}. For instance, for finite state and action MDPs with natural parameterization, \cite{bhandari2024global} show that $\ell(\theta)$ has no suboptimal stationary point. In addition, \citep{agarwal2020optimality} show that $\nabla \ell(\theta)$ is Lipschitz continuous. 

From the gradient formula \eqref{eq:policy_gradient}, we can take $Q^{\theta}(s,a)\nabla_{\theta}\log\pi^{\theta}(a|s)$ as a gradient estimator. However, if $(s,a)$'s are not sampled from $\nu^{\theta}(s,a)$, the gradient estimator is biased. In addition, $Q^{\theta}(s,a)$ also needs to be estimated. One way to overcome the challenge is to simulate the Markov chain under policy $\pi^{\theta}$ for a long enough time so that we get an accurate enough estimate $Q^{\theta}$ and the distribution of $(s_t,a_t)$ is close enough to $\nu^{\theta}(s,a)$. This idea has been employed in the literature (see, e.g., \cite{wang2019neural, xu2020improving, xiong2021non}). In this section, we are interested in understanding how adaptive the policy gradient algorithm can be while still achieving fast convergence to a stationary point. 
%This idea has been employed in \cite{wang2019neural} where they require $O(T^8)$ steps in the TD update to estimate $Q$ function. Here, $T$ is the number of policy improvement steps.

We consider an actor-critic scheme where we update the state-action value function using the following temporal-difference (TD) update: 
\begin{equation}
\label{eqn:QMC}
\begin{split}
&Q_{t+1}(s_t,\hat{a}_t)=Q_t(s_t,\hat{a}_t)+\alpha[c(s_t,\hat{a}_t)-Q_t(s_t,\hat{a}_t)+\gamma  Q_t(s'_{t+1},a'_{t+1})]\\
&Q_{t+1}(s,a)=Q_t(s,a) \mbox{ for $(s,a)\neq (s_t,\hat{a}_t)$,}
\end{split}
\end{equation}
where $\hat{a}_t$ is sampled uniformly at random from $\mathcal{A}$, and $(s'_{t+1},a'_{t+1})$ is a random state generated from $\mathcal{P}^{\theta_t}((s_t,\hat{a}_{t}), \cdot)$, independent of $(s_{t+1},a_{t+1})$, which is generated from $\mathcal{P}^{\theta_t}((s_t,a_t), \cdot)$. Denote $Z_t=(s_t, a_t, Q_t)$. 

We first establish the ergodicity property of the Markov chain $Z_t=(s_t,a_t, Q_t)$ under a fixed policy $\pi^{\theta}$. In particular, given $Z_t$, $Z_{t+1}$ is generated as follows: Sample $\hat{a}_t$ uniformly at random from $\mathcal{A}$. Sample $(s_{t+1}, a_{t+1})$ from $\mathcal{P}^{\theta_t}((s_t,a_t), \cdot)$, and independently, sample $(s_{t+1}', a_{t+1}')$ from $\mathcal{P}^{\theta_t}((s_t,\hat{a}_t), \cdot)$. Update $Q_{t+1}$ according to \eqref{eqn:QMC}. 

For a Markov chain $s_t$ with a finite state space $\mathcal{S}$ and a unique stationary distribution $\nu$.
Let $\kappa_s=\inf\{t\geq 0: s_t=s\}$.
We define the hitting time \citep{levin2017markov} as
\[
t_{hit}=\max_{s,s'\in \mathcal{S}}\E_s[\kappa_{s'}].
\]
% Let $\tau_{cov}$ be the first time at which $s_t$ have visited all the states. Then we define define the cover time \citep{levin2017markov} as
% \[
% t_{cov}=\max_{s\in\mathcal{S}}\E_x[\tau_{cov}].
% \] 
% %define the covering time as in  \citep{levin2017markov}
% %\[
% %t_{cov}:=\max_{s\in \mathcal{S}}\E_s\left[\max_{s'\in \mathcal{X}}\kappa_{s'}\right].
% %\]
% Note that (see Theorem 11.2 in \citep{levin2017markov})
% \[
% t_{cov} \leq t_{hit}\sum_{k=1}^{|\mathcal{S}|-1}\frac{1}{k}.
% \]
In addition, recall that under the total variation distance $\|\cdot\|_{\text{TV}}$, the mixing time of $s_t$, $t_{mix}$, satisfies
\[
\max_{s\in \mathcal{S}}\|P_s^{t_{mix}}-\nu\|_{\text{TV}}\leq \frac14.
\]
% We also define $\eta_{\epsilon}$ as an $\epsilon$-mixing time if 
% \[
% \max_{x\in \mathcal{X}}\|P_x^{\eta_{\epsilon}}-\nu\|_{\text{TV}}\leq \epsilon.
% \]
% \begin{itemize}
%     \item Total-variation mixing time $t_{mix}=\arg\min_{t\geq0} \{\max_{x\in \mathcal{X}}\|P_x^t-\nu\|_{tv}\leq \frac14\}$
%     \item Hitting time: $t_{hit}=\max_{x,x'\in \mathcal{X}}\E_x[\tau_{x'}]$
%     \item Covering time: $t_{cov}=\max_{x\in \mathcal{X}}\E_x[\max_{x'\in \mathcal{X}}\tau_{x'}]$
% \end{itemize}
% %Some understanding based on Pere's book, Theorem 10.22, $t_{mix}\leq 2 t_{hit}+1$. Theorem 11.2 says $t_{cov}\leq \log(|\mathcal{X}|)t_{hit}$. 
% %{\color{blue} Need to double check.}
% Among these stopping times, we have
% $t_{mix}\leq 2 t_{hit}+1$ (Theorem 10.22 in \cite{levin2017markov})  {\color{red} requires reversibility}, and
%$t_{cov}\leq \log(|\mathcal{X}|)t_{hit}$ (Theorem 11.2. in \cite{levin2017markov})
% We also define the $\epsilon$ mixing time as
% \[
% \eta_{\epsilon} := =\arg\min_{t\geq0} \{\max_{x\in \mathcal{X}}\|P_x^t-\nu\|_{tv}\leq \epsilon\}
% \]
%Let $M=\sup_{t} \max_{s\in \mathcal{S}, a\in \mathcal{A}}Q_t(s,a)$. We assume $M<\infty$.
% \red{I added a Lemma \ref{lem:tcov} to use $t_{cov}$. That will make a better packaging? If so, replace $t'_{cov}$ with $t_{cov}$ in below. }
\begin{proposition}
\label{prop:policy_gradient_ergo}
For a fixed value of $\theta$ (i.e., under a fixed policy $\pi_{\theta}$), suppose $s_t$ is a finite-state Markov chain with a mixing time $t_{mix}$ under the total variation distance and a finite hitting time $t_{hit}$. In addition, suppose $Q_t(s,a) \leq M$ for some $M<\infty$.
%When $\theta$ is fixed, 
%and $(s_t,a_t)$ has an $\epsilon$-mixing time $\eta_{\epsilon}$ under the total variation distance. 
Then, the triad $Z_t=(s_t,a_t,Q_t)$ induced by the TD sampler is a Markov chain with an $\epsilon$-mixing time, $\bar \eta_\epsilon$, satisfying
\[\begin{split}
\bar \eta_\epsilon \leq %\frac{|\log \epsilon|}{|\log (1/4)|} 
&\frac{|\log (\epsilon/(8M+4))|}{|\log (1/4)|}t_{mix}\\
&+\max\left\{\frac{|\log (\epsilon/(4M))|}{|\log (1-\alpha+\alpha\gamma)|}, 12|\log(\epsilon/(8M+4))|\right\}(1+t_{hit})|\mathcal{A}|\sum_{k=1}^{|\mathcal{S}||\mathcal{A}|-1}\frac{1}{k}
\end{split}\]
under the distance 
\[
\tilde d(Z,\tilde Z)=1_{s=\tilde s,a=\tilde a}+\|Q-\tilde Q\|_\infty. 
\]
\end{proposition}

\begin{proposition} \label{prop:policy_graident_stationary}
%For the TD sampler with a fixed value of $\theta$ (i.e., under a fixed policy $\pi^{\theta}$), suppose $(s_t,a_t)$ is a finite-state Markov chain with a mixing time $t_{mix}$ under the total variation distance, and $Q_t(s,a) \leq M$ for some $M<\infty$.
Under the assumptions of Proposition \ref{prop:policy_gradient_ergo},
for $\alpha$ in the TD sampler takes value outside a finite set of values, the invariant distribution of the triad $Z_t=(s_t,a_t,Q_t)$, $\mu^{\theta}$, satisfies 
\[
\E_{\mu^\theta}[1_{s_t=s,a_t=a}]=\nu^\theta(s,a),
~~ \E_{\mu^\theta}[Q_t(s_t,a_t)|s_t=s,a_t=a]=Q^\theta(a,s).
\]
%which implies 
%\[
%\E_{\mu^\theta}[Q_t(s_t,a_t) \nabla_\theta \log\pi^\theta(a_t|s_t)]
%=\sum_{s\in\mathcal{S}, a\in\mathcal{A}} \nu^\theta(s,a) Q^\theta(s,a)\nabla_{\theta} \log\pi^\theta(a|s).
%\]
\end{proposition}

We note from Propostion \ref{prop:policy_graident_stationary} that $Q_{t+1}$ does not converge to $Q^\theta$ even if $(s_t,a_t)$'s are generated under $\pi^\theta$. Instead, $(s_t,a_t, Q_t)$ will be a Markov chain with the stationary measure $\mu^{\theta}$ satisfying
\[
\E_{\mu^{\theta}}[Q_t(s_t,a_t)|s_t=s,a_t=a]=Q^\theta(s,a).
\]

%We will show later in Propositions \ref{prop:policy_gradient_ergo} and \ref{prop:policy_graident_stationary} that  $Z_t$ generated under a fixed policy $\pi^{\theta}$ is a properly defined ergodic Markov chain with stationary distribution $\mu^{\theta}$, where $\mu^{\theta}$ satisfies
%\[
%\E_{\mu^\theta}[Q_t(s_t,a_t) \nabla_\theta\log\pi^\theta(a_t|s_t)]
%=\sum_{s\in\mathcal{S}, a\in\mathcal{A}} \nu^\theta(s,a) Q^\theta(s,a)\nabla_{\theta} \log\pi^\theta(a|s).
%\]
%Thus, based on Theorem \ref{thm:main1}, 
Based on Propositions \ref{prop:policy_gradient_ergo} and \ref{prop:policy_graident_stationary}, $Z_t$ generated under a fixed policy $\pi^{\theta}$ is a properly defined ergodic Markov chain with stationary distribution $\mu^{\theta}$, where $\mu^{\theta}$ satisfies
\[
\E_{\mu^\theta}[Q_t(s_t,a_t) \nabla_\theta\log\pi^\theta(a_t|s_t)]
=\sum_{s\in\mathcal{S}, a\in\mathcal{A}} \nu^\theta(s,a) Q^\theta(s,a)\nabla_{\theta} \log\pi^\theta(a|s).
\]
This indicates that we only need ``one transition" for each policy update. Algorithm \ref{alg:TD} summarizes our actor-critic scheme. %where only two samples, i.e., $(s_{t+1}, a_{t+1})$ and $(s_{t+1}', a_{t+1}')$, are drawn at each iteration.

% \begin{algorithm}
%   Initialize $s_0,a_0,Q_0, \theta_0$. Set $t=0$. \;
%     Sample $(s_{t+1}, a_{t+1})$ and $(s_{t+1}', a_{t+1}')$ as two independent samples from $\mathcal{P}^{\theta_t}((s_t,a_t), \cdot)$.
%     Set
%     \[\begin{split}
%         &Q_{t+1}(s_t,a_t)=Q_t(s_t,a_t)+\alpha[c(s_t,a_t)-Q_t(s_t,a_t)+\gamma  Q_t(s'_{t+1},a'_{t+1})]\\
%         &Q_{t+1}(s,a)=Q_t(s,a) \mbox{ for $(s,a)\neq (s_t,a_t)$,}
%     \end{split}\]
%     and  
%     $\theta_{t+1}=\theta_t - \eta_t Q_{t+1}(s_{t+1},a_{t+1}) \nabla_{\theta} \log\pi^{\theta_t}(a_{t+1}|s_{t+1})$ \;
% Set $t=t+1$. If $t<T$, go back to Step 2; otherwise, output $\theta_T$
%   \caption{Actor-critic based policy gradient} \label{alg:TD}
% \end{algorithm}

\begin{algorithm}
  Initialize $s_0,a_0, Q_{0}, \theta_0$. Set $t=0$. \;
    Sample $\hat{a}_{t}$ uniformly at random from $\mathcal{A}$. Sample $(s_{t+1}, a_{t+1})$ and $(s_{t+1}', a_{t+1}')$ as two independent samples from $\mathcal{P}^{\theta_t}((s_t,a_t), \cdot)$ and $\mathcal{P}^{\theta_t}((s_t,\hat{a}_{t}), \cdot)$ respectively.
    Set
    \[\begin{split}
        &Q_{t+1}(s_t,\hat{a}_t)=Q_t(s_t,\hat{a}_t)+\alpha[c(s_t,\hat{a}_t)-Q_t(s_t,\hat{a}_t)+\gamma  Q_t(s'_{t+1},a'_{t+1})] \\
        &Q_{t+1}(s,a)=Q_t(s,a) \mbox{ for $(s,a)\neq (s_t,\hat{a}_{t})$,}
    \end{split}\]
    and  
    $\theta_{t+1}=\theta_t - \eta_t Q_{t+1}(s_{t+1},a_{t+1}) \nabla_{\theta} \log\pi^{\theta_t}(a_{t+1}|s_{t+1})$ \;
Set $t=t+1$. If $t<T$, go back to Step 2; otherwise, output $\bar \theta_T$
  \caption{Actor-critic based policy gradient} \label{alg:TD}
\end{algorithm}

% {\color{red} Need to discuss how to write this section. Need geometric ergodicity, which I think the discount-induced regeneration will satisfy this. Lipchitz continuity: we probably only need to make assumptions on $\pi^{\theta}$. Boundedness: since we are focusing on finite-state Markov chains, we probably only need some assumptions on $c(s,a)$. In this case, we can apply Theorem 1 to prove convergence to a stationary point.

% Suppose $c(s,a)\in(0,M)$. Then $Q \leq M/(1-\gamma)$. If $Q_0\leq M/(1-\gamma)$, then $Q_t\leq M/(1-\gamma)$ for any $t\geq 0$.}

We next establish the convergence of Algorithm \ref{alg:TD}. We first introduce some assumptions about the MDP. 

\begin{assumption}
\label{ass:pg-bounded}
The instantaneous costs are bounded, i.e., $|c(s,a)|\leq M$, almost surely. $Q_0$ is initialized such that $\|Q_0\|_{\infty} \leq M/(1-\gamma)$.
\end{assumption}

\begin{assumption}
\label{ass:pg-init-dist}
The initial distribution $\rho(s)>0$ for all $s\in\mathcal{S}$.
\end{assumption}

\begin{assumption}
\label{ass:pg-score}
For all $\theta\in\Theta$ and all $a\in\mathcal{A}$ and $s\in\mathcal{S}$, %the gradient of $\log \pi^{\theta}(a|s)$, 
$\nabla_{\theta}\log\pi_{\theta}(a|s)$ is bounded and Lipschitz continuous in $\theta$, i.e. there exists $R,L\in (0,\infty)$ such that for any $\theta,\theta'\in \Theta$,
\[
||\nabla_{\theta}\log\pi_{\theta}(a|s)||\leq R, \quad ||\nabla_{\theta}\log\pi_{\theta}(a|s)-\nabla_{\theta'}\log\pi_{\theta'}(a|s)||\leq L||\theta-\theta'||.
\]
\end{assumption}

Assumption~\ref{ass:pg-bounded} is a mild assumption since the state and action spaces are finite, and this assumption is standard in the literature (see, e.g., Assumption 1 in \cite{mei2020global}).
Assumption~\ref{ass:pg-init-dist} ensures sufficient exploration of the state space and is also a standard assumption for the convergence analysis of vanilla policy gradient (see, e.g., Assumption 2 in  \cite{mei2020global}).
Assumption~\ref{ass:pg-score} is a regularity condition on the score function that prevents the policy gradient estimator from having an infinite variance and is a standard assumption in the analysis of policy gradient under estimated gradients (see, e.g., Assumption 3.1 in  \cite{zhang2020global}). This is naturally satisfied under the softmax parameterization $\pi(a|s) \propto \exp(\theta_{s,a})$  with $\Theta$ being a bounded subset of $ \R^{|\mathcal{S}||\mathcal{A}|}$.

%Under these assumptions, Algorithm~\ref{alg:TD} converges to a stationary point at rate $\tilde O(1/\sqrt{T})$.
\begin{theorem}
\label{thm:pg-convergence} Suppose Assumptions~\ref{ass:pg-bounded}-\ref{ass:pg-score} hold.
Consider step size $\eta_{t} = \eta_{0}/\sqrt{t}$ and let $\theta_{t}$ denote the policy parameters under Algorithm~\ref{alg:TD}. Then
\[
\min_{0\leq t< T}\E \| \nabla \ell(\theta_t)\|^2=O\left(\tau (\log T)^2/\sqrt{T}\right).
\]
where $\tau$ is the mixing time of $Z_{t} = (s_{t},a_{t},Q_{t})$ and is $O\left( \frac{|\mathcal{A}|\min_{s\in\mathcal{S}} \rho(s)^{-1}}{1-\gamma} \right)$.
\end{theorem}

Theorem \ref{thm:pg-convergence} indicates that Algorithm~\ref{alg:TD} is able to attain an $\epsilon$-stationary point with $O(\epsilon^{-2}\log(1/\epsilon))$ samples. This order bound also depends polynomially on $|S|,|A|$, and $\frac{1}{1-\gamma}$, and we omit the full dependence on these terms here (more details can be found in the proof of Theorem \ref{thm:pg-convergence} in Appendix \ref{sec:proof-pg-convergence}).

% {\color{green}
% The full bound is:
% \[
% \min_{0\leq t< T}\E \| \nabla \ell(\theta_t)\|^2=O\left(\left(\frac{1}{1-\gamma} \right)^{15} |A|^{7}|S|^{7}(\log T)^2/\sqrt{T}\right).
% \]
% }

Let $T$ denote the total number of iterations. For a similar $\tilde O(1/\sqrt{T})$ convergence, \cite{wang2019neural} requires running $O(T^8)$ steps in the TD update to estimate the Q-function in each iteration and the ability to sample from $(s,a)$ from $v^{\theta}(s,a)$ directly. In particular, for a policy $\pi^\theta$, they approximate the gradient using
\[
\frac{1}{B}\sum_{k=1}^{B}\hat Q^{\theta}(\tilde s_k, \tilde a_k)
\]
where $\hat Q^{\theta}$ is estimated by running TD under $\pi^{\theta}$ for $O(T^8)$ steps and $(\tilde s_k, \tilde a_k)$, $k=1,\dots, B$ are $B$ iid samples drawn from $v^{\theta}$. When using the REINFORCE gradient estimator, \cite{yuan2022general} requires running the Markov chain under policy $\pi^{\theta}$ for $O(\log(1/T))$ steps in each iteration and the overall sample complexity is $\tilde O(\epsilon^{-4})$.
Meanwhile, we would also like to acknowledge that \cite{wang2019neural, yuan2022general} study more complicated settings than the tabular setting we study here. For example, \cite{wang2019neural} considers using neural networks to approximate the policies and the state-action value functions. \cite{yuan2022general} considers different and more general regularity conditions than what we assume in this section. Our sample complexity results are comparable to some of the best-known sample complexity results for policy gradient algorithms. In particular, \cite{xiong2021non} establishes an $\tilde O(\epsilon^{-2})$ complexity for an Adam-type policy gradient algorithm, which requires sampling a long enough trajectory under a policy $\pi$ to estimate the corresponding Q-function. 
\cite{xu2020improving} establishes an $\tilde O(\epsilon^{-2})$ complexity for a mini-batch actor-critic policy gradient algorithm. Unlike these works, we do not assume that the Markov chain on the states induced by the policy is uniformly ergodic across policies, which is a strong assumption that is difficult to verify. Rather, we use the regenerations induced by the discount factor $\gamma$ to control the mixing rate, which allows our analysis to be applied to a wider range of problems.

\section{Numerical Experiments} \label{sec:num}

Motivated by our theoretical results, we proceed to empirically study the performance of the SGD algorithm with adaptive data. Specifically, for the policy optimization examples considered in Sections \ref{sec:inventory} - \ref{sec:RL}, we examine how the adaptivity of the algorithm
%/non-stationary of the data 
affects performance by varying the batch size: the number of data points collected before updating the policy parameters. 

We observe broadly that even in the fully adaptive setting where policy parameters are updated after every data point, i.e., the batch size is $1$, SGD can achieve an equivalent rate of convergence to larger-batch variants, which is consistent with our theoretical results. This holds even for highly non-stationary environments where the policy parameters change the environment quite a bit and convergence to stationarity is slow. Nevertheless, a carefully chosen larger batch size can provide small improvements in convergence speed in some cases (likely by improving the constant term), especially if the step sizes are tuned appropriately. Above all, our results show that under the ergodicity and smoothness conditions characterized in our theoretical analysis, in non-stationary, adaptive environments, SGD is robust to the level of adaptivity and the convergence speed is largely similar to the iid setting.

% In addition, we also observe a stark difference between the adaptive data setting and the iid setting: while it is known that for iid data, SGD empirically converges well (to a neighborhood of the optional) even with constant step sizes or step sizes larger than the theoretically recommended $1/t$ schedule, in the adaptive setting, small step sizes as elaborated in our theoretical analysis is important for convergence. {\color{red} Add some discussion/explanation. Design some experiments.}
%in the joint pricing and capacity sizing problem for the $M/M/1$ queue instabilities emerge even when increasing the step-size schedule from $\eta_{t} = 1/t$ to $\eta_{t} = 1/t^{1/3}$. This suggests that setting small step sizes as elaborated in our theoretical analysis is much more important in the adaptive data setting.

For the following numerical results, we compare the performance of batch sizes $B \in \{1, 10, 100 \}$. We index the number of data points by $t\in\mathbb{N}_{+}$. For each example, we test out a step-size schedule $\eta_{t} \propto t^{-1/2}$ with iterate averaging $\bar{\theta}_{t}$ and average the performance across 100 independent runs of the algorithm. For the queuing and inventory examples, we also look at the loss of $\theta_{t}$ under the step-size schedule $\eta_{t} \propto t^{-1}$.
We plot the performance as a function of the number of data samples rather than SGD iterations. In this setting, running the algorithm with a larger batch size will have identical $\theta_t$ across multiple data points until an update of the policy is made.

% We utilize a learning rate schedule $\eta_{k} = \eta_{0}(k/B)^{-1}$, where $k$ is the $k$-th SGD update. The step sizes are chosen so that larger step sizes are used for larger batch sizes. {\color{red} For the policy gradient example, since we don't have convexity, we probably need to use $1/\sqrt{t}$ step sizes?}
%while fixing $\eta_{0}$ across all batch sizes. 
% When evaluating performance, we consider the averaged iterate $\bar{\theta}_{t}$ and average the performance across $100$ independent runs of the algorithm. {\color{red} For the strongly convex case, we don't need to average $\theta$'s.} 

\subsection{Inventory Control with Stock-Out Damping}

We empirically test the performance of the adaptive SGD algorithm for the inventory control problem discussed in Section~\ref{sec:inventory}. Recall that in this problem, demand is endogenously affected by the base-stock level $\theta$. We consider a setting where the newsvendor loss is parameterized by an overage cost of $h = 1$, an underage cost of $b = 10$ and the demand process has a demand drift term $m = 5$ and a noise level $\sigma = 1.0$. We consider two values of the $AR(1)$ parameter, $\alpha = 0.8$ and $\alpha = 0.9$, to compare performance across relatively low and high levels of non-stationarity/mixing rate. 
Note that $\alpha$ also calibrates the degree to which $\theta$ affects the dynamics, with $\alpha = 0$ involving zero damping of the demand from stockouts.

\begin{figure}[ht]
\centering
\includegraphics[width=7.3cm]{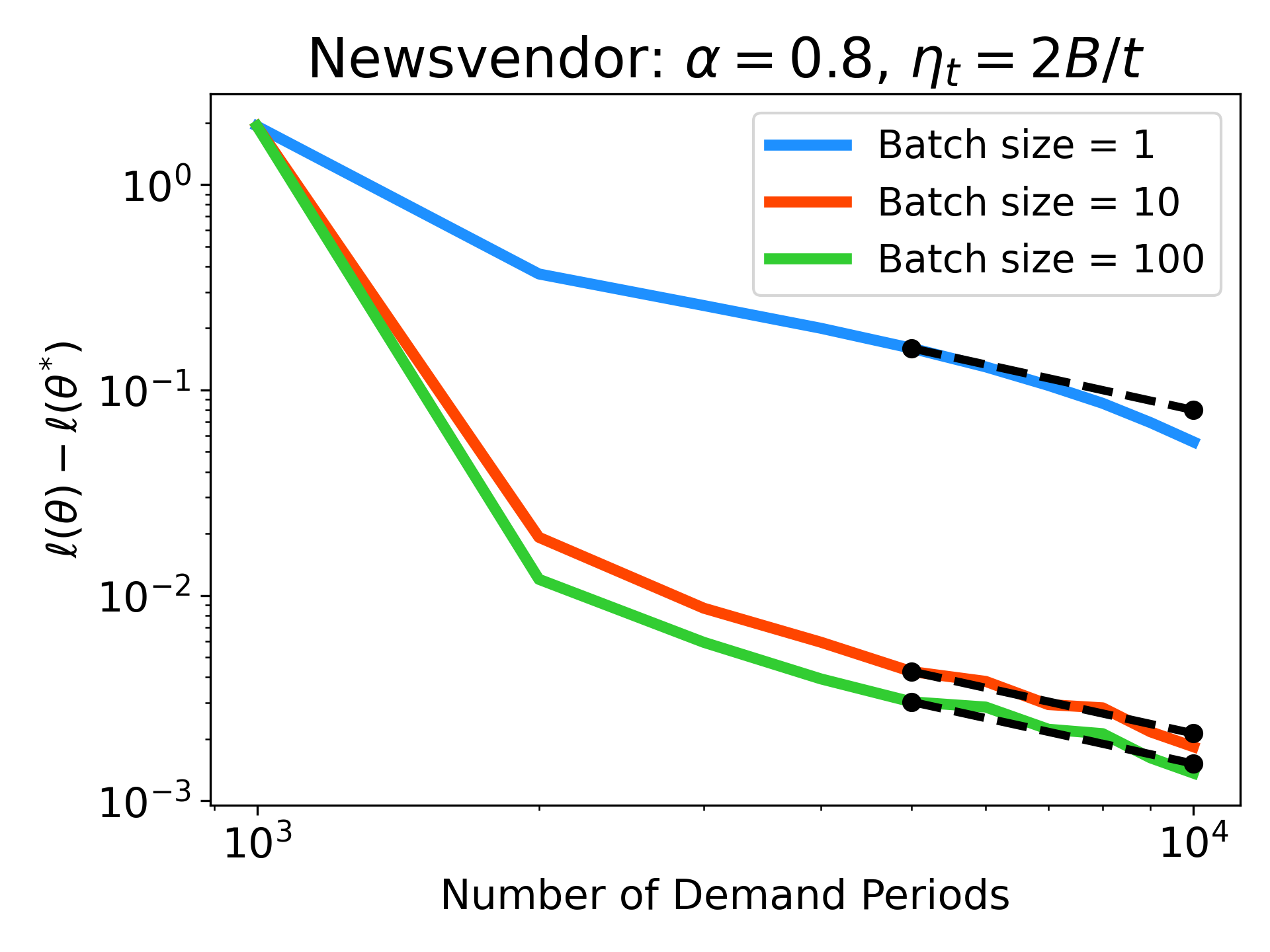}
\includegraphics[width=7.3cm]{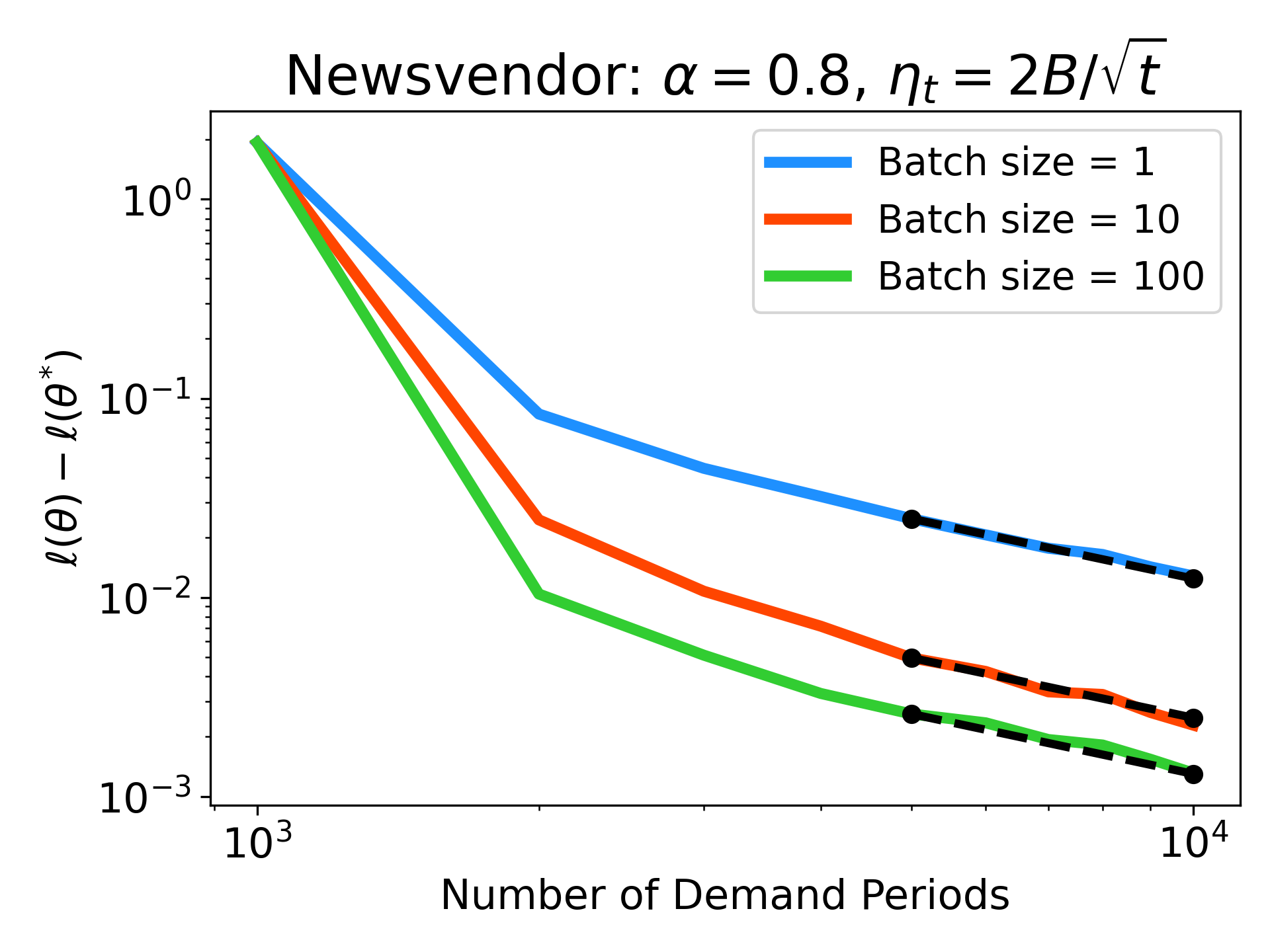}
\includegraphics[width=7.3cm]{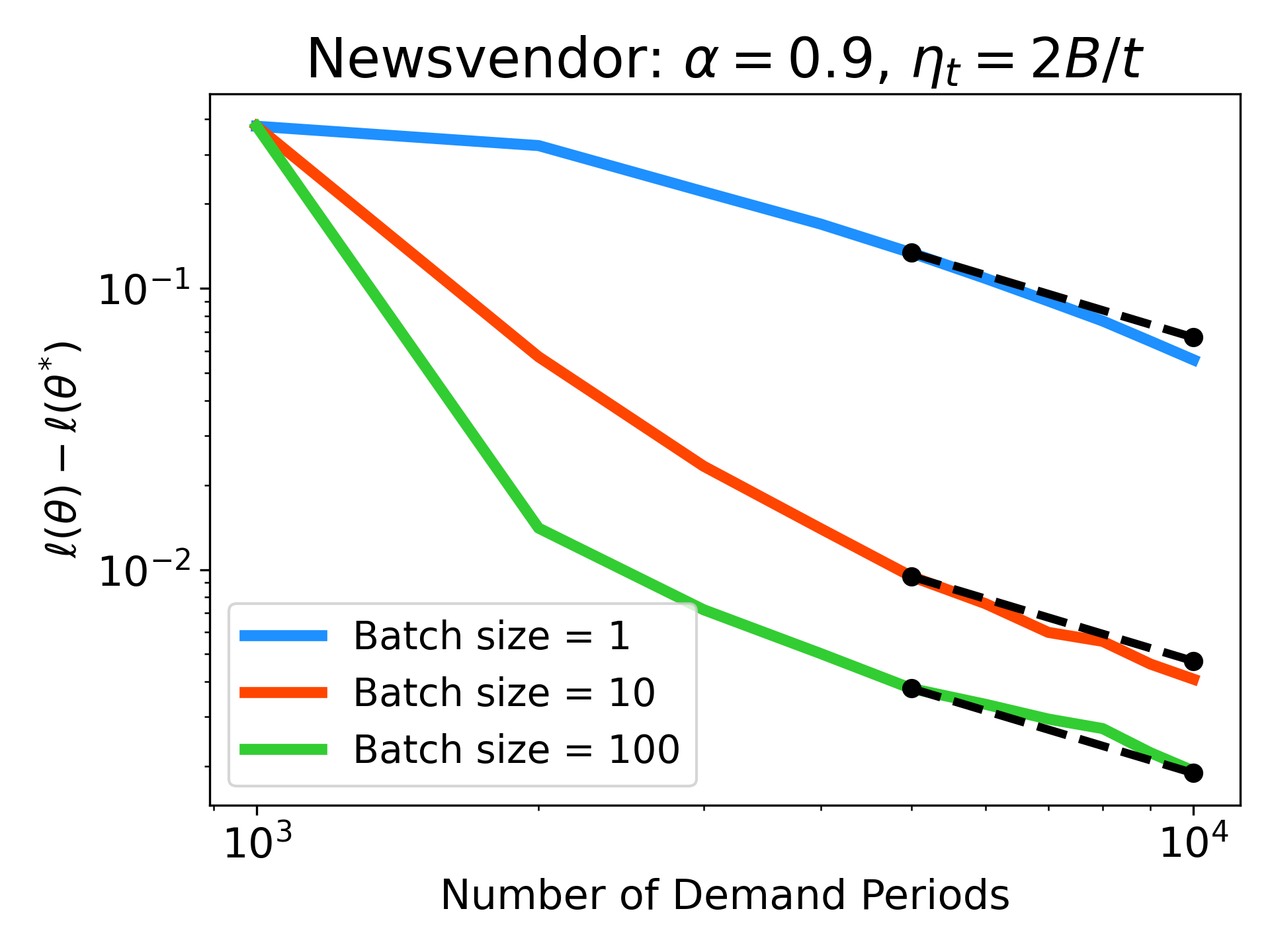}
\includegraphics[width=7.3cm]{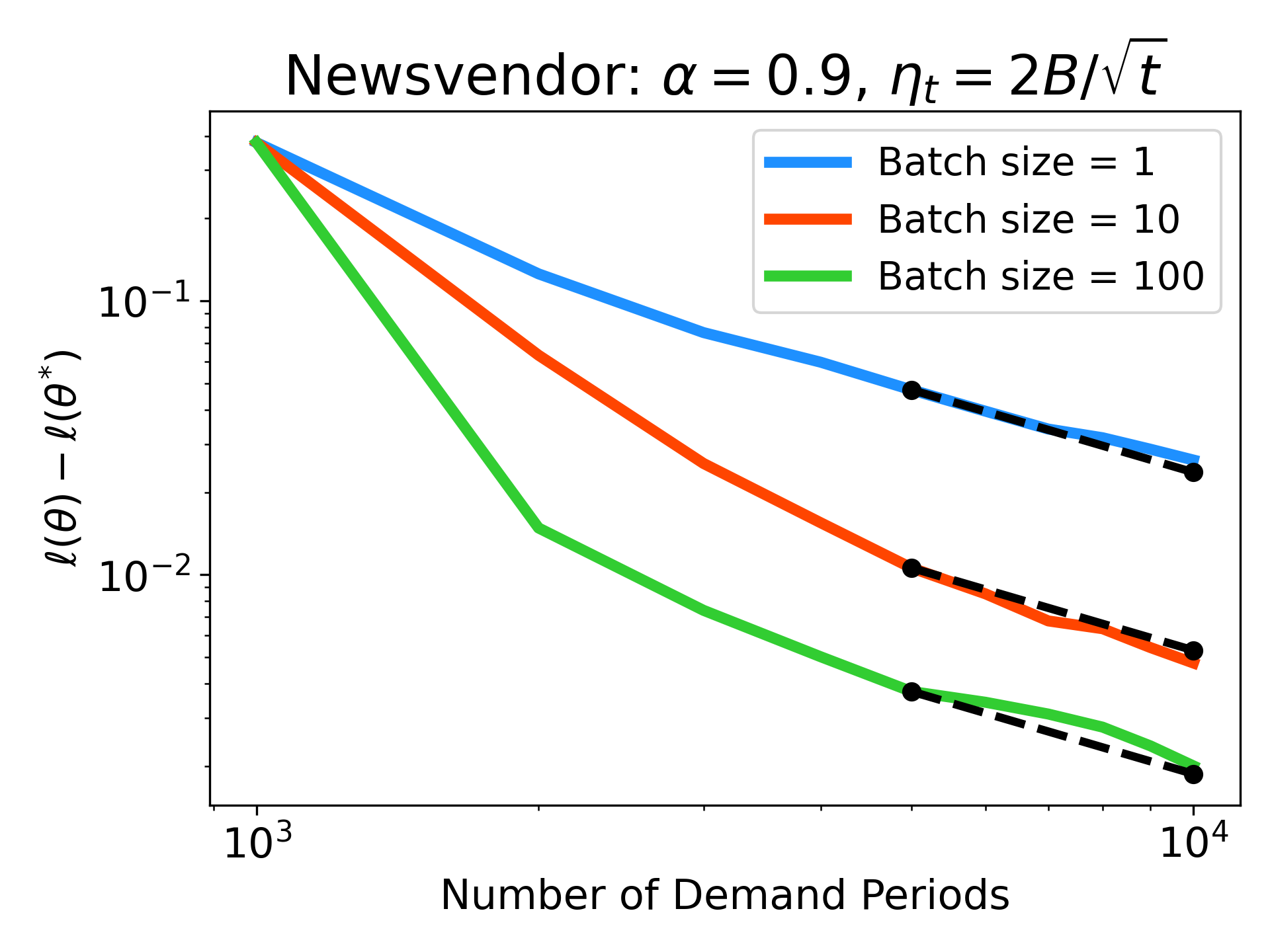}
\caption{Inventory control with stock-out damping (overage cost $b=10$, underage cost $h=1$, and noise level $\sigma = 1.0$). The dotted line indicates $1/t$ convergence rate in all plots. (Top) Newsvendor loss gap for the SGD iterates with $AR(1)$ parameter $\alpha = 0.8$. (Top Left)  Step-size schedule $\eta_{t} = 2Bt^{-1}$ across batch sizes $B \in \{1, 10, 100\}$. (Top Right) Step size schedule $\eta_{t} = 2Bt^{-1/2}$ with iterate averaging. (Bottom) Newsvendor loss gap for the SGD iterates with $AR(1)$ parameter $\alpha = 0.9$. (Bottom Left) Step-size schedule $\eta_{t} = 2Bt^{-1}$. (Bottom Right) Step-size schedule $\eta_{t} = 2Bt^{-1/2}$ with iterate averaging.}  
\label{fig:inventory}
\end{figure}

We compare the SGD algorithm with step sizes scaled by the batch size $B$ across different batch sizes
for the two settings of $\alpha$ in Figure \ref{fig:inventory}. We compare two versions of the SGD algorithm, one with step-size schedule $\eta_{t} \propto t^{-1/2}$ and with iterate averaging; the other with $\eta_{t} \propto t^{-1}$ and without iterate averaging. We find that for both $\alpha = 0.8$ and $\alpha = 0.9$, the fully adaptive SGD algorithm ($B = 1$) achieves an equivalent rate of convergence as larger batch variants, with the large batch sizes performing better (as the step sizes are scaled with the batch size). Interestingly, both the $t^{-1/2}$ and $t^{-1}$ step-sizes empirically achieve a convergence rate of $1/t$ 
%with the number of data points
(indicated by the dotted black line). In addition, while we observe that convergence of SGD is slower for the larger value of $\alpha = 0.9$, because the underlying Markov chain mixes slower in this case and the base-stock policy has a greater impact on the dynamics, the overall rate of convergence resembles the rate observed for $\alpha = 0.8$. This provides numerical evidence that when the conditions of Theorem~\ref{thm:main1} are satisfied, the performance of SGD in the adaptive environment resembles that in the iid setting.

\subsection{Pricing and Capacity Sizing in Single-Server Queue}

We consider the pricing and capacity sizing problem studied in Section~\ref{sec:queue}. Following the numerical example described in \cite{chen2023online}, we consider an $M/M/1$ queue where arrivals to the queue follow a Poisson process with rate $\lambda(p) = n \lambda_{0}(p)$ with $n>0$ and 
\[
\lambda_{0}(p) = \frac{\exp(a - p)}{1 + \exp(a - p)}
\]
for some $a >0$. The service time is exponentially distributed and the service rate $\mu$ entails a quadratic cost $c(\mu)=c_{0}\mu^{2}$ with $c_{0}>0$. There is a holding cost $h_{0} > 0$. In this simple example, the pricing and capacity sizing problem \eqref{eq:gg1_orig_objective} can be written in closed form as
\begin{equation}
\label{eq:mm1_pricing_service}
\max_{p,\mu \in \Theta} \left\{ np\lambda_{0}(p) - c_{0}\mu^{2} - h_0\frac{\lambda(p)/\mu}{1 - \lambda(p)/\mu} \right\}.
\end{equation}
We set $n = 10$, $a = 4.1$, $c_{0} = 0.1$, $h_{0} =1$, and the step-size parameter $\eta_{0} = 1$. 

\begin{figure}[ht]
\centering
    \includegraphics[height=6cm]{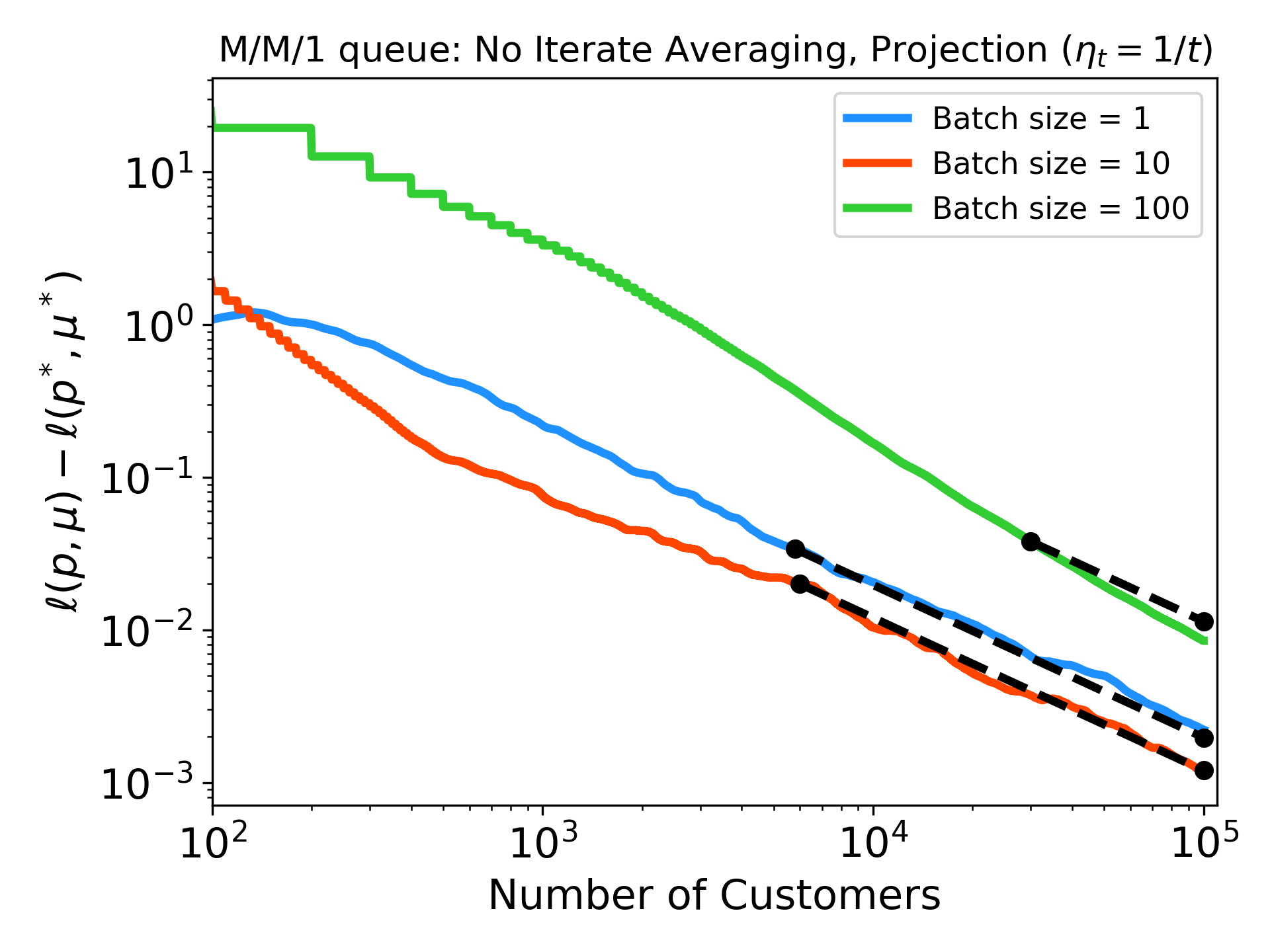}
    \includegraphics[height=6cm]{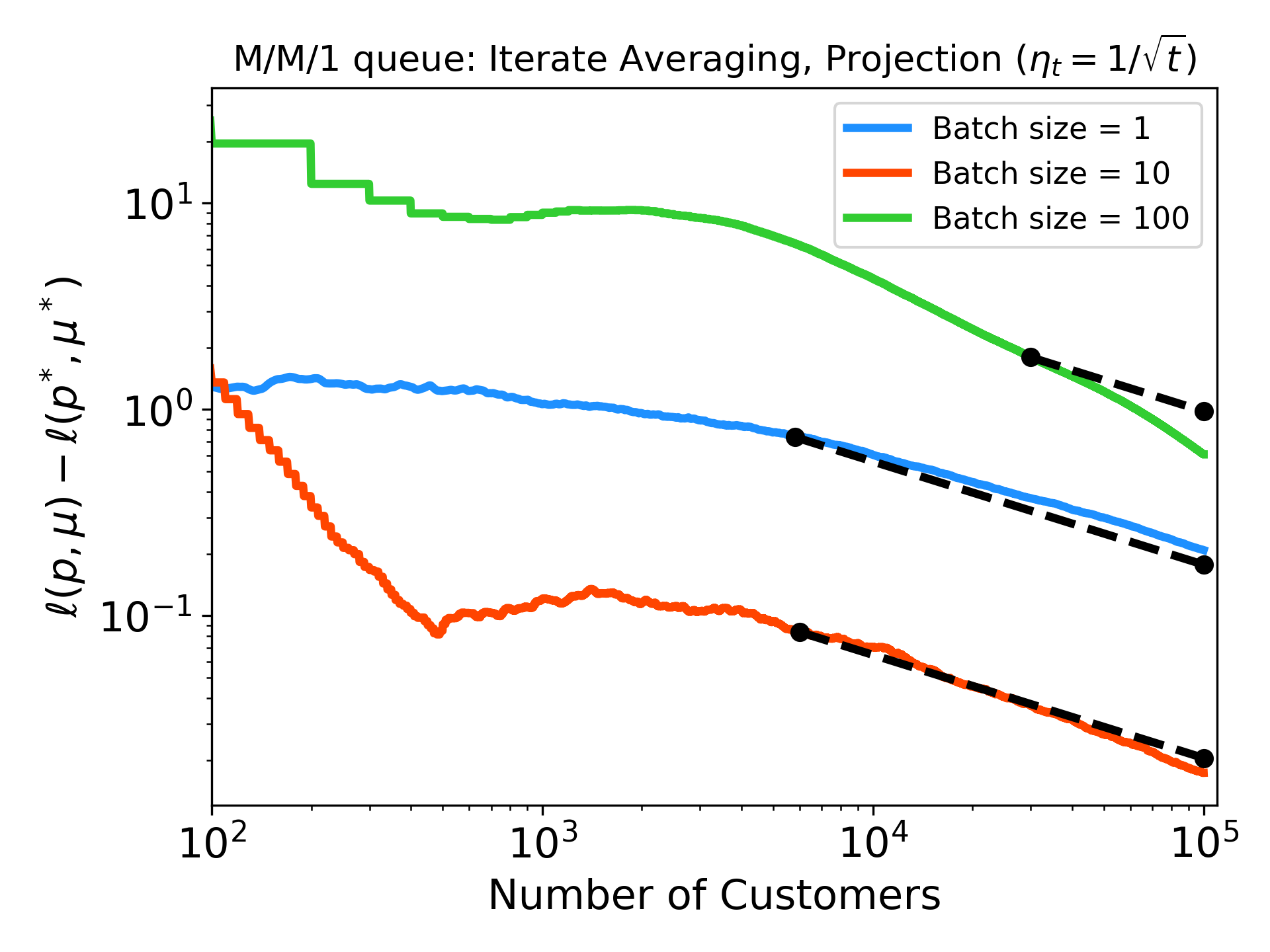}
    \caption{Pricing and capacity sizing in the single server queue.  (Left) Last iterate loss gap for the SGD iterates with $\eta_{t} = 1/t$ across batch sizes $B \in \{1, 10, 100\}$. The dotted line displays the convergence rate of $t^{-1}$. (Right) Loss gap for the average iterate with $\eta_{t} = 1/\sqrt{t}$. The dotted line displays the convergence rate of $t^{-1/2}$. }
    \label{fig:queuing}
\end{figure}

The joint pricing and capacity sizing problem~\eqref{eq:mm1_pricing_service} is known to be strongly convex. As in the inventory example, we evaluate two versions of the SGD algorithm, one with step-size schedule $\eta_{t} \propto t^{-1/2}$ and with iterate averaging; the other with step-size schedule $\eta_{t} \propto t^{-1}$ and 
without iterate averaging. Figure~\ref{fig:queuing} displays the performance for both versions of the algorithm. As predicted by our theoretical results, across all batch sizes, step-size schedule $\eta_{t} = 1/\sqrt{t}$ with iterate averaging achieves an $O(t^{-1/2})$ convergence. Whereas step-size schedule $\eta_{t} = 1/t$ without iterate averaging achieves an $O(t^{-1})$ convergence. %Despite the fact that the queueing network exhibits slow mixing and even possible instability in parts of the parameter space, convergence appears similar to the i.i.d. setting.

\begin{figure}[ht]
\centering
    \includegraphics[height=6cm]{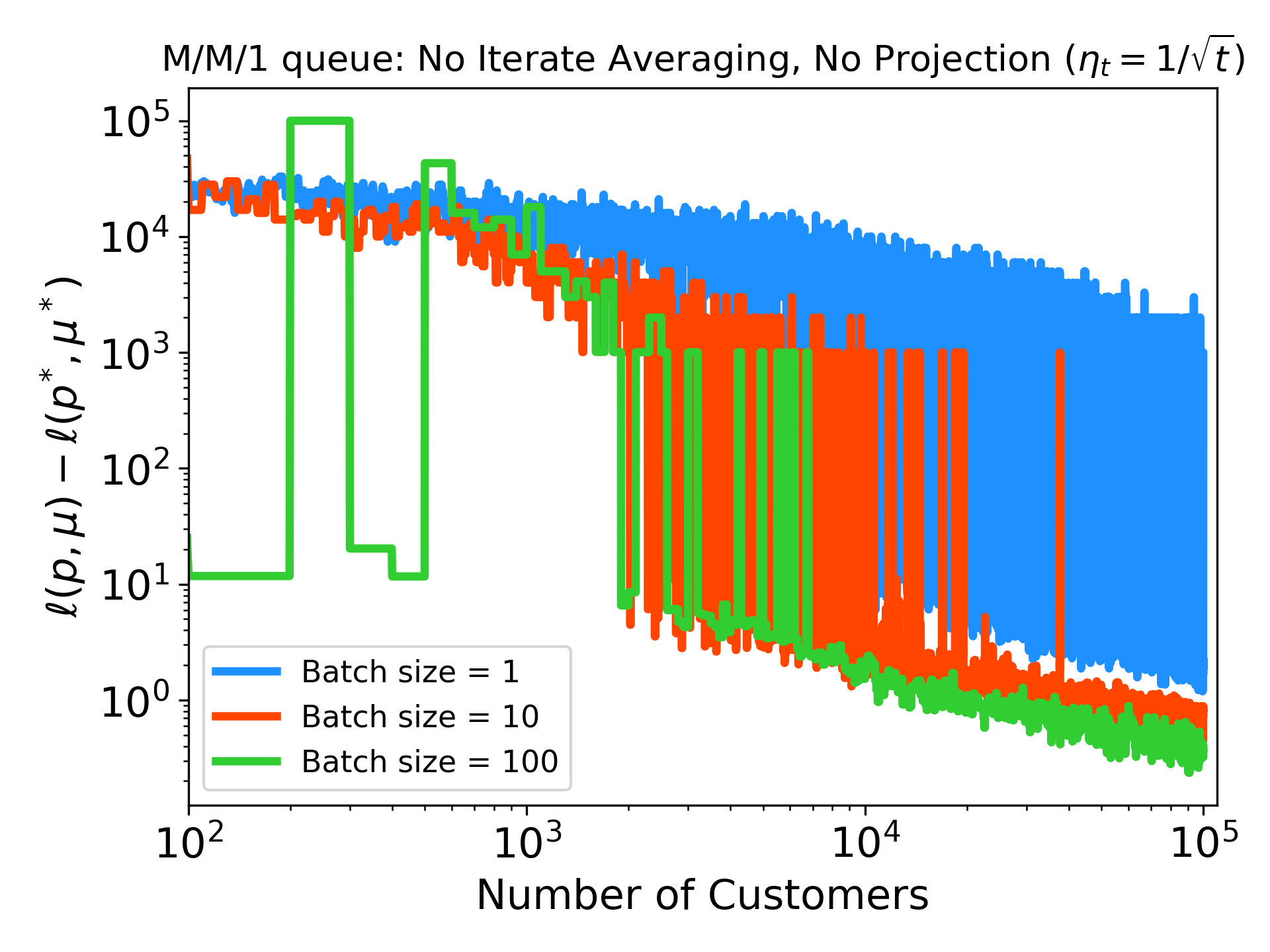}
    \includegraphics[height=6cm]{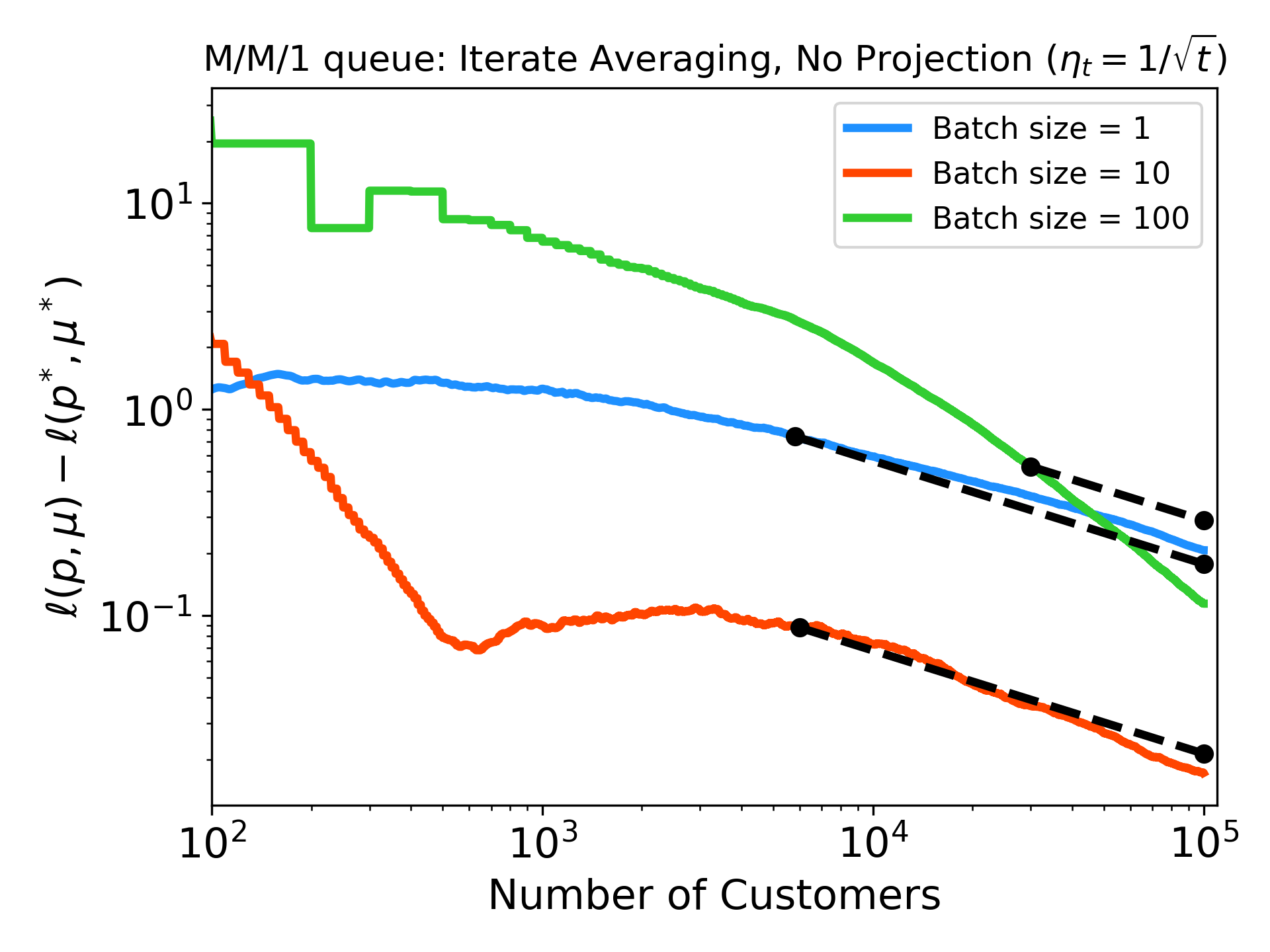}   
    \caption{Pricing and capacity sizing in the single server queue. (Left) Loss gap for the SGD iterates without projection and without iterate averaging with $\eta_{t} = 1/\sqrt{t}$ across batch sizes $B \in \{1, 10, 100\}$. (Right) Loss gap for the SGD iterates without projection but with iterate averaging and $\eta_{t} = 1/\sqrt{t}$. The dotted line displays the convergence rate of $t^{-1/2}$.
    % (Bottom Left) Loss of projected SGD iterates without Polyak averaging with the same step size $\eta_{t} = 1/\sqrt{t}$. (Bottom right) Loss of SGD with i.i.d. stochastic gradients, i.e. $\nabla \ell (\theta) + \epsilon$ with $\epsilon \sim N(0,1)$. No Polyak averaging or projection.
    }
    \label{fig:queuing2}
\end{figure}

We also see in our experiments that instabilities can appear under adaptive feedback if the assumptions for Theorem~\ref{thm:main2} are not satisfied.  For example, Figure~\ref{fig:queuing2} compares the empirical convergence of SGD with step-size schedule $\eta_{t} = 1/\sqrt{t}$, without versus with averaging, and without projection. 
%performing ablation of iterate averaging vs last iterate convergence. 
Note that the two scenarios are outside the scope of our theoretical results because we do not project the parameters to the set in which the queue will be uniformly stable. 
%It's worth mentionging that in the standard i.i.d. noise setting, it is well known that SGD is able to converge without iterate averaging and empirically has strong last iterate performance.
%We observe that iterate averaging is very important for stabilizing the iterates in the adaptive setting. In particular, 
We observe in the left panel of Figure~\ref{fig:queuing2} that without projection and iterate averaging, the adaptive algorithm can be highly unstable. For example, when $B=1$, the cost/loss oscillates between $10$ and $10^{4}$. In the right panel, with iterate averaging, the algorithm converges at rate $t^{-1/2}$. The instabilities in the left panel are a result of the bias incurred by the adaptive feedback. Intuitively, instabilities emerge because the algorithm lowers the service capacity without fully anticipating the increase in congestion that this will incur, since it takes time for customers to arrive. Waiting for more customers to arrive before updating parameters allows the algorithm to better gauge congestion, which leads to better performance for larger batch variants. Averaging or smaller step size, i.e., $t^{-1}$, can help smooth things in this case and lead to more stable learning.
%These instabilities can also be avoided by projecting the parameters to enforce stability or decreasing the step-size schedule to $t^{-1}$ instead of $t^{-1/2}$. 
Overall, this demonstrates that features of the algorithm such as projection, iterate averaging, and small and properly decreasing step sizes, which are usually benign for the empirical performance of SGD with iid data, can be crucial for stable convergence in the adaptive setting.

% Thus,  This explains why in the top right panel of Figure~\ref{fig:queuing} we observe that Polyak averaging, even without projecting a stable parameter set, results in stabler performance of SGD and matches the performance in the stochastic i.i.d. case. While many iterates cause instability, the average iterate is less affected to the oscillations caused by the adaptive feedback. Note that this is unlike the standard i.i.d. noise setting, where the practical performance of SGD is strong even without Polyak averaging.

% Finally, projecting to a stable set, i.e. a set where $\lambda(p) < \mu$, is effective in stabilizing the iterates by preventing the queues from blowing up, as shown in the bottom left panel of Figure~\ref{fig:queuing}. Ultimately, this empirical example shows that broadly the conditions of Theorem~\ref{thm:main1} are useful for guaranteeing the practical performance of SGD in adaptive environments.

%Finally, note that the SGD algorithm featured in \cite{chen2023online} involves increasing batch sizes to mitigate the bias. In particular, the batch sizes scale as $\log(k)$ with $k$ being the number of iterations. Our empirical results illustrate that increasing batch sizes are largely unnecessary for convergence. The algorithm can converge even with a constant batch size of $B = 1$, as established by our theoretical results.

\subsection{Policy Gradient in Reinforcement Learning}

\begin{figure}[!ht]
\centering
    \includegraphics[height=6cm]{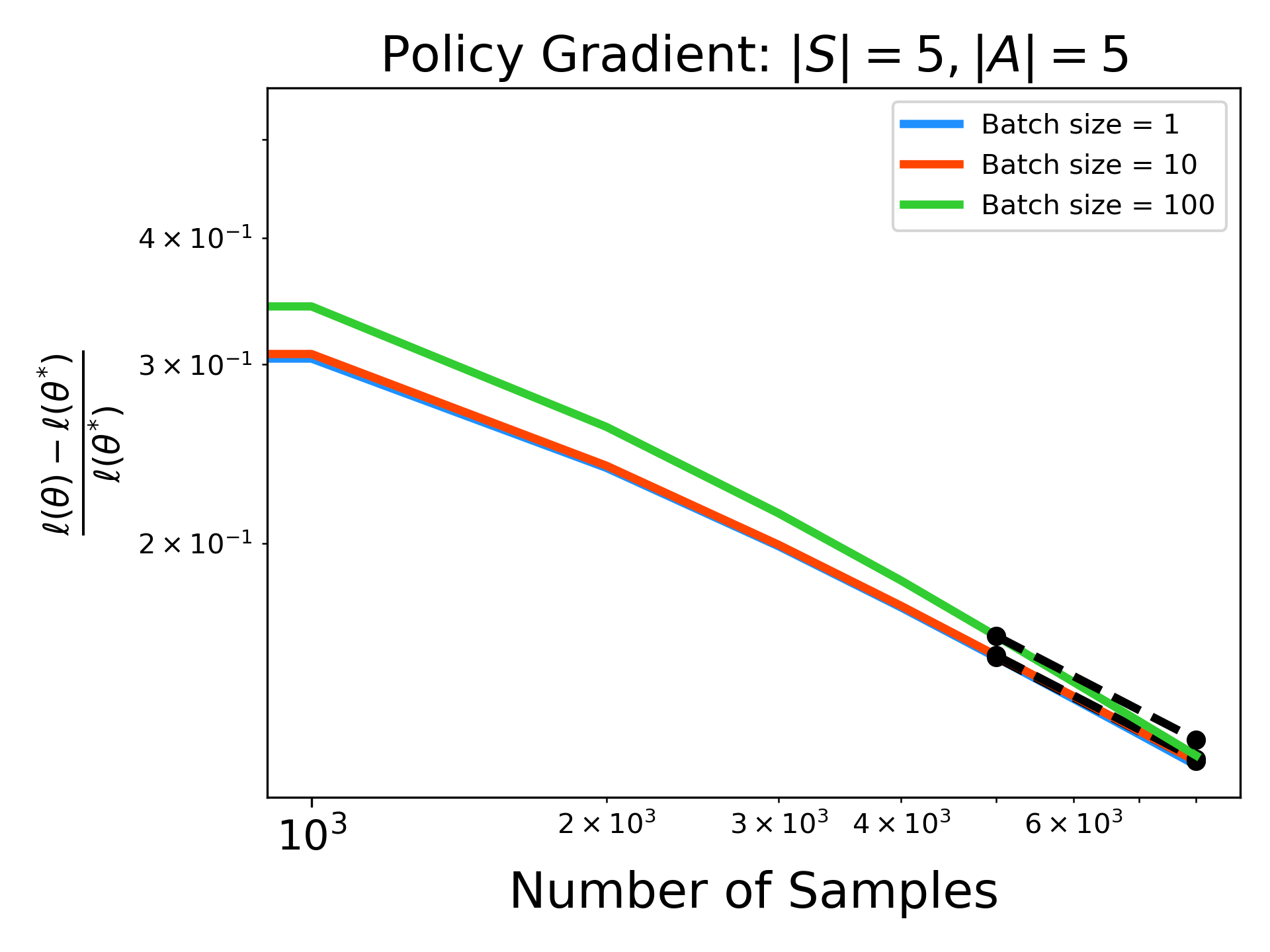}
    \includegraphics[height=6cm]{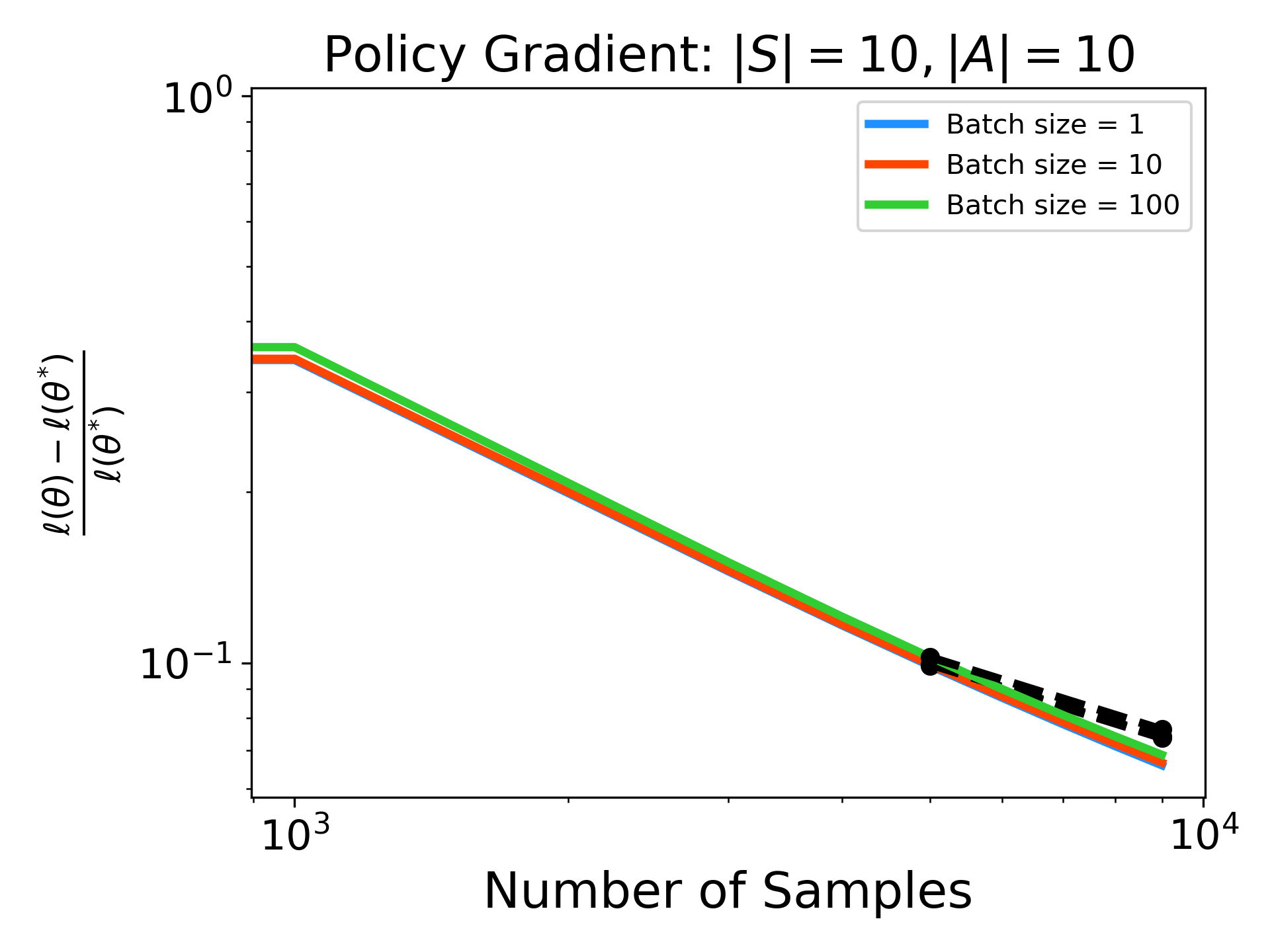}
    \caption{Policy gradient for tabular MDPs. Dotted line indicates a $t^{-1/2}$ convergence rate. (Left) Scaled loss gap for the averaged SGD iterate for the reinforcement learning problem across batch sizes $B \in \{1, 10, 100\}$ for 100 randomly generated MDPs with $|\mathcal{S}|=|\mathcal{A}|=5$. Step-size schedule is $\eta_{t} = 2B/t^{-1/2}$. (Right) Scaled loss gap for the averaged SGD iterates for the reinforcement learning problem for 100 randomly generated MDPs with $|\mathcal{S}|=|\mathcal{A}|=10$. Step-size schedule is $\eta_{t} = 2B/t^{-1/2}$.}
    \label{fig:policy_gradient}
\end{figure}

We evaluate the performance of the actor-critic algorithm (Algorithm~\ref{alg:TD}) in simple tabular RL examples. We consider a direct softmax parameterization of the policy over all states and actions, i.e. $\theta \in \mathbb{R}^{|S| \times |A|}$ and
\[
\pi_{\theta}(a|s) = \frac{\exp(\theta_{s,a})}{\sum_{a\in A} \exp(\theta_{s,a})}.
\]
Since the loss function $\ell(\theta) = \E^{\theta}_{\rho} \left[ \sum_{t=0}^{\infty} \gamma^{t} c(s_{t},a_{t}) \right]$ is not strongly convex in $\theta$, we consider step size schedule with $\eta_{t} \propto t^{-1/2}$ and iterate averaging.
We set the discount factor $\gamma = 0.8$, the TD update parameter $\alpha = 0.5$, and the step-size schedule $\eta_{t} = 2B/t^{-1/2}$, which scales linearly with the batch size. We consider randomly generated MDPs with $|\mathcal{S}| = 5$ states and $|\mathcal A| = 5$ actions, and larger instances with $|\mathcal S| = 10$ states and $|\mathcal A| = 10$ actions. For each instance type, we average performance across 100 randomly generated MDPs by rescaling the cost in each MDP as
\begin{equation}\label{eq:scale_gap}
\frac{\ell(\bar{\theta}_{t}) - \ell(\theta^{*})}{\ell(\theta^{*})}.
\end{equation}
Figure~\ref{fig:policy_gradient} displays the averaged scaled optimality gap \eqref{eq:scale_gap} across different batch sizes. We observe that the fully adaptive actor-critic algorithm, which updates the policy after only a single TD update to the Q-function, is able to achieve an almost identical convergence rate with variants that perform multiple TD updates before updating the policy. This convergence rate is $O(t^{-1/2})$ as indicated by the black line. 

%This is true even for the larger-scale instances, i.e., $|\mathcal{S}|=10$ and $|\mathcal{A}|=10$. 

\section{Conclusion} \label{sec:conclude}
In this work, we study SGD with adaptive data, which arises in many online learning problems in operations research. We provide easy-to-verify conditions under which the fully adaptive SGD update achieves a similar convergence speed as the classical stationary setting. Our results provide guidance and assurance as to how to choose the appropriate batch size in online learning problems where stationary (long-run average) performance is involved. For example,  we demonstrate how to apply the results to study online learning algorithms for some service and inventory management problems.

We conclude the paper with some remarks for future research. First, the conditions we required in Theorems \ref{thm:main1} and \ref{thm:main2} are only sufficient conditions. It would be an interesting future research direction to see both theoretically and empirically, whether similar convergence speeds hold under more general conditions. In particular, while we think the ergodicity and smoothness conditions are important for the convergence of the algorithm, it would be interesting to see if these conditions only need to hold locally, i.e., around the optimal solution. Second, the upper bounds we established in Theorems \ref{thm:main1} and \ref{thm:main2} are unlikely to be tight, especially regarding the logarithmic terms. We leave it as a future research direction to establish tight bounds. Lastly, from our numerical experiments, even though the batch size does not affect the convergence rate of the algorithm, it can improve the convergence speed through the constant term. It would be valuable to develop theoretical results that can guide the choice of the optimal batch size.

\bibliographystyle{plainnat}
\bibliography{RL_ref}

%\newpage

\begin{appendices}

\section{Proofs of Theorems \ref{thm:main1} and \ref{thm:main2}}
Define $\eta_{t:t+n}:=\sum_{s=t}^{t+n-1} \eta_s$, $\eta_{t:t+n}^2:=\sum_{s=t}^{t+n-1} \eta_s^2$, $(V\eta)_{t:t+n}=\sum_{s=t}^{t+n-1}V(z_s)\eta_s$, $(V\eta)_{t:t+n}^2=\sum_{s=t}^{t+n-1}V(z_s)^2\eta_s^2$ and $P_{\theta_{m:n}}=P_{\theta_m}P_{\theta_1} \dots P_{\theta_{n-1}}$.
We also define the $\epsilon$-mixing time $\tau_{\epsilon}$ as a time for which under Assumption \ref{ass:ergodicity0},
\begin{equation}\label{eq:mixing_time}
d(\mu_{\theta},\delta_{z}P_{\theta}^{\tau_\epsilon}) \leq \epsilon V(z).
\end{equation}
%When we update with projected gradient descent \eqref{eq:genSGD2}, we only require \eqref{eq:mixing_time} to hold for $\theta\in \mathcal{C}$.

We first prove some auxiliary lemmas. These lemmas hold for both the unprojected and projected updates, i.e., \eqref{eq:genSGD} and \eqref{eq:genSGD2} respectively. For concision, we only list the assumptions for the unprojected case. For the projected case, these assumptions are only required to hold for $\theta\in \Theta$.

\begin{lemma}\label{lm:epsilon_mix}
If $P_{\theta}$ has a mixing time $\tau$, then for $\tau_\epsilon=\lceil |\log \epsilon|/|\log(1/4)|\rceil \tau $, we have
\[
d(\delta_{z} P_{\theta}^{\tau_\epsilon},\mu_{\theta}(\cdot))\leq \epsilon V(z),
\]
%\[
%\|\Prob_{\mu}(z_{\tau_\epsilon}\in \,\cdot\,)-\mu(\cdot)\|_{TV}\leq \epsilon
%\]
i.e., $\tau_{\epsilon}$ is an $\epsilon$-mixing time.
\end{lemma}
\proof{Proof.}
For any function $f$ with $|f(z)|\leq V(z)$ for all $z$,
we define $f_0(z):=f(z)-\sum_{z'} \mu_{\theta}(z') f(z')$. Note that $\sum_z \mu_{\theta}(z)f_0(z)=0$.
%{\color{red} and $|f_0(z)|\leq V(z)$}.
We also define $f_1(z):=\E_{z} f_0(z_\tau)$. Note that
\[
\sum_{z}\mu_{\theta}(z) f_1(z)=\sum_{z} \mu_{\theta}(z)\sum_{z'}\Prob_z(z_\tau=z')f_0(z')=\sum_{z'} \mu_{\theta}(z') f_0(z')=0
\]
and
\[
|f_1(z)|= \left|\langle \delta_{z} P_{\theta}^{\tau}-\mu_{\theta}, f(z')\rangle\right|\leq \frac14 V(z). 
\]
We can repeat the above procedure and define a sequence of $f_k$'s.
For $f_k(z):=\E_z[f_0(z_{k\tau})]$ for $k\geq 2$, suppose $|f_{k-1}(z)|\leq \frac{1}{4^{k-1}}V(z)$. Then, we have
\[
\sum_{z}\mu(z) f_k(z)=\sum_{z} \mu_{\theta}(z)\sum_{z'}\Prob_z(z_{k\tau}=z') f_0(z')=\sum_{z'} \mu_{\theta}(z') f_0(z')=0,
\]
and
\[
|f_k(s)|=\frac{1}{4^{k-1}}\left|\langle \delta_{z} P_{\theta}^{\tau}-\mu_{\theta}, 4^{k-1} f_{k-1}(z')\rangle\right| \leq \frac{1}{4^{k}}V(z).
\]
Let $n=\lceil |\log \epsilon|/|\log(1/4)|\rceil$. Then,
\[
f_n(z)=\E_{z} f(z_{\tau_\epsilon}) -\sum_{z'} \mu(z') f(z')
\mbox{ and }
|f_n(z)|\leq\frac{1}{4^n} V(z)\leq \epsilon V(z).
\] 
Since $f$ is any function with $|f(z)|\leq V(z)$, we have $d(\delta_{z} P_{\theta}^{\tau_\epsilon},\mu(\cdot))\leq \epsilon V(z)$.

Next 
\[\begin{split}
&\left|\E_t \langle \bar g(\theta_t)-g(\theta_t,z_{t+\tau_{\epsilon}}),\theta_t-\theta^*\rangle\right|\\
%=&\left|\left\langle\sum_{z'}(\mu(z') - \Prob_{z_t}(z_{t+\tau_{\epsilon}}=z'))g(\theta_t,z'),\theta_t-\theta^*\right\rangle\right|\\
\leq&\left\|\sum_{z'}(\mu(z') - \Prob_{z_t}(z_{t+\tau_{\epsilon}}=z'))g(\theta_t,z')\right\|\|\theta_t-\theta^*\| \mbox{ by Cauchy-Schwarz inequality}\\
\leq&\epsilon C_0\|\theta_t-\theta^*\| \leq 2C_0C_1\epsilon.
\end{split}\]
\Halmos
\endproof

\begin{lemma}
\label{lem:lip0}
Under Assumptions \ref{ass:bound0}, the iterates satisfy
\[
\|\theta_{t+n}-\theta_t\|\leq M(V\eta)_{t:t+n}.
\]

\end{lemma}
\proof{Proof.}
We first note that
for the unprojected cases, $\|\theta_{t+1}-\theta_t\|=\|\eta_t\ghat_t(\theta_t, z_t)\|$.
For the projected case, since $\mathcal{C}$ is convex,
$\|\theta_{t+1}-\theta_t\|\leq \|\eta_t\ghat_t(\theta_t, z_t)\|$.
Thus,
\[
\|\theta_{t+1}-\theta_t\|\leq \|\eta_t\ghat_t(\theta_t, z_t)\|\leq MV(z_t)\eta_t.
\]
%\red{(we need $\|\ghat\|\leq M$?)}
Then,
\[
\|\theta_{t+n}-\theta_t\| \leq \sum_{k=t}^{t+n-1}\|\theta_{k+1}-\theta_k\| \leq M(V\eta)_{t:t+n}.
\]
\Halmos
\endproof

%{\color{blue} Need to discuss $\theta_{0:n}$. Is it ``given"?}
\begin{lemma} \label{lm:ergo}
Under Assumptions \ref{ass:ergodicity0} and \ref{ass:lip0}, the following holds almost surely,
\[
d(\delta_x P_{\theta_{0:n}}, \delta_xP_{\theta_{0}}^n )\leq 
\frac{LK^2MV(x)}{(1-\rho)^2}(V\eta)_{0:n}.
\]
\end{lemma}
\proof{Proof.}
First, let $\Pi$ denote the optimal coupling between $\mu$ and $\nu$ under $d$. Then,
\[\begin{split}
d(\mu P_{\theta_0}^{k}, \nu P_{\theta_0}^{k})&\leq \sum_{x,y} d(\delta_x P_{\theta_0}^{k}, \delta_y P_{\theta_0}^{k})\Pi(x,y) \\
&\leq  K\rho^k \sum_{x,y} d(x,y)\Pi(x,y)=K\rho^k d(\mu,\nu).
\end{split}\]
Next,
\[
d(\mu P_\theta, \mu P_{\theta'})\leq \sum_{x}  
d(\delta_x P_\theta, \delta_x P_{\theta'})\mu(x)
\leq L\|\theta-\theta'\|\sum_{x} V(x)\mu(x). 
\]
We also note that
\[\begin{split}
 \langle\delta_x P_{\theta_{0:m}}, V\rangle
 &\leq \rho P_{\theta_{0:m-1}}V(x) + K\\
 &\leq \rho^mV(x) + \rho^{m-1} K + \dots + \rho K+ K \\
& \leq \frac{KV(x)}{1-\rho}. 
\end{split}\]
Lastly, 
\begin{align*}
d(\delta_xP_{\theta_{0:n}}, \delta_x P_{\theta_0}^n)
&\leq \sum_{m=2}^{n} d(\delta_x P_{\theta_{0:m}}P_{ \theta_0}^{n-m}, \delta_xP_{\theta_{0:m-1}}P_{ \theta_0}^{n-m+1})\\
&\leq \sum_{m=2}^{n} K\rho^{n-m} d(\delta_x P_{\theta_{0:m-1}}P_{\theta_{m-1}}, \delta_x P_{\theta_{0:m-1}} P_{\theta_{0}} )\\
&\leq \sum_{m=2}^{n} K\rho^{n-m}L\|\theta_{m-1}-\theta_0\| \langle \delta_x P_{\theta_{0:m-1}}, V\rangle \\
&\leq \sum_{m=2}^{n} K\rho^{n-m}LM (V\eta)_{0:m-1}\frac{KV(x)}{1-\rho}
\leq \frac{LK^2MV(x)}{(1-\rho)^2}(V\eta)_{0:n}.
\end{align*}
\Halmos
\endproof

Similar to Lemma \ref{lm:ergo}, we also have
\begin{lemma} \label{lm:ergo2}
Under Assumptions \ref{ass:ergodicity0} and \ref{ass:lip0},
\[
d(\delta_x P_{\theta_{0:n}}, \delta_xP_{\theta_{n}}^n )\leq 
\frac{LK^2MV(x)}{(1-\rho)^2}(V\eta)_{0:n}.
\]
\end{lemma}
\proof{Proof.}
%First, let $\Pi$ denote the optimal coupling between $\mu$ and $\nu$ under $d$. Then,
%\[\begin{split}
%d(\mu P_{\theta_n}^{k}, \nu P_{\theta_n}^{k})&\leq \int \Pi(dx,dy)  d(\delta_x P_{\theta_n}^{k}, \delta_y P_{\theta_n}^{k})\\
%&\leq \int K\rho^k \int \Pi(dx,dy)d(x,y)=K\rho^k d(\mu,\nu).
%\end{split}\]
%Next,
%\[
%d(\mu P_\theta, \mu P_{\theta'})\leq \int \mu(dx) 
%d(\delta_x P_\theta, \delta_x P_{\theta'})
%\leq L\|\theta-\theta'\|\int V(x)\mu(dx) 
%\]
%We also note that
%\[\begin{split}
% \langle\delta_x P_{\theta_{0:m}}, V\rangle
% &\leq \rho P_{\theta_{0:m-1}}V(x) + K\\
% &\leq \rho^mV(x) + \rho^{m-1} K + \dots + \rho K+ K \\
%& \leq \frac{KV(x)}{1-\rho}. 
%\end{split}\]
%Lastly, 
Define $P_{\theta_{m:m}}=I$ and $P_{\theta_n}^0=I$.
\begin{align*}
d(\delta_xP_{\theta_{0:n}}, \delta_x P_{\theta_{n}}^n)
&\leq \sum_{m=1}^{n} d(\delta_x P_{\theta_{0:m}}P_{ \theta_n}^{n-m}, \delta_xP_{\theta_{0:m-1}}P_{ \theta_n}^{n-m+1})\\
&\leq \sum_{m=1}^{n} K\rho^{n-m} d(\delta_x P_{\theta_{0:m-1}}P_{\theta_{m-1}}, \delta_x P_{\theta_{0:m-1}} P_{\theta_{n}} )\\
&\leq \sum_{m=1}^{n} K\rho^{n-m}L\|\theta_{m-1}-\theta_n\| \langle \delta_x P_{\theta_{0:m-1}}, V\rangle \\
&\leq \sum_{m=1}^{n} K\rho^{n-m}LM (V\eta)_{m-1:n}\frac{KV(x)}{1-\rho}
\leq \frac{LK^2MV(x)}{(1-\rho)^2}(V\eta)_{0:n}.
\end{align*}
\Halmos
\endproof

%\blue{Later on, we use $\nabla l(\theta_t)\rangle$, not $\theta_t-\theta^*\rangle$}
\begin{lemma}\label{lm:mix}
Under Assumptions \ref{ass:ergodicity0} and \ref{ass:lip0}, we have 
%\[
%\left|\E_t \langle \bar g(\theta_{t})-g(\theta_{t},z_{t+\tau_{\epsilon}}),\theta_t-\theta^*\rangle\right| \leq 
%2C\left(L\epsilon+ \frac{L^2K^2M}{(1-\rho)^2} \eta_{t:t+\tau_\epsilon}\right) V(z_t).
%\]
%\blue{Maybe we just show }
\[
\|\nabla \ell(\theta_{t})-\E_t g(\theta_{t},z_{t+\tau_{\epsilon}})\| \leq 
L\epsilon V(z_{t-1})+\frac{L^2K^2M}{(1-\rho)^2} \eta_{t:t+\tau_\epsilon}(V(z_{t-1})+K)^2.
\]
%{\color{blue} Define $z_{t+\tau_{\epsilon}}$.}
\end{lemma}
\proof{Proof.}
%Since $\|g(\theta, z)-g(\theta, z^{\prime})\|\leq G_2d(z,z^{\prime})$,
Note that 
%\blue{(I am not sure if this should be $V(z_t)$ or $V(z_{t-1})$, can you check?)}
\[\begin{split}
&\|\nabla \ell(\theta_{t})-\E_t g(\theta_{t},z_{t+\tau_{\epsilon}})\|\\
\leq&\left\|\langle \mu_{\theta_t}-\delta_{z_{t-1}}P_{\theta_t}^{\tau_\epsilon},g(\theta_t,\,\cdot\,)\rangle\right\| +\left\|\E_t\langle \delta_{z_{t-1}}P_{\theta_{t:t+\tau_\epsilon}}-\delta_{z_{t-1}}P_{\theta_t}^{\tau_\epsilon},g(\theta_t,\,\cdot\,)\rangle\right\|\\
\leq& L\epsilon V(z_{t-1})+ \frac{L^2K^2MV(z_{t-1})}{(1-\rho)^2}\E_{t}[(V\eta)_{t:t+\tau_\epsilon}] \\
\leq& L\epsilon V(z_{t-1})+ \frac{L^2K^2MV(z_{t-1})}{(1-\rho)^2}(V(z_{t-1})+K)\eta_{t:t+\tau_\epsilon} 
%\leq& 2C\left(L\epsilon+ \frac{L^2K^2V(z_t)}{(1-\rho)^2} (V\eta)_{t:t+\tau_\epsilon}\right) V(z_t)
\end{split}\] 
by Lemma \ref{lm:ergo}, Kantorovich-Rubenstein duality, and the fact that $V$ is a Lyapunov function.
\Halmos
\endproof

%Similar to Lemma \ref{lm:mix}, we also have the following bound.

\begin{lemma}\label{lm:mix2}
Under Assumptions \ref{ass:ergodicity0} and \ref{ass:lip0}, we have
%\[
%\left|\E_t \langle \bar g(\theta_{t+\tau_\epsilon})-g(\theta_{t+\tau_\epsilon},z_{t+\tau_{\epsilon}}),\theta_t-\theta^*\rangle\right| \leq 
%2C\left(L\epsilon+ \frac{L^2K^2M}{(1-\rho)^2} \eta_{t:t+\tau_\epsilon}\right) V(z_t).
%\]
\[
\left\|\E_t[\nabla \ell(\theta_{t+\tau_\epsilon})-g(\theta_{t+\tau_\epsilon},z_{t+\tau_{\epsilon}})]\right\| \leq 
L\epsilon V(z_{t-1})+\frac{L^2K^2M}{(1-\rho)^2} \eta_{t:t+\tau_\epsilon}(V(z_{t-1})+K)^2.
%\left(L\epsilon+ \frac{L^2K^2M}{(1-\rho)^2} \eta_{t:t+\tau_\epsilon}\right) \blue{(V(z_{t-1})+K)^2}.
\]
\end{lemma}
\proof{Proof.}
%Since $\|g(\theta, z)-g(\theta, z^{\prime})\|\leq G_2d(z,z^{\prime})$,
Note that
\[\begin{split}
&\left\|\E_t[\nabla \ell(\theta_{t+\tau_\epsilon})-g(\theta_{t+\tau_\epsilon},z_{t+\tau_{\epsilon}})]\right\|\\
\leq&\left\|\langle \mu_{\theta_{t+\tau_\epsilon}}-\delta_{z_{t-1}}P_{\theta_{t+\tau_\epsilon}}^{\tau_\epsilon},g(\theta_{t+\tau_\epsilon},\,\cdot\,)\rangle\right\| +\left\|E_t\langle \delta_{z_{t-1}}P_{\theta_{t:t+\tau_\epsilon}}-\delta_{z_{t-1}}P_{\theta_{t+\tau_\epsilon}}^{\tau_\epsilon},g(\theta_{t+\tau_\epsilon},\,\cdot\,)\rangle\right\|\\
\leq& L\epsilon V(z_{t-1})+ \frac{L^2K^2MV(z_{t-1})}{(1-\rho)^2}\E_t[(V\eta)_{t:t+\tau_\epsilon}]\\
\leq& L\epsilon V(z_{t-1})+ \frac{L^2K^2MV(z_{t-1})}{(1-\rho)^2}(V(z_{t-1})+K)\eta_{t:t+\tau_\epsilon}
%\leq& 2C\left(L\epsilon+ \frac{L^2K^2V(z_t)}{(1-\rho)^2} \eta_{t:t+\tau_\epsilon}\right) V(z_t)
\end{split}\]
by Lemma \ref{lm:ergo2}, Kantorovich-Rubenstein duality, and the fact that $V$ is a Lyapunov function.
\Halmos
\endproof

\begin{lemma}
\label{lem:lip}
Under Assumptions \ref{ass:bound0} and \ref{ass:lip0}, the iterates satisfy the following 
\[
\|\E_t [g(\theta_t, z_{t+\tau_\epsilon})-g(\theta_{t+\tau_\epsilon},z_{t+\tau_\epsilon})]\| \leq \left(2L\epsilon+ 2\frac{L^2K^2M}{(1-\rho)^2} \eta_{t:t+\tau_\epsilon}+ML\eta_{t:t+\tau_\epsilon}\right) (V(z_{t-1})+K)^2.
%+ML\eta_{t:t+\tau_\epsilon}.
\]
\end{lemma}
\proof{Proof.}
Note that
\[\begin{split}
&\|\E_t[g(\theta_t, z_{t+\tau_\epsilon})-g(\theta_{t+\tau_\epsilon},z_{t+\tau_\epsilon})]\|\\
\leq &\| \E_t g(\theta_t, z_{t+\tau_\epsilon})-\nabla l(\theta_{t})\|+
\| \E_t[g(\theta_{t+\tau_\epsilon}, z_{t+\tau_\epsilon})-\nabla l(\theta_{t+\tau_\epsilon})]\|+
\| \E_t\nabla l(\theta_{t+\tau_\epsilon})-\nabla l(\theta_{t})\|\\
\leq & 2L\epsilon V(z_{t-1})+ 2\frac{L^2K^2M}{(1-\rho)^2} \eta_{t:t+\tau_\epsilon}(V(z_{t-1})+K)^2+ML(V(z_{t-1})+K)\eta_{t:t+\tau_\epsilon} 
\end{split}\]
by Lemmas \ref{lm:mix} and \ref{lm:mix2}.
\Halmos
\endproof

We are now ready to prove the main theorems. 
%To make the presentation concise, we divide the proof into three parts. We first prove Cases 1 and 2 in Theorem \ref{thm:main1}, i.e., the cases without strong convexity. We then prove Case 1 (the convex case) in Theorem \ref{thm:main2}, which is different from and simpler than the proof of Case 2 in Theorem \ref{thm:main1} due to projection. Lastly, we prove Case 2 in Theorem \ref{thm:main1} and Case 2 in Theorem \ref{thm:main2}, i.e., the strongly convex cases, which share the same proof.

\proof{Proof of Theorem \ref{thm:main1}.}

Since $\theta_{t+1}=\theta_t-\eta_t \ghat_t(\theta_t,z_t)
$ and $\nabla \ell(\theta)$ is $L$-smooth,
\begin{align*}
\ell(\theta_{t+1})&\leq \ell(\theta_{t})-\eta_t\langle \nabla \ell(\theta_t), \ghat_t(\theta_t, z_t)\rangle+\frac{L}{2}\eta_t^2\|\ghat_t(\theta_t, z_t)\|^2.
\end{align*}
%\blue{Issue: this seems not to be true in general. }

The above inequality can be rearranged as
\[
\eta_t\langle \nabla \ell(\theta_t), \nabla \ell(\theta_t)\rangle \leq \ell(\theta_{t})-\ell(\theta_{t+1}) + \frac{L}{2} \eta_t^2M^2V(z_t)^2
+\eta_t\langle \nabla \ell(\theta_t)-\ghat_t(\theta_t, z_t),\nabla \ell(\theta_t)\rangle. 
\]
Taking the sum over $t$, we have
\begin{align*}
\sum_{t=0}^{T-1} \eta_t\|\nabla \ell(\theta_t)\|^2\leq \ell(\theta_0)+\frac12 LM^2(V\eta)_{0:T}^2 +\sum_{t=0}^{T-1} \eta_t\langle \nabla \ell(\theta_t)-\ghat_t(\theta_t, z_t), \nabla \ell(\theta_t)\rangle. 
\end{align*}

We next establish an appropriate bound for $\langle \nabla \ell(\theta_t)-\ghat_t(\theta_t, z_t), \nabla \ell(\theta_t)\rangle$.
Consider the decomposition
\begin{align*}
&\langle\nabla \ell(\theta_t)-\ghat_t(\theta_t, z_t), \nabla \ell(\theta_t)\rangle\\
=&\underbrace{\langle g(\theta_t,z_t)-\ghat_t(\theta_t, z_t), \nabla \ell(\theta_t)\rangle}_{(a)}+
\underbrace{\langle g(\theta_{t+\tau_\epsilon},z_{t+\tau_\epsilon})-g(\theta_t,z_t), \nabla \ell(\theta_t)\rangle}_{(b)}\\
&+\underbrace{\langle g(\theta_{t},z_{t+\tau_\epsilon})-g(\theta_{t+\tau_\epsilon},z_{t+\tau_\epsilon}), \nabla \ell(\theta_t)\rangle}_{(c)}
+\underbrace{\langle \nabla \ell(\theta_t)-g(\theta_{t},z_{t+\tau_\epsilon}), \nabla \ell(\theta_t)\rangle}_{(d)},
\end{align*}
where $z_{t+\tau_\epsilon}\sim P_{\theta_{t:t+\tau_{\epsilon}}}$. \\
%\blue{We need to add in $\E_t$ in below, do we add that here as well?}\\
For (a), we have
\[
|\E_t\langle g(\theta_t,z_t)-\ghat_t(\theta_t,z_t), \nabla \ell(\theta_t)\rangle|
\leq \| \nabla \ell(\theta_t)\|\E_t\|g(\theta_t,z_t)-\ghat_t(\theta_t,z_t)\|\leq M\E_t e_t.
\]
For (c), by Lemma \ref{lem:lip}, we have
\[\begin{split}
&|\E_t\langle g(\theta_{t},z_{t+\tau_\epsilon})-g(\theta_{t+\tau_\epsilon},z_{t+\tau_\epsilon}),  \nabla \ell(\theta_t)\rangle| \\
\leq&\|\E_t[g(\theta_{t},z_{t+\tau_\epsilon})-g(\theta_{t+\tau_\epsilon}, z_{t+\tau_\epsilon})]\| \|\nabla \ell(\theta_t)\|\\
\leq& M\left(2L\epsilon+ 2\frac{L^2K^2M}{(1-\rho)^2} \eta_{t:t+\tau_\epsilon}+ML\eta_{t:t+\tau_\epsilon}\right) (V(z_{t-1})+K)^2\\
%+M^2L\eta_{t:t+\tau_\epsilon}\\
\leq& M\left(2L\epsilon+ 2\frac{L^2K^2M}{(1-\rho)^2} \tau_\epsilon\eta_{t}+ML\tau_\epsilon\eta_{t}\right) (V(z_{t-1})+K)^2.
%+M^2L\tau_\epsilon\eta_{t}
%\leq& L\|\theta_t - \theta_{t+\tau_{\epsilon}}\|M
%\leq LM^2 \eta_{t:t+\tau_\epsilon}\leq 
%\blue{2C\left(L\epsilon+ \frac{L^2K^2M}{(1-\rho)^2} \eta_{t:t+\tau_\epsilon}\right) V(z_t)+L\eta_{t:t+\tau_\epsilon}}
\end{split}\]
For (d), following Lemma \ref{lm:mix}, we have
\[\begin{split}
&\left|\E_t[\langle \nabla \ell(\theta_t)-g(\theta_t,z_{t+\tau_{\epsilon}}), \nabla \ell(\theta_t)\rangle]\right|\\
%\leq&\E_t[\|\bar g(\theta_t)-g(\theta_t,z_{t+\tau_{\epsilon}})\|]\|\nabla l(\theta_t)\|\\
%\leq& M\epsilon V(z_t)
%\leq& 2C_0C_1\epsilon V(z_t)+ \blue{2LC_1\tau_\epsilon \eta_{t:t+\tau_\epsilon}}
\leq&\|\E_t[\nabla \ell(\theta_t)-g(\theta_t,z_{t+\tau_{\epsilon}})]\| \|\nabla \ell(\theta_t)\|\\
\leq& M\left(L\epsilon+ \frac{L^2K^2M}{(1-\rho)^2} \eta_{t:t+\tau_\epsilon}\right) (V(z_{t-1})+K)^2\\
\leq &M\left(L\epsilon+ \frac{L^2K^2M}{(1-\rho)^2} \tau_\epsilon\eta_{t}\right) (V(z_{t-1})+K)^2.
\end{split}\]
Lastly, for (b), taking the sum over $t$, we have
\begin{align*}
&\left|\E\sum_{t=0}^{T-1} \eta_t \langle g(\theta_{t+\tau_\epsilon},z_{t+\tau_\epsilon})-g(\theta_t,z_t), \nabla \ell(\theta_t)\rangle\right|\\
\leq &\left|\E\sum_{t=\tau_\epsilon}^{T-1} \langle g(\theta_{t},z_t), \eta_{t-\tau_\epsilon}\nabla \ell(\theta_{t-\tau_\epsilon})-\eta_t\nabla \ell(\theta_t)\rangle\right|
+\left|\E\sum_{t=0}^{\tau_\epsilon-1} \langle g(\theta_{t},z_t), \eta_t\nabla \ell(\theta_t)\rangle\right|\\
&+\left|\E\sum_{t=T}^{T+\tau_\epsilon-1} \langle g(\theta_{t},z_t), \eta_{t-\tau_\epsilon}\nabla \ell(\theta_{t-\tau_\epsilon})\rangle\right|\\
\leq &\left|\E\sum_{t=\tau_\epsilon}^{T-1} \langle g(\theta_{t},z_t), \eta_{t}\nabla \ell(\theta_{t-\tau_\epsilon})-\eta_t\nabla \ell(\theta_t)\rangle\right|+
\left|\E\sum_{t=\tau_\epsilon}^{T-1} \langle g(\theta_{t},z_t), (\eta_{t-\tau_\epsilon}-\eta_t)\nabla \ell(\theta_{t-\tau_\epsilon})\rangle\right|\\
&+\left|\E\sum_{t=0}^{\tau_\epsilon-1} \langle g(\theta_{t},z_t), \eta_t\nabla \ell(\theta_t)\rangle\right|+
\left|\E\sum_{t=T}^{T+\tau_\epsilon-1} \langle g(\theta_{t},z_t), \eta_{t-\tau_\epsilon}\nabla \ell(\theta_{t-\tau_\epsilon})\rangle\right|\\
\leq& M^2L\E\sum_{t=\tau_\epsilon}^{T-1}\eta_tV(z_t)(V\eta)_{t-\tau_\epsilon:t}
+M^2(V(z_0)+K)\eta_{0:\tau_{\epsilon}} \\
&+M^2\E(V\eta)_{0:\tau_{\epsilon}} + M^2(V(z_0)+K)\eta_{T-\tau_{\epsilon}:T}\\
%+2C_0C_1\eta_0+2C_0C_1\tau_\epsilon+2C_0C_1\tau_\epsilon\\
\leq& M^2L\tau_\epsilon\sum_{t=0}^{T-1}\E(V(z_t)+K)^2\eta_t^2 + 3M^2\tau_\epsilon(V(z_0)+K).
\end{align*}
%where $\eta_{t:t+n}^2:=\sum_{s=t}^{t+n-1}\eta_s^2$.

Putting the bounds for (a) -- (d) together, we have
\[\begin{split}
&\left|\E\sum_{t=0}^{T-1}\eta_t \langle \nabla \ell(\theta_t) -\ghat_t(\theta_t, z_t), \nabla \ell(\theta_t)\rangle\right|\\
\leq& M\sum_{t=0}^{T-1}\eta_t\E e_t 
%+2M^2L\tau_\epsilon\eta_{0: T}^2
+3M^2\tau_\epsilon(V(z_0)+K)
+3ML\epsilon \sum_{t=0}^{T-1}\eta_t\E (V(z_{t})+K)^2 \\
&+ \left(3\frac{M^2L^2K^2}{(1-\rho)^2} \tau_\epsilon+2M^2L\tau_\epsilon \right)\sum_{t=0}^{T-1}\eta^2_{t}\E (V(z_{t})+K)^2.
%&+4C_0C_1\tau_\epsilon
%+2GC_0C_1\tau_{\epsilon}\eta^2_{1:(T+1)}+2C_0C_1 \epsilon \sum_{t=1}^T \blue{\eta_tV(z_t)}+ \blue{2\tau_\epsilon LC_1 \eta^2_{1:T}} 
\end{split}\]

Thus,
\begin{align*}
\sum_{t=0}^{T-1} \eta_t\E\|\nabla \ell(\theta_t)\|^2
\leq& \ell(\theta_0)+\frac12 LM^2(V\eta)_{0:T} +\sum_{t=0}^{T-1} \eta_t\E\langle \bar g(\theta_t)-\ghat_t(\theta_t, z_t), \nabla \ell(\theta_t)\rangle \\
\leq& \ell(\theta_0)+3M^2\tau_\epsilon(V(z_0)+K)
%+3LM^2\tau_{\epsilon}\eta^2_{0:T}
+M\sum_{t=0}^{T-1}\eta_t\E e_t +3ML\epsilon \sum_{t=0}^{T-1}\eta_t \E(V(z_{t})+K)^2 \\
&+ \left(3\frac{M^2L^2K^2}{(1-\rho)^2} \tau_\epsilon+2M^2L\tau_{\epsilon}\right)\sum_{t=0}^{T-1}\eta^2_{t}\E(V(z_{t})+K)^2.
\end{align*}
For $\epsilon=1/\sqrt{T}$, $\tau_{\epsilon}=O(\tau\log T)$. Let $\tilde\eta_t=\eta_t/\eta_{0:T}$. Then, we have
% \[
% \sum_{t=0}^{T-1} \eta_t\E\langle \bar g(\theta_t), \nabla l(\theta_t)\rangle
% =O\left(\tau\log T + \tau\log T\eta_{0:T}^2+\frac{1}{\sqrt{T}}\eta_{0:T}+\sum_{t=0}^{T-1}\eta_t\E e_t \right)
% \]
\[
\sum_{t=0}^{T-1} \tilde \eta_t\E\| \nabla \ell(\theta_t)\|^2
=O\left(\frac{1}{\eta_{0:T}}\left(\tau\log T + \tau\log T\eta_{0:T}^2+\frac{1}{\sqrt{T}}\eta_{0:T}+\sum_{t=0}^{T-1}\eta_t\E e_t \right)\right).
\]
This implies that
\[
\min_{0\leq t< T}\E \| \nabla \ell(\theta_t)\|^2=O\left(\frac{1}{\eta_{0:T}}\left(\tau\log T + \tau\log T\eta_{0:T}^2+\frac{1}{T}\eta_{0:T}+\sum_{t=0}^{T-1}\eta_t\E e_t \right)\right).
\]

\Halmos
\endproof

Before we prove Case 1 of Theorem \ref{thm:main2}, we provide a bound for $\E\|\theta_t-\theta^*\|^2$ first. Note that when $\Theta$ is bounded, $\|\theta_t-\theta\|^2$ is bounded. We will show in Lemma \ref{lem:thetabound} that even when $\Theta$ is not bounded, we can still establish appropriate bounds for $\E\|\theta_t-\theta^*\|^2$.

\begin{lemma}
\label{lem:thetabound}
Assume $l$ is convex and $\E e_t=O(1/\sqrt{t})$.
Let $\eta_t=\eta_0/\sqrt{t}$. %and $\epsilon=1/\sqrt{T}$},
%Then, 
%$\E V(z_t)^2\leq M$ and
There exists a constant $C\in(0,\infty)$, such that
for any fixed $T>0$, we have 
\[\E\|\theta_T-\theta^*\|^2\leq C^2\left(\|\theta_0-\theta^*\|+\tau_{\epsilon}\log T+\frac{1}{\sqrt{T}}\sqrt{t}+(V(z_0)+K)^2\right)^2,\]
for $t\leq T$.
%(There might be a factor of $\log t$ missing somewhere)
\end{lemma}
\proof{Proof.}
Let $\epsilon=1/\sqrt{T}$. Then $\tau_{\epsilon}=O(\tau\log T)$.

By the convexity of $\Theta$ and the fact that $\theta^*\in \Theta$, we have
\begin{align*}
\|\theta_{t+1}-\theta^*\|^2&\leq\|\theta_{t}-\theta^*-\eta_t \ghat_t(\theta_t, z_t) \|^2\\
%&=\|\theta_{t}-\theta^*\|^2+2\eta_t \langle \ghat_t(\theta_t, z_t),\theta^*-\theta_t\rangle+\eta_t^2 \|\ghat_t(\theta_t, z_t)\|^2\\
&\leq \|\theta_{t}-\theta^*\|^2+2\eta_t \langle \ghat_t(\theta_t, z_t),\theta^*-\theta_t\rangle+\eta_t^2M^2V(z_t)^2\\
&= \|\theta_{t}-\theta^*\|^2+2\eta_t \langle \nabla \ell(\theta_t),\theta^*-\theta_t\rangle+\eta_t^2M^2 V(z_t)^2
+2\eta_t \langle \ghat_t(\theta_t,z_t)-\nabla \ell(\theta_t),\theta^*-\theta_t\rangle\\
&\leq \|\theta_{t}-\theta^*\|^2+\eta_t^2M^2 V(z_t)^2
+2\eta_t \langle \ghat_t(\theta_t,z_t)-\nabla \ell(\theta_t),\theta^*-\theta_t\rangle,
\end{align*}
since $\ell$ is convex.
By induction, we have
\[
\|\theta_{t}-\theta^*\|^2
\leq \|\theta_{0}-\theta^*\|^2
+ M^2(V\eta)_{0:t}^2
+2\sum_{s=0}^{t-1}\eta_s \langle \ghat_s(\theta_s,z_s)-\nabla \ell(\theta_s),\theta^*-\theta_s\rangle
\]
Next, note that
\begin{align*}
&\langle\nabla \ell(\theta_s)-\ghat_s(\theta_s, z_s), \theta_s-\theta^*\rangle\\
=&\underbrace{\langle g(\theta_s,z_s)-\ghat_s(\theta_s, z_s), \theta_s-\theta^*\rangle}_{(a)}+
\underbrace{\langle g(\theta_{s+\tau_\epsilon},z_{s+\tau_\epsilon})-g(\theta_s,z_s), \theta_s-\theta^*\rangle}_{(b)}\\
&+\underbrace{\langle g(\theta_{s},z_{s+\tau_\epsilon})-g(\theta_{s+\tau_\epsilon},z_{s+\tau_\epsilon}), \theta_s-\theta^*\rangle}_{(c)}
+\underbrace{\langle \nabla \ell(\theta_s)-g(\theta_{s},z_{s+\tau_\epsilon}),\theta_s-\theta^*\rangle}_{(d)},
\end{align*}
We shall bound each of the terms in the decomposition. For (a), we have
\[\begin{split}
\eta_s|\E_s\langle g(\theta_s,z_s)-\ghat_s(\theta_s,z_s), \theta_s-\theta^*\rangle|
\leq& \eta_s\|\E_s[g(\theta_s,z_s)-\ghat_s(\theta_s,z_s)]\|\|\theta_s-\theta^*\|\\
\leq& \eta_s\E_se_s \|\theta_s-\theta^*\|.
\end{split}\]
For (c), by Lemma \ref{lem:lip}, we have
\[\begin{split}
&\eta_s|\E_s\langle g(\theta_{s},z_{s+\tau_\epsilon})-g(\theta_{s+\tau_\epsilon},z_{s+\tau_\epsilon}), \theta_s-\theta^*\rangle| \\
\leq&\eta_s\|\E_s[g(\theta_{s},z_{s+\tau_\epsilon})-g(\theta_{s+\tau_\epsilon},z_{s+\tau_\epsilon})]\|\|\theta_s-\theta^*\|\\
\leq &\eta_s\left(2L\epsilon+2\frac{L^2K^2M}{(1-\rho)^2}\eta_{s:s+\tau_\epsilon}+ML\eta_{s:s+\tau_\epsilon}\right) (V(z_{s-1})+K)^2\|\theta_s-\theta^*\|.
%+ \eta_sML \eta_{s:s+\tau_\epsilon}\|\theta_s-\theta^*\|.
\end{split}\]
For (d), by Lemma \ref{lm:mix}, we have
\begin{align*}
&\eta_s\left|\E_s \langle \nabla \ell(\theta_s)-g(\theta_s,z_{s+\tau_{\epsilon}}),\theta_s-\theta^*\rangle\right|\\
&\leq 
\eta_s\left(L\epsilon+\frac{L^2K^2M}{(1-\rho)^2}\eta_{s:s+\tau_\epsilon}\right) (V(z_{s-1})+K)^2\|\theta_s-\theta^*\|.    
\end{align*}

Lastly, for (b), taking the sum over $s$, we have
%\begin{align*}
%&\left|\sum_{t=0}^{T-1}\langle g(\theta_{t+\tau_\epsilon},z_{t+\tau_\epsilon})-g(\theta_t,z_t), \theta_t-\theta^*\rangle\right|\\
%\leq &\left|\sum_{t=\tau_\epsilon}^{T-1} \langle g(\theta_{t},z_t), \theta_{t-\tau_\epsilon}-\theta_t\rangle\right|
%+\left|\sum_{t=0}^{\tau_\epsilon-1} \langle g(\theta_{t},z_t), \theta_t-\theta^*\rangle\right|+
%\left|\sum_{t=T}^{T+\tau_\epsilon-1} \langle g(\theta_{t},z_t), \theta_{t-\tau_\epsilon}-\theta^*\rangle\right|\\
%\leq& M\sum_{t=\tau_\epsilon}^{T-1}M\eta_{t-\tau_\epsilon:t}+2MC\tau_\epsilon+2MC\tau_\epsilon\\
%\leq& M^2\tau_\epsilon\eta_{0:T}+4MC\tau_\epsilon.
%\end{align*} 
\begin{align*}
&\left|\E\sum_{s=0}^{t-1}\eta_s\langle g(\theta_{s+\tau_\epsilon},z_{s+\tau_\epsilon})-g(\theta_s,z_s), \theta_s-\theta^*\rangle\right|\\
%\leq &\left|\sum_{t=\tau_\epsilon}^{T-1} \langle g(\theta_{t},z_t), \eta_{t-\tau_\epsilon}\theta_{t-\tau_\epsilon}-\eta_t\theta_t\rangle\right|
%+\left|\sum_{t=0}^{\tau_\epsilon-1} \langle g(\theta_{t},z_t), \theta_t-\theta^*\rangle\right|+
%\left|\sum_{t=T}^{T+\tau_\epsilon-1} \langle g(\theta_{t},z_t), \theta_{t-\tau_\epsilon}-\theta^*\rangle\right|\\
\leq & \left|\E\sum_{s=\tau_\epsilon}^{t-1} \langle g(\theta_{s},z_s), \eta_{s}(\theta_{s-\tau_\epsilon}-\theta_s)\rangle\right| +   \left|\E\sum_{s=\tau_\epsilon}^{t-1} \langle g(\theta_{s},z_s), (\eta_{s-\tau_\epsilon}-\eta_s)(\theta_{s-\tau_\epsilon}-\theta^*)\rangle\right|\\
&+\left|\E\sum_{s=0}^{\tau_\epsilon-1} \eta_s\langle g(\theta_{s},z_s), \theta_s-\theta^*\rangle\right|+
\left|\E\sum_{s=t}^{t+\tau_\epsilon-1} \eta_{s-\tau_\epsilon}\langle g(\theta_{s},z_s), \theta_{s-\tau_\epsilon}-\theta^*\rangle\right|\\
\leq& \E\sum_{s=\tau_\epsilon}^{t-1}M^2V(z_s)\eta_s(V\eta)_{s-\tau_\epsilon:s}+M\E\sum_{s=0}^{t-\tau_{\epsilon}-1} \eta_{s}^3\tau_\epsilon\|\theta_s-\theta^*\|V(z_{s+\tau_\epsilon})\\
&\quad+M\E\sum_{s=0}^{\tau_\epsilon-1}\eta_{s}V(z_s)\|\theta_s-\theta^*\| + M\E\sum_{s=t-\tau_{\epsilon}}^{t-1}\eta_s V(z_{s+\tau_\epsilon})\|\theta_s-\theta^*\|
\end{align*}
% \red{Here we used that}
% \[
% \red{\eta_{t-\tau_\epsilon}-\eta_t
% =O(\frac{1}{\sqrt{t-\tau_\epsilon}}-\frac{1}{\sqrt{t}})=O(\frac{\tau_\epsilon}{(t-\tau_\epsilon)^{3/2}})}
% \]
%For this to hold, we need to assume $\eta_t=1/\sqrt{t}$ up front.
%In view of the blue term in ineq. next line, it seems we just need $\eta_t^2$? Another idea that not involve up front assumption, is to carry $\eta_{t}-\eta_{t+\tau_\epsilon}$ in the derivation below. 
%Using the same derivation on page 25, we should be able to obtain
Putting the bounds for (a) to (d) together, we have
\[\begin{split}
&\left|\E\sum_{s=0}^{t-1}\eta_s \langle \nabla \ell(\theta_s)-\ghat_s(\theta_s, z_s), \theta_s-\theta^*)\rangle\right|\\
\leq& \sum_{s=0}^{t-1}\eta_s\E e_s 
\E\|\theta_s-\theta^*\|
%+ML\tau_\epsilon \sum_{s=0}^{t-1}\eta_s^2\E(V(z_{s})+K)^2\|\theta_s-\theta^*\|
+M^2\tau_\epsilon\sum_{s=\tau_\epsilon}^{t-1}\eta_{s}^2\E(V(z_{s})+K)^2
\\
&+M\tau_{\epsilon}\sum_{s=1}^{t-1}\eta_s^3\E(V(z_{s})+K)^2\|\theta_s-\theta^*\|+M\sum_{s=0}^{\tau_\epsilon-1} \eta_s\E V(z_s)\|\theta_s-\theta^*\|\\
&+M\sum_{s=t-\tau_\epsilon}^{t-1} \eta_s\E V(z_s)\|\theta_s-\theta^*\|
+3L\epsilon \sum_{s=0}^{t-1}\eta_s \E (V(z_{s})+K)^2\|\theta_s-\theta^*\|\\
&+ \left(3\frac{ML^2K^2}{(1-\rho)^2} \tau_\epsilon+ML\tau_{\epsilon}\right)\sum_{s=0}^{t-1}\eta^2_{s} 
\E (V(z_{s})+K)^2\|\theta_s-\theta^*\|.
\end{split}\]
%\blue{This seems wrong with $V(z_{s-1})$, since $s$ starts with zero? Maybe we should start $s$ with 1?}

Next, we prove the result by induction. Suppose for $s\leq t$, 
\[
\E\|\theta_s-\theta^*\|^2
\leq C^2\left(\|\theta_0-\theta^*\|+\tau_{\epsilon}\log t+\epsilon\sqrt{s}+(V(z_0)+K)^2\right)^2.
\]
%\red{The induction should be on $\E\|\theta_s-\theta^*\|^2$?}
Then, since $\E (V(z_{s-1})+K)^2\|\theta_{s-1}-\theta^*\|\leq
\sqrt{\E (V(z_s)+K)^4}\sqrt{\E\|\theta_s-\theta^*\|^2}$
\[\begin{split}
&\left|\E\sum_{s=0}^{t-1}\eta_s \langle \nabla \ell(\theta_s)-\ghat_s(\theta_s, z_s), \theta_s-\theta^*)\rangle\right|\\
\leq& \sum_{s=0}^{t-1}\eta_s\E e_s C(\|\theta_0-\theta^*\|+\tau_{\epsilon}\log t+\epsilon\sqrt{s}+(V(z_0)+K)^2)
%+ML\tau_\epsilon \sum_{s=0}^{t-1}\eta_s^2C(\|\theta_0-\theta^*\|+\tau_{\epsilon}\log t+\epsilon\sqrt{s}+\blue{(V(z_0)+K)^2})\\
+M^2\tau_\epsilon\sum_{s=\tau_\epsilon}^{t-1}\eta_{s}^2 \E(V(z_{s})+K)^2\\
&+M\tau_{\epsilon}\sum_{s=1}^{t-1}\eta_s^3\sqrt{\E (V(z_{s})+K)^4}C(\|\theta_0-\theta^*\|+\tau_{\epsilon}\log t+\epsilon\sqrt{s}+(V(z_0)+K)^2)\\
&+M\sum_{s=0}^{\tau_\epsilon-1} \eta_s \sqrt{\E V(z_s)^2} C(\|\theta_0-\theta^*\|+\tau_{\epsilon}\log t+\epsilon\sqrt{s}+(V(z_0)+K)^2)\\
&+ M\sum_{s=t-\tau_\epsilon}^{t-1} \eta_s\sqrt{\E V(z_s)^2}C(\|\theta_0-\theta^*\|+\tau_{\epsilon}\log t+\epsilon\sqrt{s}+(V(z_0)+K)^2)\\
&+3L\epsilon \sum_{s=0}^{t-1}\eta_s\sqrt{\E (V(z_{s})+K)^4} C(\|\theta_0-\theta^*\|+\tau_{\epsilon}\log t+\epsilon\sqrt{s}+(V(z_0)+K)^2)\\ 
&+ \left(3\frac{ML^2K^2}{(1-\rho)^2} \tau_\epsilon+ML\tau_{\epsilon}\right)\sum_{s=0}^{t-1}\eta^2_{s}
\sqrt{\E (V(z_{s})+K)^4}C(\|\theta_0-\theta^*\|+\tau_{\epsilon}\log t+\epsilon\sqrt{s}+(V(z_0)+K)^2)\\
\leq & C'(\|\theta_0-\theta^*\|+\tau_{\epsilon}\log t+(V(z_0)+K)^2)C(\tau_{\epsilon}\log t + \epsilon\sqrt{t}+(V(z_0)+K)^2)
\end{split}\]
where $C'$ is a constant that does not depend on $C$.
%\red{We can use Cauchy Schwarz if we assume $\E V(z_s)^2\leq C^2(V(z_0)^2+1)$, which is similar to assuming $V(z_0)^2$ is also a Lyapunov function. Then 
%\[
%\E V(z_{s-1})\|\theta_s-\theta^*\|\leq
%\sqrt{\E V(z_s)^2}\sqrt{\E\|\theta_s-\theta^*\|^2}
%\]}
Thus,
\begin{align*}
\E\|\theta_{t}-\theta^*\|^2&\leq
\|\theta_0-\theta^*\|^2+M^2\log t(V(z_0)+K)^2\\
&+C'C(\|\theta_0-\theta^*\|+\tau_\epsilon \log t+(V(z_0)+K)^2)(\tau_\epsilon\log t+\epsilon\sqrt{t}+(V(z_0)+K)^2)\\
&\leq C^2(\|\theta_0-\theta^*\|+\tau_\epsilon \log t+\epsilon\sqrt{t}+(V(z_0)+K)^2)^2
\end{align*}
\Halmos
\endproof

\proof{Proof of Case 1 of Theorem \ref{thm:main2}.}
By the convexity of $\Theta$ and the fact that $\theta^*\in \Theta$, we have
\begin{align*}
\|\theta_{t+1}-\theta^*\|^2&\leq\|\theta_{t}-\theta^*-\eta_t \ghat_t(\theta_t, z_t) \|^2\\
&=\|\theta_{t}-\theta^*\|^2+2\eta_t \langle \ghat_t(\theta_t, z_t),\theta^*-\theta_t\rangle+\eta_t^2 \|\ghat_t(\theta_t, z_t)\|^2\\
&\leq \|\theta_{t}-\theta^*\|^2+2\eta_t \langle \ghat_t(\theta_t, z_t),\theta^*-\theta_t\rangle+\eta_t^2M^2V(z_t)^2.
\end{align*}
Therefore, 
\[
\langle \ghat_t(\theta_t, z_t),\theta^*-\theta_t\rangle\geq \frac{1}{2\eta_t}(\|\theta_{t+1}-\theta^*\|^2-\|\theta_{t}-\theta^*\|^2)-\frac12\eta_tM^2 V(z_t)^2. 
\]
Next, by the convexity of $\ell$, 
\begin{align*}
\ell(\theta^*)-\ell(\theta_t)&\geq \langle \nabla \ell(\theta_t), \theta^*-\theta_t\rangle\\
&=\langle \ghat_t(\theta_t, z_t), \theta^*-\theta_t\rangle+\langle \nabla \ell(\theta_t)-\ghat_t(\theta_t, z_t), \theta^*-\theta_t\rangle\\
&\geq  \frac{1}{2\eta_t}(\|\theta_{t+1}-\theta^*\|^2-\|\theta_{t}-\theta^*\|^2)-\frac12\eta_t V(z_t)^2M^2
+\langle \nabla \ell(\theta_t)-\ghat_t(\theta_t, z_t), \theta^*-\theta_t\rangle.
\end{align*}
This leads to 
%\begin{align*}
%\sum_{t=0}^{T-1} l(\theta_t)-l(\theta^*)
%\leq& \frac{1}{2\eta_0}\|\theta_{0}^2-\theta^*\|^2+\sum_{t=1}^{T-1}\left(\frac{1}{2\eta_t}-\frac{1}{2\eta_{t-1}}\right)\|\theta_{t}^2-\theta^*\|^2+\frac12 M^2\eta_{0:T}\\
%&+\sum_{t=0}^{T-1} \langle \bar g(\theta_t)-\ghat_t(\theta_t, z_t), \theta_t-\theta^*\rangle\\
%\leq& \frac{1}{\eta_T} C^2+\frac12 M^2\eta_{0:T} +\sum_{t=0}^{T-1} 
%\langle \bar g(\theta_t)-\ghat_t(\theta_t, z_t), \theta_t-\theta^*\rangle. 
%\end{align*}
\begin{align*}
\frac{1}{\eta_{0:T}}\sum_{t=0}^{T-1} \eta_t(\ell(\theta_t)-\ell(\theta^*))
\leq& \frac{1}{2\eta_{0:T}}\|\theta_{0}-\theta^*\|^2+\frac12 M^2\frac{(V\eta)_{0:T}^2}{\eta_{0:T}}\\
&+\frac{1}{\eta_{0:T}}\sum_{t=0}^{T-1} \eta_t\langle \nabla \ell(\theta_t)-\ghat_t(\theta_t, z_t), \theta_t-\theta^*\rangle
\end{align*}

From the proof of Lemma \ref{lem:thetabound}, for $\epsilon=1/\sqrt{T}$, $\tau_{\epsilon}=O(\tau\log T)$, we have
\[\begin{split}
&\left|\E\sum_{t=0}^{T-1}\eta_t \langle \nabla \ell(\theta_t)-\ghat_t(\theta_t, z_t), \theta_t-\theta^*\rangle\right|\\
%\leq& \sum_{t=0}^{T-1}\eta_s^2
%\E\|\theta_t-\theta^*\|
%+ML\tau_\epsilon \sum_{t=0}^{T-1}\eta_t^2\E\|\theta_t-\theta^*\|
%+M^2\tau_\epsilon\sum_{t=\tau_\epsilon}^{T-1}\eta_{t}^2
%+\tau_{\epsilon}\sum_{t=1}^{T-1}\eta_t^3\E\|\theta_t-\theta^*\|\\
%&+M\sum_{t=0}^{\tau_\epsilon-1} \eta_t\E\|\theta_t-\theta^*\|+
%M\sum_{t=T-\tau_\epsilon}^{T-1} \eta_t\E\|\theta_t-\theta^*\|\\
%&+3L\epsilon \sum_{t=0}^{T-1}\eta_t\E V(z_{t-1}) \|\theta_t-\theta^*\|
%+ 3\frac{ML^2K^2}{(1-\rho)^2} \tau_\epsilon\sum_{t=0}^{T-1}\eta^2_{t}
%\E V(z_{t-1})\|\theta_s-\theta^*\|\\
\leq& C'(\|\theta_0-\theta^*\|+\tau_{\epsilon}\log T+(V(z_0)+K)^2)C(\tau_{\epsilon}\log T + \epsilon\sqrt{T}+(V(z_0)+K)^2).
\end{split}\]
%\blue{it's a mess here. Can you clean it up?}
Then,
\begin{align*}
&\frac{1}{\eta_{0:T}}\sum_{t=0}^{T-1} \eta_t(\ell(\theta_t)-\ell(\theta^*))\\
\leq& \frac{\|\theta_0-\theta^*\|^2}{\eta_{0:T}} + \frac{M^2}{2\eta_{0:T}}\sum_{t=0}^{T-1}\eta_t^2\E V(z_t)^2\\
&+\frac{C'C}{\eta_{0:T}}(\|\theta_0-\theta^*\|+\tau_{\epsilon}\log T+(V(z_0)+K)^2)(\tau_{\epsilon}\log T + \epsilon\sqrt{T}+(V(z_0)+K)^2)\\
=&O\left(\frac{1}{\sqrt{T}}+\frac{\log T}{\sqrt{T}} + \frac{\tau(\log T)^2}{\sqrt{T}} + \frac{\tau^2(\log T)^4}{\sqrt{T}}\right)
\end{align*}

% \begin{align*}
% &\frac{1}{\eta_{0:T}}\sum_{t=0}^{T-1} \eta_t(l(\theta_t)-l(\theta^*))\\
% \leq& \frac{C}{\eta_{0:T}}+\left(\frac{M^2}{2} + M^2\tau_\epsilon 
%  + ML\tau_\epsilon C(\log T)^2+C(\log T)^2\right)\frac{\eta_{0:T}^2}{\eta_{0:T}}
% %+\frac{C(\log T)^2}{\eta_{0:T}}\sum_{t=0}^{T-1}\eta_t\E e_t\\
% + \frac{3MC(\log T)^2}{\eta_{0:T}}\tau_\epsilon \\
% &+\frac{3L\epsilon C(\log T)^2}{\eta_{0:T}} \sum_{t=0}^{T-1}\eta_t\E V(z_{t-1})
% +\frac{3L^2K^2MC(\log T)^2}{\eta_{0:T}(1-\rho)^2}\tau_{\epsilon}\sum_{t=0}^{T-1}\eta_{t}^2 \E V(z_{t-1})
% \end{align*}
%\[\begin{split}
%\frac{1}{T} \sum_{t=1}^T \E l(\theta_t)-l(\theta^*)
%\leq& \frac{1}{T\eta_T} C^2+\frac{1}{2T} M^2\eta_{0:T}
%+2C\frac{1}{T}\sum_{t=0}^{T-1}\E e_t + M^2\tau_\epsilon\frac{\eta_{0:T}}{T}\\
%&+4MC\frac{\tau_\epsilon}{T} + 2CML\frac{\tau_\epsilon \eta_{0:T}}{T}
%+6CL\epsilon \frac{1}{T}\sum_{t=0}^{T-1}\E V(z_{t-1})\\
%&+6C\frac{L^2K^2M}{(1-\rho)^2}\frac{\tau_{\epsilon}}{T}\sum_{t=0}^{T-1}\eta_{t} \E V(z_{t-1})
%\end{split}\]
%For $\epsilon=1/\sqrt{T}$, $\tau_{\epsilon}=O(\tau\log T)$, we have
% \[
% \frac{1}{\eta_{0:T}}\sum_{t=0}^{T-1} \eta_t(l(\theta_t)-l(\theta^*))=O\left(\frac{1}{\eta_{0:T}}\left(\tau\log T+ \tau\log T\eta_{0:T}^2+\frac{1}{\sqrt{T}}\eta_{0:T}+\sum_{t=1}^T\eta_t\E e_t\right)\right).
% \]
% Since $\eta_t=\eta_0/\sqrt{t}$, we have
% \[
% \frac{1}{\eta_{0:T}}\sum_{t=0}^{T-1} \eta_t(l(\theta_t)-l(\theta^*))=O\left(\tau(\log T)^4/\sqrt{T}\right)
% \]
%\[
%\frac{1}{T} \sum_{t=1}^T \E l(\theta_t)-l(\theta^*)=O\left(\frac{1}{T}\left(\tau\log T\eta_{0:T}+\frac{1}{\eta_T}+\sum_{t=1}^T\E e_t\right)\right).
%\]
For $\bar\theta_T=\frac{1}{\eta_{0:T}}\sum_{t=0}^{T-1}\eta_t\theta_t$, by the convexity of $\ell$, we have 
\[
\E \ell(\bar{\theta}_T)-\ell(\theta^*)\leq
\frac{1}{\eta_{0:T}}\sum_{t=0}^{T-1} \eta_t(\ell(\theta_t)-\ell(\theta^*))
=O\left(\tau^2(\log T)^4/\sqrt{T}\right).
\]
%O\left(\frac{1}{\eta_{0:T}}\left(\tau\log T+ \tau\log T\eta_{0:T}^2+\frac{1}{\sqrt{T}}\eta_{0:T}+\sum_{t=1}^T\eta_t\E e_t\right)\right).
%\end{split}\]
%\[
%\E l(\bar{\theta}_T)-l(\theta^*)\leq\frac{1}{T} \sum_{t=1}^T \E l(\theta_t)-l(\theta^*)
%=O\left(\frac{1}{T}\left((\tau\log T)\eta_{0:T}+\frac{1}{\eta_T}+\sum_{t=1}^T\E e_t\right)\right).
%\]
\Halmos
\endproof

We next present an auxiliary result about our choice of step size in the strongly convex case.
\begin{lemma} \label{lem:stepsize}
If $\eta_t=2\eta_0/(ct)$ and $\eta_0>2$, we have $\exp(-\frac12c\eta_{t+1:T})\eta_t$ is increasing in $t$.  
%If $\eta_t=8\eta_0/(c(t+1))$, we have $\exp(-\frac12c\eta_{t+1:T})\eta_t^2$ is increasing in $t$. 
\end{lemma}
\proof{Proof.}
Through induction, we only need to show that 
$\exp\left(-\frac12 c\eta_{t+1}\right)\eta_t\leq \eta_{t+1}$,
which is equivalent to showing 
\[
\exp(-\frac12 c\eta_{t+1})\leq \frac{t}{t+1},
\]
which is true with $\eta_{t+1}\geq \frac{4}{c(t+1)}$.
\Halmos
\endproof

\proof{Proof of Case 2 in Theorem \ref{thm:main2}.}
First, note that
\[\begin{split}
\|\theta_{t+1}-\theta^*\|^2 &= \|\theta_t - \eta_t\ghat(\theta_t,z_t)-\theta^*\|^2\\
&\leq \|\theta_t-\theta^*\|^2-2\langle \theta_t-\theta^*,\ghat(\theta_t,z_t)\rangle\eta_t+M^2\eta_t^2V(z_t)^2.
\end{split}\]

Consider the following decomposition of $\ghat(\theta_t,z_t)$:
\begin{equation}\label{eq:g_decomp}
\begin{split}
\ghat(\theta_t,z_t)=&\underbrace{[\ghat(\theta_t,z_t)-g(\theta_t,z_t)]}_{(a)}+\underbrace{[g(\theta_t,z_t)-g(\theta_{t+\tau_\epsilon},z_{t+\tau_\epsilon})]}_{(b)}+
\underbrace{[g(\theta_{t+\tau_\epsilon},z_{t+\tau_\epsilon})-g(\theta_{t},z_{t+\tau_\epsilon})]}_{(c)}\nonumber\\
&+\underbrace{[g(\theta_{t},z_{t+\tau_\epsilon})-\nabla \ell(\theta_t)]}_{(d)}+\underbrace{\nabla \ell(\theta_t)}_{(e)}. 
\end{split}
\end{equation}
We next bound the inner product of each part in \eqref{eq:g_decomp} with $\theta^*-\theta_t$, except for part (b). Part (b) will be treated separately later.

For (e), since $\ell$ is strongly convex with convexity constant $c$
\[
-\langle\nabla \ell(\theta_t),\theta_t-\theta^*\rangle\leq -c\|\theta_t-\theta^*\|^2. 
\]
For (a), under Assumption \ref{ass:error0},
\[\begin{split}
-\langle \ghat(\theta_t,z_t)-g(\theta_t,z_t), \theta_t-\theta^*\rangle
&\leq \frac{1}{c}\|\ghat(\theta_t,z_t)-g(\theta_t,z_t)\|^2+\frac{1}{4}c \|\theta_t-\theta^*\|^2\\ 
&\leq\frac{1}{c}e_t^2+\frac{1}{4}c \|\theta_t-\theta^*\|^2.
\end{split}\]
For (c), 
%we first note that by Lemma \ref{lem:lip0}, 
%$\|\theta_{t+\tau_\epsilon}-\theta_t\|\leq M\eta_{t:(t+\tau_\epsilon)}\leq M\tau_{\epsilon}\eta_t$.
%%since $\eta_t$ is decreasing in $t$.
following Lemma \ref{lem:lip}, we have
\[\begin{split}
&-\E_t\langle g(\theta_{t+\tau_\epsilon},z_{t+\tau_\epsilon})-g(\theta_{t},z_{t+\tau_\epsilon}),\theta_t-\theta^*\rangle\\
\leq& \frac1c\|\E_t[g(\theta_{t+\tau_\epsilon},z_{t+\tau_\epsilon})-g(\theta_{t},z_{t+\tau_\epsilon})]\|^2+\frac14c\|\theta_t-\theta^*\|^2\\
\leq& \frac1c \left(2L\epsilon+ 2\frac{L^2K^2M}{(1-\rho)^2} \eta_{t:t+\tau_\epsilon}+ML\eta_{t:t+\tau_\epsilon}\right)^2 (V(z_{t-1})+K)^4 +\frac14c\|\theta_t-\theta^*\|^2. 
\end{split}\]
%\blue{Use the new bound, which looks like $(d)$ }
For (d), following Lemma \ref{lm:mix}, we have
\[\begin{split}
&-\E_t\langle g(\theta_{t},z_{t+\tau_\epsilon})-\nabla \ell(\theta_t),\theta_t-\theta^*\rangle\\
\leq& \frac1c\| \E_tg(\theta_{t},z_{t+\tau_\epsilon})-\nabla \ell(\theta_t)\|^2 +\frac14c\|\theta_t-\theta^*\|^2\\
\leq& \frac1c \left(L\epsilon+ \frac{L^2K^2M}{(1-\rho)^2} \eta_{t:t+\tau_\epsilon}\right)^2 (V(z_{t-1})+K)^4+
\frac14c\|\theta_t-\theta^*\|^2\\
\leq & \frac{2}{c}L^2\epsilon^2(V(z_{t-1})+K)^4+\frac{2}{c}\frac{L^4K^4M^2}{(1-\rho)^4}\tau_{\epsilon}^2\eta_t^2(V(z_{t-1})+K)^4 + \frac14c\|\theta_t-\theta^*\|^2
\end{split}\]
Putting together the bounds for parts (a), (c), (d), and (e), we have 
\[\begin{split}
\E_t\|\theta_{t+1}-\theta^*\|^2\leq& \left(1-\frac{1}{2}c\eta_t\right)\|\theta_{t}-\theta^*\|^2 + M^2\eta_t^2V(z_t)^2+\frac{2}{c}\eta_te_t^2\\
&+\frac{2}{c}\left(18L^2\epsilon^2\eta_t + 2L^2M^2\tau_{\epsilon}^2\eta_t^3 + 18\frac{L^4K^4M^2}{(1-\rho)^4}\tau_{\epsilon}^2\eta_t^3\right)(V(z_{t})+K)^4\\
&-2\eta_t\E_t\langle g(\theta_t,z_t)-g(\theta_{t+\tau_\epsilon},z_{t+\tau_\epsilon}),\theta_t-\theta^*\rangle.
\end{split}\]
Define $\eta_{T:T}=0$. Then, 
%\red{(we also have a $E[V(z_t)^2])$ down there}
\begin{align*}
\E\|\theta_{T}-\theta^*\|^2\leq &\exp\left(-\frac{1}{2}c\eta_{0:T}\right)\|\theta_{0}-\theta^*\|^2 + \sum_{t=0}^{T-1}\exp\left(-\frac{1}{2}c\eta_{(t+1):T}\right)M^2\eta_t^2V(z_t)^2\\
&+\frac2c \sum_{t=0}^{T-1}\exp\left(-\frac{1}{2}c\eta_{(t+1):T}\right)(18L^2\epsilon^2\eta_t\E(V(z_{t})+K)^4+\eta_t \E e_t^2)\\
&+\frac2c\sum_{t=0}^{T-1}\exp\left(-\frac{1}{2}c\eta_{(t+1):T}\right)\left(2L^2M^2\tau_{\epsilon}^2\eta_t^3 + 18\frac{L^4K^4M^2}{(1-\rho)^4}\tau_{\epsilon}^2\eta_t^3\right)\E(V(z_{t})+K)^4\\
&\underbrace{-2\sum_{t=0}^{T-1} \exp\left(-\frac{1}{2}c\eta_{(t+1):T}\right)\eta_t \E\langle g(\theta_{t},z_{t})-g(\theta_{t+\tau_\epsilon},z_{t+\tau_\epsilon}),\theta_{t}-\theta^*\rangle}_{(f)}.
\end{align*}
Lastly, we develop a proper bound for (f). 
We first re-arrange the summation as
\[\begin{split}
&-2\underbrace{\sum_{t=\tau_\epsilon}^{T-1}\left\langle \exp\left(-\frac{1}{2}c\eta_{(t+1):T}\right)\eta_t(\theta_t-\theta^*)- \exp\left(-\frac{1}{2}c\eta_{(t+1-\tau_\epsilon):T}\right)\eta_{t-\tau_\epsilon}(\theta_{t-\tau_\epsilon}-\theta^*),g(\theta_t,z_t)\right\rangle}_{(f1)}\\
&-2\underbrace{\sum_{t=0}^{\tau_\epsilon-1} \exp\left(-\frac{1}{2}c\eta_{(t+1):T}\right)\eta_t\langle g(\theta_{t},z_{t}),\theta_{t}-\theta^*\rangle}_{(f2)}\\
&+2\underbrace{\sum_{t=T-\tau_{\epsilon}}^{T-1}\exp\left(-\frac{1}{2}c\eta_{(t+1):T}\right)\eta_t\langle g(\theta_{t+\tau_{\epsilon}},z_{t+\tau_{\epsilon}}),\theta_{t}-\theta^*\rangle}_{(f3)}
\end{split}\]
For (f1), we have
\begin{align*}
&\left|-\left\langle \exp\left(-\frac{1}{2}c\eta_{(t+1):T}\right)\eta_t(\theta_t-\theta^*)- \exp\left(-\frac{1}{2}c\eta_{(t+1-\tau_\epsilon):T}\right)\eta_{t-\tau_\epsilon}(\theta_{t-\tau_\epsilon}-\theta^*),g(\theta_t,z_t)\right\rangle\right|\\
\leq &\left|\exp\left(-\frac{1}{2}c\eta_{(t+1):T}\right)\eta_t\langle \theta_t-\theta_{t-\tau_\epsilon}, g(\theta_t,z_t)\rangle\right|\\
&+\left|\left\langle\left(\exp\left(-\frac{1}{2}c\eta_{(t+1):T}\right)\eta_t- \exp\left(-\frac{1}{2}c\eta_{(t+1-\tau_\epsilon):T}\right)\eta_{t-\tau_\epsilon}\right)(\theta_{t-\tau_\epsilon}-\theta^*), g(\theta_t,z_t)\right\rangle\right|
\\
%\leq &\left|-\exp\left(-\frac{1}{2}c\eta_{(t+1):T}\right)\eta_t\langle \theta_t-\theta_{t-\tau_\epsilon}, g(\theta_t,s_t)\rangle\right|\\
%&+\left(\exp\left(-\frac{1}{2}c\eta_{(t+1):T}\right)\eta_t- \exp\left(-\frac{1}{2}c\eta_{(t+1-\tau_\epsilon):T}\right)\eta_{t-\tau_\epsilon}\right) 2CM \\
\leq& 
M^2\exp\left(-\frac{1}{2}c\eta_{(t+1):T}\right)\eta_tV(z_t)(V\eta)_{t-\tau_{\epsilon}:t}\\
%M^2\tau_\epsilon \log \tau_\epsilon \exp\left(-\frac{1}{2}c\eta_{(t+1):T}\right)\eta^2_{t}\\
&+\left(\exp\left(-\frac{1}{2}c\eta_{(t+1):T}\right)\eta_t- \exp\left(-\frac{1}{2}c\eta_{(t+1-\tau_\epsilon):T}\right)\eta_{t-\tau_\epsilon}\right) \|\theta_{t-\tau_\epsilon}-\theta^*\|MV(z_t).
\end{align*}
%where the last step follows from the fact that 
%$\|g\|\leq M$ and 
%by Lemma \ref{lem:lip0},
%\[\begin{split}
%\|\theta_t-\theta_{t-\tau_\epsilon}\|\leq& M\eta_{t-\tau_\epsilon:t}\\
%\leq &M\exp\left(\frac{1}{2}c\eta_{(t+1-\tau_\epsilon):t}\right)\eta_t \tau_\epsilon\\
%\leq& M{\color{blue}\left(\frac{t}{t-\tau_{\epsilon}+1}\right)^{\eta_0}\eta_t \tau_\epsilon \leq  M\tau_{\epsilon}^{1+\eta_0} \eta_t.}
%\end{split}\]
%\blue{We might want to use a large $\eta_0$, so we cannot use this relation ship. Instead}
Since,
\[
\eta_{t-\tau_\epsilon:t}
\leq \frac{\eta_{1:\tau_\epsilon}}{\eta_{\tau_\epsilon}} \eta_t 
\leq \tau_\epsilon \log \tau_\epsilon  \eta_t.
\]
\[\begin{split}
&M^2\exp\left(-\frac{1}{2}c\eta_{(t+1):T}\right)\E\eta_tV(z_t)(V\eta)_{t-\tau_{\epsilon}:t} \\
\leq& M^2\exp\left(-\frac{1}{2}c\eta_{(t+1):T}\right)C(V(z_0)^2+K)\tau_\epsilon \log \tau_\epsilon  \eta_t^2 
\end{split}\]
Next, note that
\begin{align*}
  &\sum_{t=\tau_{\epsilon}}^{T-1}\left(\exp\left(-\frac{1}{2}c\eta_{(t+1):T}\right)\eta_t- \exp\left(-\frac{1}{2}c\eta_{(t+1-\tau_\epsilon):T}\right)\eta_{t-\tau_\epsilon}\right) \E\|\theta_{t-\tau_\epsilon}-\theta^*\|MV(z_t)\\
  \leq& M\sum_{t=\tau_{\epsilon}}^{T-1}\left(\exp\left(-\frac{1}{2}c\eta_{(t+1):T}\right)\eta_t-\exp\left(-\frac{1}{2}c\eta_{(t+1-\tau_\epsilon):T}\right)\eta_{t-\tau_\epsilon}\right) \sqrt{\E\|\theta_{t-\tau_\epsilon}-\theta^*\|^2}\sqrt{\E V(z_t)^2}\\
    \leq&MC(V(z_0)^2+1)\left\{\sum_{t=\tau_\epsilon}^{T-\tau_\epsilon-1}\left(\sqrt{\E\|\theta_{t-\tau_\epsilon}-
    \theta^*\|^2}-\sqrt{\E\|\theta_t-\theta^*\|^2}\right)\exp\left(-\frac{1}{2}c\eta_{(t+1):T}\right)\eta_t \right.\\
    &\quad\left. -\sum_{t=1}^{\tau_\epsilon-1} \exp\left(-\frac{1}{2}c\eta_{(t+1):T}\right)\eta_t\sqrt{\E\|\theta_t-\theta^*\|^2}
    +\sum_{t=T-\tau_\epsilon}^{T-1}\exp\left(-\frac{1}{2}c\eta_{(t+1):T}\right)\eta_t \sqrt{\E\|\theta_{t-\tau_\epsilon}-\theta^*\|^2}\right\} \\
    \leq& MC(V(z_0)^2+1)
    \left\{\sum_{t=\tau_\epsilon}^{T-\tau_\epsilon-1}\eta_T\sqrt{\E\|\theta_{t-\tau_\epsilon}-\theta_{t}\|^2}
    -\sum_{t=1}^{\tau_\epsilon-1} \exp\left(-\frac{1}{2}c\eta_{(t+1):T}\right)\eta_t\sqrt{\E\|\theta_t-\theta^*\|^2}\right.\\
    &\quad \left.+\sum_{t=T-\tau_\epsilon}^{T-1}\eta_T \sqrt{\E \|\theta_{t-\tau_\epsilon}-\theta^*\|^2}\right\} \mbox{ since $\exp\left(-\frac{1}{2}c\eta_{t+1:T}\right)\eta_t\leq \eta_T$}\\
    \leq&MC(V(z_0)^2+1)
    \left\{\eta_T\sum_{t=\tau_\epsilon}^{T-\tau_\epsilon-1}\sqrt{\E(M(V\eta)_{t-\tau_\epsilon:t})^2}
    -\sum_{t=1}^{\tau_\epsilon-1} \exp\left(-\frac{1}{2}c\eta_{(t+1):T}\right)\eta_t\sqrt{\E\|\theta_t-\theta^*\|^2}\right.\\
    &\quad \left.+\eta_T\sum_{t=T-\tau_\epsilon}^{T-1} \sqrt{\E \|\theta_{t-\tau_\epsilon}-\theta^*\|^2}\right\}\\
    =&MC(V(z_0)^2+1)
    \left\{MC\tau_{\epsilon}\eta_T\eta_{0:T-2\tau_{\epsilon}}(V(z_0)^2+1)^{1/2}
     -\sum_{t=1}^{\tau_\epsilon-1} \exp\left(-\frac{1}{2}c\eta_{(t+1):T}\right)\eta_t\sqrt{\E\|\theta_t-\theta^*\|^2}\right.\\
    &\quad \left.+\eta_T\sum_{t=T-\tau_\epsilon}^{T-1} \sqrt{\E \|\theta_{t-\tau_\epsilon}-\theta^*\|^2}\right\}.
\end{align*}
For (f2), we have
\[\begin{split}
&\left|-\sum_{t=0}^{\tau_\epsilon-1} \exp\left(-\frac{1}{2}c\eta_{(t+1):T}\right)\eta_t\E\langle g(\theta_{t},z_{t}),\theta_{t}-\theta^*\rangle\right|\\
\leq& M\sum_{t=0}^{\tau_\epsilon-1}\exp\left(-\frac{1}{2}c\eta_{(t+1):T}\right)\eta_t\E V(z_t)\|\theta_t-\theta^*\|\\
\leq& M\sum_{t=0}^{\tau_\epsilon-1}\exp\left(-\frac{1}{2}c\eta_{(t+1):T}\right)\eta_t\sqrt{\E V(z_t)^2}\sqrt{\E\|\theta_t-\theta^*\|^2}\\
\leq &MC(V(z_0)^2+1)\sum_{t=1}^{\tau_\epsilon-1} \exp\left(-\frac{1}{2}c\eta_{(t+1):T}\right)\eta_t\sqrt{\E\|\theta_t-\theta^*\|^2}.
\end{split}\]
Similarly, for (f3), we have
\[\begin{split}
&\left|\sum_{t=T-\tau_{\epsilon}}^{T-1}\exp\left(-\frac{1}{2}c\eta_{(t+1):T}\right)\eta_t\E\langle g(\theta_{t+\tau_{\epsilon}},z_{t+\tau_{\epsilon}}),\theta_{t}-\theta^*\rangle\right|\\
% \leq& \sum_{t=T-\tau_\epsilon}^{T-1}\|\theta_t-\theta^*\|M\exp\left(-\frac{1}{2}c\eta_{(t+1):T}\right)\eta_t\\
\leq&MC(V(z_0)^2+1)\eta_T\sum_{t=T-\tau_{\epsilon}}^{T-1}\sqrt{\E\|\theta_t-\theta^*\|^2}.
\end{split}\]
Putting the bounds of (f1) -- (f3) together, we have
\begin{align*}
&\E\|\theta_{T}-\theta^*\|^2\\
\leq &\exp\left(-\frac{1}{2}c\eta_{0:T}\right)\|\theta_{0}-\theta^*\|^2\\
&+ \sum_{t=0}^{T-1}\exp\left(-\frac{1}{2}c\eta_{(t+1):T}\right)\left(M^2\eta_t^2\E V(z_t)^2+M^2C(V(z_0)^2+K)\tau_{\epsilon}\log\tau_{\epsilon}\eta_t^2\right)\\
&+\frac2c \sum_{t=0}^{T-1}\exp\left(-\frac{1}{2}c\eta_{(t+1):T}\right)(18L^2\epsilon^2\eta_t\E(V(z_{t})+K)^4+\eta_t \E e_t^2)\\
&+\frac2c\sum_{t=0}^{T-1}\exp\left(-\frac{1}{2}c\eta_{(t+1):T}\right)\left(2L^2M^2\tau_{\epsilon}^2\eta_t^3 + 18\frac{L^4K^4M^2}{(1-\rho)^4}\tau_{\epsilon}^2\eta_t^3\right)\E(V(z_{t})+K)^4\\
&+2M^2C^2(V(z_0)^2+1)^2\tau_{\epsilon}\eta_T\eta_{0:T-2\tau_{\epsilon}}+2M\eta_T\sum_{t=T-2\tau_{\epsilon}}^{T-1}\sqrt{\E\|\theta_t-\theta^*\|^2}.
\end{align*}
Since 
$\eta_t=2\eta_0/(ct)$ for some $\eta_0>0$, $\exp\left(-\frac{1}{2}c\eta_{(t+1):T}\right)\leq \left(\frac{t}{T}\right)^2$. In addition, since $\E e_t^2=O(1/t)$, we have
\begin{align*}
\E\|\theta_{T}-\theta^*\|^2
\leq & \frac{1}{T^2}\|\theta_0-\theta^*\|^2\\
&+\sum_{t=0}^{T-1}\frac{t^2}{T^2}\frac{4\eta_0^2}{c^2t^2}\left(M^2\E V(z_t)^2+M^2C(V(z_0)^2+K)\tau_{\epsilon}\log\tau_{\epsilon}\right)\\
&+\frac2c \sum_{t=0}^{T-1}\frac{t^2}{T^2}\frac{2\eta_0}{ct}(18L^2\epsilon^2\E(V(z_{t})+K)^4+\E e_t^2)\\
&+\frac2c\sum_{t=0}^{T-1}\frac{t^2}{T^2}\frac{8\eta_0^3}{c^3t^3}\left(2L^2M^2\tau_{\epsilon}^2 + 18\frac{L^4K^4M^2}{(1-\rho)^4}\tau_{\epsilon}^2\right)\E(V(z_{t})+K)^4\\
&+8M^2C^2(V(z_0)^2+1)^2\tau_{\epsilon}\frac{\eta_0^2}{c^2T}\log T+2M\eta_T\sum_{t=T-2\tau_{\epsilon}}^{T-1}\sqrt{\E\|\theta_t-\theta^*\|^2}\\
\leq& \frac{1}{T^2}\|\theta_0-\theta^*\|^2
+C_1\tau_{\epsilon}\frac{\log T}{T}
+C_2\epsilon^2
+C_3\tau_{\epsilon}\frac{\log T}{T}
+C_4\frac{1}{T}\sum_{t=T-2\tau_{\epsilon}}^{T-1}\sqrt{\E\|\theta_t-\theta^*\|^2},
\end{align*}
where $C_1, C_2, C_3, C_4$ are some suitably defined constants that do not depend on $T$.

Next, we prove by induction. Suppose for $t<T$,
\[
\E\|\theta_t-\theta^*\|^2 \leq C\left(\tau_{\epsilon}\frac{\log t}{t}+\epsilon^2\right).
\]
Then 
\[
\E\|\theta_{T}-\theta^*\|^2
\leq C\left(\tau_{\epsilon}\frac{\log T}{T}+\epsilon^2\right).
\]

Lastly, set $\epsilon=\frac{1}{\sqrt{T}}$. Then, $\tau_{\epsilon}=O(\tau\log T)$ and
\[
\E\|\theta_{T}-\theta^*\|^2=O\left(\tau\frac{(\log T)^2}{T}\right).
\]

\Halmos
\endproof

\section{Proofs of the results in Section~\ref{sec:inventory}}

\subsection{\label{sec:demand-derivative-proof}Proof of Lemma \ref{lem:demand-derivative}}
% We can express the dynamics of $D_{t}$ as $D_{t+1}=\phi(D_{t},\eta_{t+1},\epsilon_{t+1};\theta)$
% where $\phi(a,b,c;\theta)=(\alpha\min\{a+b,\theta\}+(1-\alpha)m+c)^{+}$.
% The derivative process $L_{t}=\frac{d}{d\theta}D_{t}$ itself follows
% the following dynamics

% \begin{align}
% L_{t+1}=\frac{dD_{t+1}}{d\theta} & =\frac{\partial}{\partial D_{t}}\phi(D_{t},\eta_{t+1},\epsilon_{t+1};\theta)\frac{dD_{t}}{d\theta}+\frac{\partial}{\partial\theta}\phi(D_{t},\eta_{t+1},\epsilon_{t+1};\theta)\nonumber \\
%  & =\alpha1\{D_{t+1}>0\}1\{D_{t}+\eta_{t+1}\leq\theta\}\frac{dD_{t}}{d\theta}+\alpha1\{D_{t+1}>0\}1\{\theta<D_{t}+\eta_{t+1}\}\nonumber \\
%  & =1\{D_{t+1}>0\}\alpha\left(1\{\theta<D_{t}+\eta_{t+1}\}+1\{D_{t}+\eta_{t+1}\leq\theta\}\frac{dD_{t}}{d\theta}\right)\label{eq:grad_recursion}
% \end{align}

Our proof is based on the development in \cite{glasserman1992stationary}, who establish sufficient conditions for the derivative process to be well-defined, converge to a unique stationary distribution, and be an unbiased estimator of $\frac{d}{d\theta}\mathbb{E}[D_{\infty}(\theta)]$ at stationarity.
%existence of a stationary distribution, and unbiasedness of the stationary derivative estimator.

Let 
\[
\phi(D_t, u_{t+1}, \epsilon_{t+1};\theta)=(\alpha\min\{D_{t}+u_{t+1},\theta\}+(1-\alpha)m+\epsilon_{t+1})^{+}
\]
Then, $D_{t+1}=\phi(D_t, u_{t+1}, \epsilon_{t+1};\theta)$, where $u_{t}$'s and $\epsilon_t$'s are iid respectively, and
\[
L_{t+1}=1\{D_{t+1}>0\}\alpha\left(1\{\theta<D_{t}+u_{t+1}\}+1\{D_{t}+u_{t+1}\leq\theta\}L_t\right).
\]
Let $G_{\phi}\subseteq\mathbb{R}^{4}$ be defined as the set where
$\phi$ is continuously differentiable. It follows that
$G_{\phi}$ is the complement $\mathbb{R}^{4}\backslash C_{\phi}$
where $C_{\phi}$ is 
\[
C_{\phi} =\{a,b,c,\theta:\alpha\min\{a+b,\theta\}+(1-\alpha)m+c=0\}\cup\{a,b,c,\theta:a+b=\theta\}.
\]

Note that for any $\theta$ and $t$, $\mathbb{P}((D_{t},u_{t+1},\epsilon_{t+1},\theta)\in G_{\phi})=1$
since $u_{t+1}$ and $\epsilon_{t+1}$ have densities on $\mathbb{R}$.
By Lemma 2.1 of \cite{glasserman1992stationary}, it follows that $D_{t}(\theta)$ is differentiable with probability 1
for any $\theta\in\Theta$ and $t\geq0$, and so the derivative process $L_{t}$ exists.

Moreover, note that $\phi$ is Lipschitz in $a,b,c,\theta$ with Lipschitz constant
$1$ and for all $t$,
\begin{align*}
\mathbb{E}\left[D_{t}\right] & =\mathbb{E}\left[(\alpha\min\{D_{t-1}+u_{t+1},\theta\}+(1-\alpha)m+\epsilon_{t+1})^{+}\right]\\
 & \leq\mathbb{E}\left[(\alpha\theta+(1-\alpha)m+\epsilon_{t+1})^{+}\right]<\infty.
\end{align*}
By Lemma 2.3 of \cite{glasserman1992stationary}, we have that $\mathbb{E}[D_{t}(\theta)]$ is differentiable
in $\theta$ for all $t$ and $\frac{d}{d\theta}\mathbb{E}[D_{t}(\theta)]=\mathbb{E}[\frac{d}{d\theta}D_{t}(\theta)]:=\mathbb{E}[L_{t}(\theta)]$.

Finally, it remains to show that $\frac{d}{d\theta}\mathbb{E}[D_{\infty}(\theta)]=\mathbb{E}[L_{\infty}(\theta)]$.
We first verify the conditions in Theorem 4.1 of \cite{glasserman1992stationary}, which guarantees
that $L_t$ has a stationary distribution.
First, we have that $(\epsilon_{t},\eta_{t})$ is a stationary sequence
as they are iid. Second, for any $\theta\in\Theta$, the stationary
distribution of $D_{t}(\theta)$ exists (see the proof of Theorem \ref{thm:newsvendor-conv})
and we can consider a stationary process $\widetilde{D}_{t}$ where
$\widetilde D_{0}$ is drawn from the stationary distribution of $D_{t}(\theta)$. Note that using the same sequence of $(\epsilon_{t},u_{t})$'s, the processes
$D_{t}$ and $\widetilde{D}_{t}$ will couple whenever both are equal
to zero, which will happen almost surely, since at every $t$, the
probability of hitting zero is at least $\bar{\Phi}\left(\hat{\theta}/\sigma\right)$,
where $\hat{\theta}=\alpha\bar{\theta}+(1-\alpha)m$ and $\bar{\Phi}(x)$ is the tail cumulative distribution function of the standard Normal distribution. Finally, we have that for all $\theta\in\Theta$
\[
\mathbb{P}\left(\frac{\partial}{\partial d}\phi(\widetilde{D}_{0},u_{1},\epsilon_{1};\theta)=0\right)>0,
\]
This is because the derivative $\frac{\partial}{\partial d}\phi(\widetilde{D}_{0},u_{1},\epsilon_{1};\theta)$ will be zero whenever $\alpha\min\{D_{0}+u_{1},\theta\}+(1-\alpha)m+\epsilon_{1}=0$,
which happens with probability at least $\bar{\Phi}\left(\hat{\theta}/\sigma\right)$.
By Theorem 4.1 of \cite{glasserman1992stationary}, for any $\theta\in\Theta$, $\left(D_{t}(\theta),L_{t}(\theta)\right)$
converges in distribution to $\left(D_{\infty}(\theta),L_{\infty}(\theta)\right)$.
Since $L_{\infty}(\theta)$ is bounded, it follows that $\mathbb{E}\left[L_{t}(\theta)\right]\rightarrow\mathbb{E}[L_{\infty}(\theta)]$ as $t\rightarrow\infty$.

%Finally, it remains to show that $\mathbb{E}[L_{\infty}(\theta)]=\frac{d}{d\theta}\mathbb{E}[D_{\infty}(\theta)]$.
Next, note that $\phi$ is Lipschitz in $a,b,c,\theta$ with Lipschitz
constant $1$. $\mathbb{E}[D_{t}(\theta)]<\infty$ and $\mathbb{E}[L_{t}(\theta)]<\infty$
for any $t$. Moreover, $\bar{L}_{t}(\theta):=\frac{1}{n}\sum_{s=1}^{t}L_{s}(\theta)\leq1$
almost surely, which implies that for $c>1$
\[
\sup_{t}\int_{0}^{\bar{\theta}}|\bar{L}_{t}(\theta)|1\{\bar{L}_{t}(\theta)>c\}d\theta=0.
\]
Then, by Theorem 5.1 of \cite{glasserman1992stationary}, we have 
\[
\mathbb{E}[L_{\infty}(\theta)]=\frac{d}{d\theta}\mathbb{E}[D_{\infty}(\theta)].
\]
\Halmos

% Letting $X_{t}=\inf\{h\leq t:D_{t-h-1}+\eta_{t}>\theta\}$, we observe
% that $1\{D_{t}>0\}\alpha^{1+X_{t}}$ satisfies the same recursion
% (\ref{eq:grad_recursion}):

\subsection{\label{sec:newsvendor-conv-proof}Proof of Theorem \ref{thm:newsvendor-conv}}

We prove Theorem \ref{thm:newsvendor-conv} by verifying that the conditions in Thereom \ref{thm:main2} hold for the augmented Markov chain $Z_t=(D_t, L_t)$.
%We verify the assumptions for the proof of convergence. 
We use the following for the choice of the Lyapunov function and the metric over $z=(d,l)\in\Omega$:
\begin{align*}
V(d,x) & =1+1\{d>0\text{ or }l>0\};\\
d(z,z') & =2 \cdot 1\{z\neq z'\}.
\end{align*}

We first show that $V$ and $V^4$ are both valid Lyapunov functions, and $Z_t$ satisfies Wasserstein contraction with respect to the metric $d(z,z')$, i.e., the total variation distance. In particular, we shall verify Assumptions \ref{ass:ergodicity0} and \ref{ass:bound1} holds for $Z_t$.

\paragraph*{Lyapunov function.}
The key observation is that for any $\theta\in \Theta$, $C=\{d=0,l=0\}$ is a recurrent aperiodic atom of the Markov chain. This is because
it is possible to reach zero demand with probability at least $\bar{\Phi}\left(\hat{\theta}/\sigma\right)$. Then,
%Consider a Lyapunov function of the form $V(d,l)=1+1\{d>0\text{ or }l>0\}$. We have that
\begin{align*}
&\mathbb{E}[V(D_1,L_1]-V(d_{0},l_{0})\\
=&\mathbb{E}[1+1\{D_{1}>0\text{ or } L_{1}>0\}]-(1+1\{d_{0}>0\text{ or }l_{0}>0\})\\
\leq&\mathbb{E}[1+1-1\{D_{1}=0\}]-(1+1\{d_{0}>0\text{ or }l_{0}>0\})\\
\leq&-\bar{\Phi}\left(\hat{\theta}/\sigma\right)+1-1\{d_{0}>0\text{ or }l_{0}>0\})\\
\leq&-\bar{\Phi}\left(\hat{\theta}/\sigma\right)(1+1\{d_{0}>0\text{ or }l_{0}>0\})+1\\
=&-\bar{\Phi}\left(\hat{\theta}/\sigma\right)V(d_{0},l_{0})+1.
\end{align*}
This simple That $V(d,l)$ is a Lyapunov function.

Since $V(d,x)^{4}=1+15\cdot1\{d>0,x<1\}$ this implies that $V(d,x)^{4}$
is also a Lyapunov function. We have thus verified Assumption \ref{ass:bound1}.

%Finally, we have that the gradient is bounded by the Lyapunov function, as it is bounded by a constant.

\paragraph*{Wasserstein ergodicity.}
Let $\nu(\cdot)=\bar{\Phi}\left(\hat{\theta}/\sigma\right)\delta_{(0,0)}(\cdot)$
where $\delta_{(0,0)}$ is the delta measure on $(0,0)$. 
By construction, $\nu(0,0)>0$.
%Since $(0,0)$ is the only point on which $\nu$ has positive measure, for any $z=(d,l)$ there is always a positive probability of transitioning to $(0,0)$:
In addition, we have for any $z\in\Omega$,
\[
\mathbb{P}_{z}(D_{t}=0,L_{t}=0)\geq\bar{\Phi}\left(\hat{\theta}/\sigma\right)
\]
Thus, $Z_t$ satisfies the Doeblin's condition. By Theorem 16.2.4 of \cite{meyn2012markov}, we have
\[
\sup_{z\in\Omega}||\delta_{z}P_{\theta}^{n}-\pi_{\theta}||_{\text{TV}}\leq\left(1-\bar{\Phi}\left(\hat{\theta}/\sigma\right)\right)^{n}.
\]
%which further implies Wasserstein convergence under the metric $d(z,z')= 2 \cdot 1\{z\neq z'\}$.\\

We next Assumption \ref{ass:lip0}, i.e., the Lipschitz continuity properties.

\paragraph*{Lipschitzness of the transition kernel.}
We would like to show that $P_{\theta}$ is Lipschitz in $\theta$
with respect to the total variation distance. To simplify the notation, we write $Z_t=Z_t(\theta)$ and $Z_t^{\prime}=Z_t(\theta^{\prime})$.
Since
\[\begin{split}
D_1&=(\alpha\min\{d_{0}+u_{1},\theta\}+(1-\alpha)m+\epsilon_{1})^{+},\\
D_1'&=(\alpha\min\{d_{0}+u_{1},\theta'\}+(1-\alpha)m+\epsilon_{1})^{+}
\end{split}\]
and
\[\begin{split}
L_1&=1\{D_{1}>0\}\alpha\left(1\{d_{0}+u_{1}>\theta\}+1\{d_{0}+u_{1}\leq\theta\}l_{0}\right),\\
L_1'&=1\{D_{1}'>0\}\alpha\left(1\{d_{0}+u_{1}>\theta\}+1\{d_{0}+u_{1}\leq\theta'\}l_{0}\right).
\end{split}\]
we have
\begin{align*}
 & \sup_{|f|\leq1}|\mathbb{E}\left[f(D_{1},L_{1})-f(D'_{1},L'_{1})\right]|\\
 %=&\sup_{|f|\leq1}|\mathbb{E}\left[f((\alpha\min\{d_{0}+\eta_{1},\theta\}+(1-\alpha)m+\epsilon_{1})^{+}\\
 %&,1\{D_{1}>0\}\alpha\left(1\{d_{0}+\eta_{1}>\theta\}+1\{d_{0}+\eta_{1}\leq\theta\}l_{t}\right)\right]\\
 %& -\mathbb{E}\left[f(D'_{t+1},1\{D'_{t+1}>0\}\alpha\left(1\{D_{t}+\eta_{t+1}>\theta'\}+1\{D_{t}+\eta_{t+1}\leq\theta'\}l_{t}\right)\right]|\\
%(a) =&\sup_{|g|\leq1}|\mathbb{E}g(\min\{d_{0}+\eta_{1},\theta\}+\epsilon_{1},\eta_{1}-\theta)\\
%&-\mathbb{E}g(\min\{d_{t}+\eta_{t+1},\theta'\}+\epsilon_{t+1},\eta_{t+1}-\theta')|\\
%=&\sup_{|g|\leq1}|\mathbb{E}g((\theta+\min\{d_{t}+\eta_{t+1}-\theta,0\}+\epsilon_{t+1},\eta_{t+1}-\theta) \\
%& - \mathbb{E}g(\theta'+\min\{d_{t}+\eta_{t+1}-\theta',0\}+\epsilon_{t+1},\eta_{t+1}-\theta')|\\
(a)  =&\sup_{|h|\leq1}|\mathbb{E}h(\epsilon_{1}+\alpha\theta,u_{1}-\theta)-\mathbb{E}h(\epsilon_{1}+\alpha\theta',u_{1}-\theta')|\\
=&d_{\text{TV}}\left((\epsilon_{1}+\alpha\theta,u_{1}-\theta),(\epsilon_{1}+\alpha \theta',u_{1}-\theta')\right)\\
(b) \leq &d_{\text{TV}}\left(\epsilon_{1}+\alpha\theta,\epsilon_{1}+\alpha\theta'\right)+d_{\text{TV}}\left(u_{1}-\theta,\eta_{1}-\theta'\right)\\
(c) \leq&\frac{2|\theta-\theta'|}{\sigma}.
\end{align*}
For (a), we use the fact that any bounded function $f$ of the 
update can be rewritten as a bounded function $h$ of $\epsilon_{1}+\alpha\theta$ and $\eta_{1}-\theta$. 
%$g$ over $\min\{d_{t}+\eta_{t+1},\theta\}+\epsilon_{t+1}$
%and $\eta_{t+1}-\theta$. In (b), we can further rewrite this as a bounded function $\eta_{t+1}-\theta$ and $\epsilon_{t+1}+\theta$.
For (b), we use the independence of $u_{1}$ and $\epsilon_{1}$ to bound the total variation of the pair $(\epsilon_{1}+\theta,u_{1}-\theta)$
by total variation of the marginals. The inequality (c) follows
from a standard upper bound on the total variation between two normal distributions. The above bound implies
\[
\sup_{z\in\Omega}||\delta_{z}P_{\theta}-\delta_{z}P_{\theta'}||_{\text{TV}}\leq\frac{2}{\sigma}|\theta-\theta'|.
\]

\paragraph*{Lipschitzness of the gradient.}
We can show that $\nabla\ell(\theta)$ is Lipschitz in $\theta$ as a consequence of Theorem 3.1 of \cite{rudolf2018perturbation}, which shows that Lipschitzness of the transition kernel, which we have established above, implies Lipschitzness of the stationary distribution under
the total variation distance:
\begin{align*}
\|\mu_{\theta}-\mu_{\theta'}\|_{\text{TV}} & \leq\frac{\alpha/\sigma}{\bar{\Phi}\left(\hat{\theta}/\sigma\right)^{2}}\|\theta-\theta'\|.
\end{align*}
\[
g(\theta,Z_{t})=\left(b1\{D_{t}>\theta\}-h1\{D_{t}<\theta\}\right)\left(1+L_{t}\right)
\]
Since $\nabla\ell(\theta)=\E_{\mu_{\theta}}[\left(b1\{D_{t}>\theta\}-h1\{D_{t}<\theta\}\right)\left(1+L_{t}\right)]$, 
%involves the expectation of a bounded function of $D_{t}$ and $L_{t}$, this guarantees that 
$\nabla\ell(\theta)$ is Lipschitz. 
%with constant $C$. 

In addition, we note that since $|g(\theta,z)|\leq2\max\{h,b\}$,
\[
\|g(\theta,z)-g(\theta,z')\|\leq4\max\{h,b\}1\{z\neq z'\}.
\]
% we can observe that the
% gradient estimator $g$ is bounded by $2\max\{h,b\}$ which implies
% that it is Lipschitz under the trivial metric
% \[
% |g(\theta,z)|\leq2\max\{h,b\}\implies||g(\theta,z)-g(\theta,z')||\leq4\max\{h,b\}1\{z\neq z'\}
% \]

Lastly, we verify Assumption \ref{ass:bound0}
\paragraph*{Bounded functions and gradient}
For the objective function, we have 
\[\ell(\theta)\leq\bar{\theta}+m+5\sigma.\]
For the gradients, we have 
%are bounded almost surely by a constant for all $\theta$: 
\[\nabla\ell(\theta)\leq2\max\{h,b\}\]
and 
\[|g(\theta,z)|\leq2\max\{h,b\}\]
for any $\theta\in\Theta$.
\Halmos

%\section{Proofs for the results in Section~\ref{sec:queue}}

\section{Proof of Theorem \ref{thm:gg1-convex}}\label{sec:proof-theorem-gg1}

We prove Theorem \ref{thm:gg1-convex} by verifying that the conditions in Theorem \ref{thm:main2} hold for the augmented Markov chain $Z_t=(W_t, X_t)$.

We first verify Assumptions \ref{ass:ergodicity0} and \ref{ass:bound1}, which requires establishing the Lyapunov drift conditions and Wasserstein contraction of $P_{\theta}$ for $\theta\in\Theta$. 
Let
\[V(w,x)=e^{\alpha_{1}w+\alpha_{2}x}\]
for $0<\alpha_2<\alpha_{1}<\min\{\alpha^{*}/\underline{\mu},\alpha^{*}/\lambda(\underline{p})\}$,
where $\alpha^{*}, \alpha_1,\alpha_2$ are defined in Assumption \ref{assu:dist-assumption}.

\begin{lemma}[Lyapunov drift condition] \label{lm:lyapunov}
There exists $\rho\in(0,1)$ such
that 
\[
\sup_{\theta\in \Theta}P_{\theta}V(w_{0},x_{0})\leq\rho V(w_{0},x_{0})+1
\]
and
\[
\sup_{\theta\in \Theta}P_{\theta}V(w_{0},x_{0})^4\leq\rho V(w_{0},x_{0})^4+1.
\]
\end{lemma}

\proof{Proof.}
Let $W_{0}$ and $X_{0}$ be the current waiting time and server's busy time respectively, and $Y=S/\mu-T/\lambda(p)$ be the increment brought by the next customer. %To show the ergodicity of the Markov chain $Z_{t}$, 
% We consider the Lyapunov function 
% \[V(w,x)=e^{\alpha_{1}w+\alpha_{2}x}\]
% for $\alpha_{1}<\min\{\alpha^{*}/\underline{\mu},\alpha^{*}/\lambda(\underline{p})\}$
% where $\alpha^{*}$ is defined in Assumption \ref{assu:dist-assumption}.
\begin{align*}
\mathbb{E}[e^{\alpha_{1}W_{1}+\alpha_{2}X_{1}}] & =\mathbb{E}\left[e^{\alpha_{1}(W_{0}+Y)^{+}+\alpha_{2}\left(X_{0}+\frac{T}{\lambda(p)}\right)1\{(W_{0}+Y)^{+}>0\}}\right]\\
 & \leq\mathbb{E}\left[e^{\left(\alpha_{1}(W_{0}+Y)+\alpha_{2}(X_{0}+\frac{T}{\lambda(p)})\right)1\{(W+Y)^{+}>0\}}\right]+\mathbb{E}[1\{(W+Y)^{+}=0\}]\\
 & \leq\mathbb{E}\left[e^{\alpha_{1}(W_{0}+Y)+\alpha_{2}(X_{0}+\frac{T}{\lambda(p)})}\right]+1\\
&=e^{\alpha_{1}W_{0}+\alpha_{2}X_{0}}\mathbb{E}\left[e^{\alpha_{1}\frac{S}{\mu}+(\alpha_{2}-\alpha_{1})\frac{T}{\lambda(p)}}\right]+1.
\end{align*}
By Assumption \ref{assu:dist-assumption}, there exists $0<\alpha_{2}<\alpha_{1}<\alpha^{*}$
to be small enough so that 
\[\mathbb{E}\left[e^{4\alpha_{1}\frac{S}{\underline{\mu}}}\right]\mathbb{E}\left[e^{-4(\alpha_{1}-\alpha_{2})\frac{T}{\lambda(\underline{p})}}\right]<1,
\]
which is possible because $\lambda(\underline{p})<\underline\mu$. By the convexity of $h(x)=x^4$ and Jensen's inequality, we have:
\[
\mathbb{E}\left[e^{\alpha_{1}\frac{S}{\underline{\mu}}}\right]^{4}\mathbb{E}\left[e^{-(\alpha_{1}-\alpha_{2})\frac{T}{\lambda(\underline{p})}}\right]^{4}
\leq \mathbb{E}\left[e^{4\alpha_{1}\frac{T}{\underline{\mu}}}\right]\mathbb{E}\left[e^{-4(\alpha_{1}-\alpha_{2})\frac{S}{\lambda(\underline{p})}}\right] <1
\]
Let $\rho=\mathbb{E}[e^{\alpha_{1}\frac{S}{\underline{\mu}}}]\mathbb{E}[e^{-(\alpha_{1}-\alpha_{2})\frac{T}{\lambda(\underline{p})}}]$.
This condition implies that
%\begin{equation}
\[P_{\theta}V(w_{0},x_{0})\leq\rho V(w_{0},x_{0})+1.\]
%\label{eq:basic_drift_condition}
%\end{equation}
%which implies that for $\theta\in\Theta$, the Markov chain is uniformly ergodic.

Similarly (by repeating the same arguments), we can show that
\[P_{\theta}V(w_{0},x_{0})^4\leq\rho V(w_{0},x_{0})^4+1.\]
\Halmos
\endproof

The above Lyapunov drift condition (Lemma \ref{lm:lyapunov}) implies for $\beta=\frac{1}{2}(1-\rho)$ and a set 
\[
C=\{z\in\mathbb{R}^{2}:V(z)\leq1/\beta\},
\]
we have, by Lemma 15.2.8 of \citet{meyn2012markov},
\begin{equation}
P_{\theta}V(z)-V(z)\leq-\beta V(z)+\mathbf{1}_{C}(z).\label{eq:drift_condition}
\end{equation}

Consider the metric
\begin{equation}
d_{V}(z,z'):=(V(z)+V(z'))1\{z\neq z'\}\label{eq:v_norm}
\end{equation}

% that a uniform Wasserstein coupling
% constant. For any function $V\geq1$, consider the metric $d_{V}(z,z')$
% for $z=(w,x)\in\mathbb{R}^{2}$ given by:

% \begin{equation}
% d_{V}(z,z'):=(V(z)+V(z'))1\{z\neq z'\}\label{eq:v_norm}
% \end{equation}

% By Lemma 15.2.8 in \citet{meyn2012markov} above condition implies
% that there exists $\lambda\in(0,1)$ and a petite set $C\subseteq\mathbb{R}^{2}$
% such that

% \begin{equation}
% P_{\theta}V(z)-V(z)\leq-\beta V(z)+\mathbf{1}_{C}(z)\label{eq:drift_condition}
% \end{equation}
% where $\beta=\frac{1}{2}(1-\gamma)$ and 
% \[
% C=\{z\in\mathbb{R}^{2}:V(z)\leq1/\beta\}
% \]
% where $C$ will be identical across parameters $\theta\in\Theta$. 

% This in turn will imply that $(W_{n},X_{n})$ is $V$-uniform ergodic.
% We use standard techniques to show that the uniform drift condition
% for the $G/G/1$ queue implies a uniform rate of convergence to stationarity.
\begin{lemma}[Wasserstein contraction] \label{lm:wasserstein}
There exists $K\in(0,\infty)$ and $\rho\in(0,1)$
such that for all $t\in\mathbb{N}$,
\end{lemma}

\[
\sup_{\theta\in \Theta}\sup_{z,z'\in \Omega}\frac{||P_{\theta}^{t}(z,\cdot)-P_{\theta}^{t}(z',\cdot)||_{V}}{d_{V}(z,z')}\leq K\rho^{t}
\]

\proof{Proof.}
Let ${\bf 0}:=(0,0)$, which is a recurrent atom for the Markov
chain $Z_{t}=(W_{t},X_{t})$. Let $\tau_{0}$ denote the first return time
to ${\bf 0}$, i.e.,
\[
\tau_0=\inf\{t\geq 1: Z_t={\bf 0}\},
\]
and $\pi_{\theta,0}$ denote the probability of
${\bf 0}$ under the stationary measure of $Z_{t}$.
We first note the following bound 
on the generating function of the distance to the stationary
distribution, which was developed in Proposition 4.2 of \citep{baxendale2005renewal}, for any $r>1$,
\begin{align*}
\sum_{t=1}^{\infty}r^{t}||P_{\theta}^{t}(z,\cdot)-\mu_{\theta}||_{V}  \leq& \mathbb{E}_{z}\left[\sum_{t=0}^{\tau_{0}}V(Z_{t})r^{t}\right]+\mathbb{E}_{\bf 0}\left[\sum_{t=0}^{\tau_{0}}V(Z_{t})r^{t}\right]\frac{\mathbb{E}_{z}[r^{\tau_{0}}]-1}{r-1}\\
 & +\mathbb{E}_{z}[r^{\tau_{0}}]\mathbb{E}_{0}\left[\sum_{t=0}^{\tau_{0}}V(Z_{t})r^{t}\right]\left|\sum_{t=1}^{\infty}(P_{\theta}^{t}({\bf 0}, {\bf 0})-\pi_{\theta,0})r^{t}\right|\\
 & +\frac{\mathbb{E}_{\bf 0}\left[\sum_{t=0}^{\tau_{0}}V(Z_{t})r^{t}\right]+r\mathbb{E}_{\bf 0}\left[\sum_{t=0}^{\tau_{0}}V(Z_{t})\right]}{r-1}
\end{align*}

Since $V(z)\geq1$, we have $\mathbb{E}_{z}[r^{\tau_{0}}]\leq\mathbb{E}_{z}\left[\sum_{t=0}^{\tau_{0}}V(Z_{t})r^{t}\right]$.
Theorem 15.2.5 of \citet{meyn2012markov} shows that the geometric
drift condition (\ref{eq:drift_condition}) implies a bound on the generating function of $V(Z_{t})$. In particular, for any $r\in(1,(1-\beta)^{-1})$,
let $\varepsilon=r^{-1}-(1-\beta)>0$. Then,
\begin{align*}
\mathbb{E}_{z}\left[\sum_{t=0}^{\tau_{0}}V(Z_{t})r^{t}\right] & \leq(r\varepsilon)^{-1}V(z)+\varepsilon^{-1}\mathbb{E}_{z}\left[\sum_{t=0}^{\tau_{0}}{\bf 1}_{C}(Z_{t})r^{t}\right]\\
 & \leq(r\varepsilon)^{-1}V(z)+\frac{r\varepsilon^{-1}}{r-1}\sup_{z\in C}\mathbb{E}_{z}\left[r^{\tau_{0}}\right].
\end{align*}

In what follows, we first develop a bound for $\sup_{\theta\in\Theta}\sup_{z\in C}\mathbb{E}_{z}\left[r^{\tau_{0}}\right]$. We write $\tau_{0}(\theta)$ and $Y(\theta)=S/\mu-T/\lambda(p)$ to mark dependence of the distribution of $\tau_{0}$ and $Y(\theta)$ on $\theta$ explicitly.  
Conditional on that $Z_0\in C$, the maximum workload is $W_{0}=\frac{1}{\alpha_{1}}\log\frac{1}{\beta}$.
%Thus, %$\sup_{z\in C}\mathbb{E}_{z}\left[r^{\tau_{0}}\right]$
%$\tau_0$ is bounded by the first hitting time of zero for a random walk starting from $\frac{1}{\alpha_{1}}\log\frac{1}{\beta}$ and with increments distributed as $Y$. 
Meanwhile, by coupling, for any fixed starting workload
$w_{0}$ and $(\mu,p)$, $\tau_{0}(\mu,p)\leq_{st}\tau_{0}(\underline{\mu},\underline{p})$.
Above all, we can bound $\tau_{0}(\theta)$ by the first return time to zero of a random walk starting from $\frac{1}{\alpha_{1}}\log\frac{1}{\beta}$ and with increments distributed as $Y(\underline{\mu},\underline{p})$.
%we can bound $\sup_{\theta\in\Theta}\sup_{z\in C}\mathbb{E}_{z}\left[r^{\tau_{0}}\right]$
%by considering the random walk with the worst-case negative drift
%$\bar{\theta}=(\underline{\mu},\underline{p})$. For the random walk with $\bar{\theta}$, 
%Note that for the random walk with increments distributed as $Y(\underline{\mu},\underline{p})$, 
For this random walk, let $\tilde \tau_0$ denote its first return time to $0$. Then, there exists $\kappa>1$ such that $\mathbb{E}_{z}\left[\kappa^{\tilde \tau_{0}}\right]<\infty$
for all $z\in C$. Let $\bar{r}=\min\{\kappa,(1-\beta)^{-1}-\varepsilon\}$
for any fixed $\varepsilon>0$. Let 
$\bar K=\mathbb{E}_{\frac{1}{\alpha_{1}}\log\frac{1}{\beta}}\left[
\bar r^{\tau_{0}(\underline{\mu},\underline{p})}\right]$.
%and let $\bar{K}$ denote
%the constant that simultaneously upper bounds $\sup_{\theta\in\Theta}\sup_{z\in C}\mathbb{E}_{z}\left[\bar{r}^{\tau_{0}}\right]$
Then, 
\begin{equation}\label{eq:bound1}
\sup_{\theta\in\Theta}\sup_{z\in C}\mathbb{E}_{z}\left[r^{\tau_{0}}\right]\leq\bar K 
\mbox{ and } \sup_{\theta\in\Theta}\sup_{z\in C}\mathbb{E}_{z}\left[\sum_{t=0}^{\tau_{0}}V(Z_{t})\bar{r}^{t}\right]\leq\bar{K}V(z).
\end{equation}

We next bound $\left|\sum_{n=1}^{\infty}(P_{\theta}^{n}({\bf 0},{\bf 0})-\pi_{\theta,0})r^{n}\right|$ following the construction in Theorem 3.2 of \cite{baxendale2005renewal}.
% we can use a quantitative version of Kendall's renewal theorem in
% \citet[Theorem 3.2]{baxendale2005renewal}. 
The sequence $P_{\theta}^{n}({\bf 0}, {\bf 0})$
can be seen as a renewal process with increment distribution $b_{n}:=\mathbb{P}_{\bf 0}(\tau_{0}=n)$.
The key conditions required for bounding $\sum_{n=1}^{\infty}|P_{\theta}^{n}({\bf 0}, {\bf 0})-\pi_{\theta,0}|r^{n}$
uniformly across $\theta \in \Theta$ are some bounds for $b_n$.
%is a geometric moment condition for $b_{n}$ and the existence of $\beta>0$ such that $b_{1}\geq\beta$. 
First, note that 
\[\sum_{n=0}^{\infty}b_{n}\bar{r}^{n}=\mathbb{E}_{0}[\bar{r}^{\tau_{0}}]\leq \bar K.\]
In addition, 
%we can obtain a lower bound of $b_{1}$ as the probability that the workload remains zero when starting from zero under the worst-case parameters $\bar{\theta}$, i.e.,
\[
b_{1}=\mathbb{P}_{0}(\tau_{0}=1)=\mathbb{P}\left(\frac{S}{\mu}-\frac{T}{\lambda(p)}\leq0\right)\geq\mathbb{P}\left(\frac{S}{\underline{\mu}}-\frac{T}{\lambda(\underline{p})}\leq0\right)=:\beta
\]
By Theorem 3.2 of \cite{baxendale2005renewal}, the radius of convergence
of $\sum_{t=1}^{\infty}|P_{\theta}^{t}({\bf 0})-\pi_{\theta,0}|r^{t}$
is at least $R_{1}>1$, where 
\[
R_{1}:=\left\{ r\in(1,\bar{r}):\frac{r-1}{r(\log\bar{r}/r)^{2}}=\frac{e^{2}\beta}{8\frac{\bar{K}-1}{\bar{r}-1}}\right\}.
\]
Then, for any $r\leq R_{1}$, there exists $K_{1}$ that only depends on
$r,\beta,\bar{r},\bar{K}$ such that 
\begin{equation}\label{eq:bound2}
\sup_{\theta\in\Theta} \left|\sum_{t=1}^{\infty}|P_{\theta}^{t}({\bf 0},{\bf 0})-\pi_{\theta,0}|r^{t}\right|\leq K_{1}
\end{equation}

Putting the two bounds \eqref{eq:bound1} and \eqref{eq:bound2} together, 
taking any $\hat{r}<R_{1}$, there exists $K_2$ depending
only on $\kappa,\beta,\bar{K},K_{1},\bar{r},\epsilon$ such that 
\[
\sup_{\theta\in\Theta} \sum_{n}\hat{r}^{t}||P_{\theta}^{t}(z,\cdot)-\mu_{\theta}||_{V}\leq K_2V(z)
\]
 This implies that there exists $\rho<\frac{1}{\hat{r}}<1$ and a
constant $K$ such that

\[
\sup_{\theta\in\Theta}\|P_{\theta}^{t}(z,\cdot)-\mu_{\theta}\|_{V}\leq K\rho^{t}V(z),
\]
%Since these bounds hold uniformly across $\theta$, this implies that
which implies
\[
\sup_{\theta\in \Theta}\sup_{z\in \Omega}\frac{||P_{\theta}^{t}(z,\cdot)-\mu_{\theta}||_{V}}{V(z)}\leq K\rho^{t}.
\]
By Lemma 3.2 of \citet{rudolf2018perturbation}, this further implies 
%that ergodicity coefficient is uniformly bounded by $K\rho^{t}$:
\[
\sup_{\theta\in \Theta}\sup_{z,z'\in \Omega}\frac{||P_{\theta}^{n}(z,\cdot)-P_{\theta}^{n}(z',\cdot)||_{V}}{d_{V}(z,z')}\leq K\rho^{t}.
\]
\Halmos
\endproof

We next verify Assumption \ref{ass:lip0}, i.e., the smoothness of the one-step transition kernel and the gradients.

\begin{lemma}[Lipschitz continuity of the one-step transition kernel] \label{lm:smooth_P}
There exists a constant $\Gamma\in(0,\infty)$ such that for all $z\in\mathbb{R}_{+}^{2}$
\[
d(\delta_{z}P_{\theta}^{n},\delta_{z}P_{\theta'}^{n})\leq\Gamma||\theta-\theta'||V(z)
\]
\end{lemma}

\proof{Proof.}
Let $C$ be a generic constant whose value may change from line
to line. By Assumption \ref{assu:dist-assumption}, $S$ and $T$ have $\mathcal{C}^{1}$ densities.
The density of $S/\mu$ is $\mu f_{S}(\mu x)$ and the density of
$T/\lambda(p)$ is $\lambda(p)f_{T}(\lambda(p)x)$. A key property
we require is that these densities are sufficiently smooth in $\mu$
and $\lambda(p)$ respectively, i.e., Assumption
\ref{assu:dist-assumption} (iii).

Consider a fixed starting state $z_{0}=(w_{0},x_{0})$. Let $\phi_{\theta}$
be the joint probability density function of $(w_{0}+\frac{S}{\mu}-\frac{T}{\lambda(p)},x_{0}+\frac{T}{\lambda(p)})$
for $\theta=(\mu,p)$. By the law of the unconscious statistician, we have
\[
\mathbb{E}_{\delta_{z_{0}}P_{\theta}}[f(W_{1},X_{1})]=\int_{\mathbb{R}}\int_{\mathbb{R}}\left[f(w,x)1\{w>0\}+f(0,0)1\{w\leq0\}\right]\phi_{\theta}(w,x)dwdx
\]
Then, for any measurable $f$ with $|f|\leq V$, we have
\begin{align}
 & |\mathbb{E}_{\delta_{z_{0}}P_{\theta}}[f(W_{1},X_{1})]-\mathbb{E}_{\delta_{z_{0}}P_{\theta'}}[f(W_{1},X_{1})]|\nonumber \\
 =&\left|\int_{\mathbb{R}}\int_{\mathbb{R}}f(w,x)1\{w>0\}\phi_{\theta}(w,x)dwdx-\int_{\mathbb{R}}\int_{\mathbb{R}}f(w,x)1\{w>0\}\phi_{\theta'}(w,x)dwdx\right|\nonumber \\
 & +\left|\int_{\mathbb{R}}\int_{\mathbb{R}}f(0,0)1\{w\leq0\}\phi_{\theta}(w,x)dwdx-\int_{\mathbb{R}}\int_{\mathbb{R}}f(0,0)1\{w\leq0\}\phi_{\theta'}(w,x)dwdx\right|\nonumber \\
 \leq&\int_{\mathbb{R}}\int_{\mathbb{R}}V(w,x)|\phi_{\theta}(w,x)-\phi_{\theta'}(w,x)|dwdx\label{eq:v_term}\\
 & +\left|\mathbb{P}\left(w_{0}+\frac{S}{\mu}-\frac{T}{\lambda(p)}\leq0\right)-\mathbb{P}\left(w_{0}+\frac{S}{\mu'}-\frac{T}{\lambda(p')}\leq0\right)\right|\label{eq:prob_term}
\end{align}

We first bound (\ref{eq:v_term}). Note that
\begin{align*}
 & \int_{\mathbb{R}}\int_{\mathbb{R}}V(w,x)|\phi_{\theta}(w,x)-\phi_{\theta'}(w,x)|dwdx\\
=&\int_{\mathbb{R}}\int_{\mathbb{R}}e^{\alpha_{1}w+\alpha_{2}x}|\phi_{\theta}(w,x)-\phi_{\theta'}(w,x)|dwdx\\ =&e^{\alpha_{1}w_{0}+\alpha_{2}x_{0}}\int_{\mathbb{R}}\int_{\mathbb{R}}e^{\alpha_{1}w+\alpha_{2}x}|\psi_{\theta}(w,x)-\psi_{\theta'}(w,x)|dwdx,
\end{align*}
where for the last equation, we take a change of variables and $\psi_{\theta}$ is the joint density
of $(\frac{S}{\mu}-\frac{T}{\lambda(p)},\frac{T}{\lambda(p)})$, i.e., 
\[
\psi_{\theta}(w,x)=\mu f_{S}(\mu(w+x))\cdot\lambda(p)f_{T}(\lambda(p)x)
\]
by independence of $f_{S}$ and $f_{T}$. 
To be more concise, in what follows we denote $\lambda=\lambda(p)$ and $\lambda'=\lambda(p')$. 
% We also use the shorthand $I(w,x)$ for the integrand:
% \begin{align}
% I(w,x) &:=e^{\alpha_{1}w+\alpha_{2}x}|\psi_{\theta}(w,x)-\psi_{\theta'}(w,x)|dwdx.
% \end{align}
% Consider the constant $c$ in Assumption \ref{assu:dist-assumption}. We can split the integral as follows
% \begin{align}
% &\int_{0}^{\infty}\int_{\mathbb{R}}I(w,x)dwdx = \int_{0}^{\infty}\int_{-x}^{\infty}I(w,x)dwdx\\
% &=\int_{c}^{\infty}\int_{c-x}^{\infty}I(w,x)dwdx+\int_{0}^{c}\int_{c-x}^{\infty}I(w,x)dwdx+\int_{c}^{\infty}\int_{-x}^{c-x}I(w,x)dwdx+\int_{0}^{c}\int_{-x}^{c-x}I(w,x)dwdx
% \label{eq:tv-integral}
% \end{align}
% since for any $\theta$, $\psi_{\theta}(w,x)$ is non-zero only when $x\geq 0$ and $w\geq -x$. 
Next,
\begin{align}
 & \int_{0}^{\infty}\int_{\mathbb{R}}e^{\alpha_{1}w+\alpha_{2}x}|\psi_{\theta}(w,x)-\psi_{\theta'}(w,x)|dwdx \nonumber\\
 %\label{eq:lipschitz_bound}\\
=&\int_{0}^{\infty}\int_{-x}^{\infty}e^{\alpha_{1}w+\alpha_{2}x}|\mu f_{S}(\mu(w+x))\lambda f_{T}(\lambda x)-\mu'f_{S}(\mu'(w+x))\lambda'f_{T}(\lambda'x)|dwdx\nonumber \\ 
\leq&\int_{0}^{\infty}\int_{-x}^{\infty}e^{\alpha_{1}w+\alpha_{2}x}|\mu f_{S}(\mu(w+x))-\mu'f_{S}(\mu'(w+x))|\cdot\lambda f_{T}(x)dwdx\label{eq:lip_first_term}\\
 & +\int_{0}^{\infty}\int_{-x}^{\infty}e^{\alpha_{1}w+\alpha_{2}x}|\lambda f_{T}(\lambda x)-\lambda'f_{T}(\lambda'x)|\cdot\mu'f_{S}(\mu'(w+x))dwdx\label{eq:lip_second_term}
\end{align}

%We wish to show that $\mu f_{S}(\mu(w+x))$ is Lipschitz in $\mu$ and $\lambda f_{T}(\lambda x)$ is Lipschitz in $\lambda$. 
Since $f_{S}$ and $f_{T}$ are $\mathcal{C}^{1}$, by the mean-value theorem applied pointwise, we have
\begin{align*}
|\mu f_{S}(\mu x)-\mu'f_{S}(\mu'x)| & \leq L_{S}(x)|\mu-\mu'|\\
|\lambda f_{T}(\lambda x)-\lambda'f_{T}(\lambda'x)| & \leq L_{T}(x)|\lambda-\lambda'|,
\end{align*}
where 
\begin{align*}
L_{S}(x) & =\sup_{\mu\in[\underline{\mu},\bar{\mu}]}\left|\frac{d}{d\mu}\mu f_{S}(\mu x)\right|\\
L_{T}(x) & =\sup_{\lambda\in[\lambda(\overline{p}),\lambda(\underline{p})]}\left|\frac{d}{d\lambda}\lambda f_{T}(\lambda x)\right|.
\end{align*}

% and 
% \[
% \frac{d}{d\mu}\left[\mu f_{S}(\mu x)\right]=f_{S}(\mu x)+\mu xf'_{S}(\mu x).
% \]
%By Assumption \ref{assu:dist-assumption}, we can obtain an upper bound for these quantities. 
Note that
\begin{align*}
L_{S}(x) 
&= \sup_{\mu\in[\underline{\mu},\bar{\mu}]} 
\left| f_{S}(\mu x)+\mu x f'_{S}(\mu x) \right|\\
&\leq K1\{0\leq x \leq c\} + 
\sup_{\mu\in[\underline{\mu},\bar{\mu}]} \left[
\left(1+ x \left| \frac{d}{dx}\log f_{S}(\mu x) \right|\right) f_{S}(\mu x)1\{x \geq c\} \right] \\
&\leq K1\{0\leq x \leq c\} + \sup_{\mu\in[\underline{\mu},\bar{\mu}]} \left[(1 +  D_{1}(\mu x) + D_{2} (\mu x)^{k+1}) f_{S}(\mu x) \right] \\
& \leq K1\{0\leq x \leq c\} +  (1 +  D_{1}(\bar{\mu} x) + D_{2} (\bar{\mu} x)^{k+1}) \sup_{\mu\in[\underline{\mu},\bar{\mu}]} f_{S}(\mu x)
\end{align*}
In the first inequality, we use the Weierstrass theorem to obtain an upper bound for $f_{S}$ and $f'_{S}$ over $[0, \bar{\mu}c]$ with the assumption that $f_{S}\in \mathcal{C}^{1}$. For the second inequality, we use Assumption \ref{assu:dist-assumption} (iii) to bound the log-derivative. 
% Similarly, we can develop an upper bound for $L_T(x)$:
% \[
% L_T(x)\leq .
% \]
Then, for \eqref{eq:lip_first_term}, we have
\begin{align*}
 & \int_{0}^{\infty}\int_{-x}^{\infty}e^{\alpha_{1}w+\alpha_{2}x}|\mu f_{S}(\mu(w+x))-\mu'f_{S}(\mu'(w+x))|\cdot\lambda f_{T}(\lambda x)dwdx\\
 =& \int_{0}^{\infty}\int_{0}^{\infty}e^{\alpha_{1}(w-x)+\alpha_{2}x}|\mu f_{S}(\mu w)-\mu'f_{S}(\mu'w))|\cdot\lambda f_{T}(\lambda x)dwdx\\
 \leq& C\lambda|\mu-\mu'|\cdot\int_{0}^{\infty}\int_{0}^{\infty}e^{\alpha_{1}w+(\alpha_{2}-\alpha_{1})x}L_{S}(w)\lambda f_{T}(\lambda x)dwdx\\
 \leq& C\lambda|\mu-\mu'|\cdot\int_{0}^{\infty}\int_{0}^{\infty}e^{\alpha_{1}w}L_{S}(w)\lambda f_{T}(\lambda x)dwdx.
\end{align*}
since $\alpha_{2} < \alpha_{1}$.
For the above integral to be finite, it is sufficient to show that $L(w)$ is exponentially integrable. 
% By Fubini's theorem (for positive integrands),
% \begin{align*}
% &\int_{0}^{\infty}\int_{\mathbb{R}}e^{\alpha_{1}(w+x)}L_{S}(w+x)\lambda f_{T}(\lambda x)dwdx\\
% =&\int_{0}^{\infty}e^{\alpha_{1}y}L_{S}(y)\left[\int_{0}^{\infty}\lambda f_{T}(\lambda x)dx\right]dy\\
%  =&\int_{0}^{\infty}e^{\alpha_{1}y}L_{S}(y)dy\\ 
%\end{align*}
% \begin{align*}
%  & \int_{0}^{\infty}\int_{\mathbb{R}}\int_{\mathbb{R}}e^{\alpha_{1}y}L_{S}(y)\lambda f_{T}(\lambda x)1\{y=w+x\}dydwdx\\
%  =&\int_{0}^{\infty}e^{\alpha_{1}y}L_{S}(y)\left[\int_{0}^{\infty}\int_{\mathbb{R}}\lambda f_{T}(\lambda x)\int_{\mathbb{R}}1\{w=y-x\}dwdx\right]dy\\
%  =&\int_{0}^{\infty}e^{\alpha_{1}y}L_{S}(y)\left[\int_{0}^{\infty}\lambda f_{T}(\lambda x)dx\right]dy\\
%  =&\int_{0}^{\infty}e^{\alpha_{1}y}L_{S}(y)dy<\infty 
% \end{align*}
% by Assumption \ref{assu:dist-assumption} (iii).
%2 (iii), this integral is finite
%\begin{align*}
\begin{align*}
&\int_{0}^{\infty}e^{\alpha_{1}w}L_{S}(w)dw \\
\leq & \int_{0}^{\infty} e^{\alpha_{1}w}K1\{0\leq w \leq c\} +  e^{\alpha_{1}w}(1 +  D_{1}(\bar{\mu} w) + D_{2} (\bar{\mu} w)^{k+1}) \sup_{\mu\in[\underline{\mu},\bar{\mu}]} f_{S}(\mu w) dw \\
\leq & Kce^{\alpha_{1}c} 
+ \int_{0}^{\infty} e^{\alpha_{1}w} (1 +  D_{1}(\bar{\mu} w) + D_{2} (\bar{\mu} w)^{k+1}) \sup_{\mu\in[\underline{\mu},\bar{\mu}]} f_{S}(\mu w) dw < \infty
\end{align*}
%given that there exists $\alpha>\alpha_{1}$ such that $\mathbb{E}[e^{\alpha S/\underline{\mu}}]<\infty$.
Thus, there exists $C'\in(0,\infty)$, such that 
\[
\int_{0}^{\infty}\int_{-x}^{\infty}e^{\alpha_{1}w+\alpha_{2}x}|\mu f_{S}(\mu(w+x))-\mu'f_{S}(\mu'(w+x))|\cdot\lambda f_{T}(\lambda x)dwdx
\leq C'|\mu-\mu'|.
\]

Following similar arguments, we can bound \eqref{eq:lip_second_term} as
\begin{align*}
 & \int_{0}^{\infty}\int_{-x}^{\infty} e^{\alpha_{1}w+\alpha_{2}x}|\lambda f_{T}(\lambda x)-\lambda'f_{T}(\lambda'x)|\cdot\mu'f_{S}(\mu'(w+x))dwdx\\
 \leq& C|\lambda-\lambda'| \int_{0}^{\infty}\int_{0}^{\infty} e^{\alpha_{1}w+(\alpha_{2}-\alpha_{1})x}L_{T}(x)\cdot\mu'f_{S}(\mu'w)dwdx \\
 \leq&C|\lambda-\lambda'| \int_{0}^{\infty} e^{\alpha_{1}w}\cdot\mu'f_{S}(\mu'w) \int_{0}^{\infty}L_{T}(x)dx dw \\
 \leq & C|\lambda-\lambda'| \left(\int_{0}^{\infty}e^{\alpha_{1}w}\mu'f_{S}(\mu'w)dw \right)\\
& \cdot \left[ Kc +  \int_{0}^{\infty}(1 + D_{1}(\bar{\lambda} x) + D_{2} (\bar{\lambda} x)^{k+1}) \sup_{\lambda\in[\underline{\lambda},\bar{\lambda}]} f_{T}(\lambda x) dx\right] \\
 \leq&  C'|\lambda-\lambda'| 
\end{align*}
where $C'\in (0,\infty)$ since $\lambda f_{ T}(\lambda x)$ and $\mu' f_{S}(\mu' w)$ have finite moment-generating functions. 
%for all $\lambda$ and $\mu'$.

% In order to bound the other terms in \eqref{eq:tv-integral}, for the terms with the domain of integration restricted to $x\in[0,c]$  or $w\in [-x,c-x]$, by Weierstrass theorem the terms $L_{T}(x)$ and $L_{S}(w)$  are uniformly bounded, as the densities are assumed to be $C^{1}$.
% \leq& C|\lambda-\lambda'|\int_{0}^{\infty}e^{\alpha_{1}y}\mu'f_{S}(\mu'y)\left[\int_{0}^{y}L_{T}(x)dx\right]dy\\
% Next,
% \begin{align*}
% &\int_{c}^{\infty}\int_{\mathbb{R}} e^{\alpha_{1}w+\alpha_{2}x}L_{T}(x)\cdot\mu'f_{S}(\mu'(w+x))dwdx\\
% \leq& \int_{0}^{\infty}e^{\alpha_{1}y}\mu'f_{S}(\mu'y)\left[\int_{0}^{\infty}L_{T}(x)dx\right]dy\\
% =& \int_{0}^{\infty}e^{\alpha_{1}y}\mu'f_{S}(\mu'y)dy{\color{red} \int_{0}^{\infty}\sup_{\lambda\in[\lambda(\overline{p}),\lambda(\underline{p})]}|f_T(\lambda x)+ \lambda xf_T^{\prime}(\lambda x)|dx}
% % \leq& C|\lambda-\lambda'|\int_{0}^{\infty}e^{\alpha_{1}y}\mu'f_{S}(\mu'y)dy\\
% % \leq& C|\lambda-\lambda'|
% \end{align*}

We next bound \eqref{eq:prob_term}.
Let $h_{\theta}$ denote the density of $\frac{S}{\mu}-\frac{T}{\lambda}$. $h_{\theta}$
is Lipschitz since it is a convolution of Lipschitz densities:
\begin{align*}
h_{\theta}(x) & 
%=(f_{T/\mu}*f_{-S/\lambda})(x)
=\int_{0}^{\infty}\mu f_S(\mu(x+t))\lambda f_T(\lambda t)dt
\end{align*}
%We next develop a Lipschitz bound for $\mathbb{P}(\frac{S}{\mu}-\frac{T}{\lambda}\leq-W_{0})$.
Then,
\begin{align*}
 & \left|\mathbb{P}\left(\frac{S}{\mu}-\frac{T}{\lambda}\leq-w_{0}\right)-\mathbb{P}\left(\frac{S}{\mu'}-\frac{T}{\lambda'}\leq-w_{0}\right)\right|\\
 \leq&\int_{-\infty}^{-w_{0}}|h_{\theta}(x)-h_{\theta}(x)|dx\\
 =&\int_{-\infty}^{-w_{0}}\left|\int_{0}^{\infty}\mu f_S(\mu(x+t))\cdot\lambda f_T(\lambda t)dt-\int_{0}^{\infty}\mu'f_S(\mu'(x+t))\cdot\lambda'f_T(\lambda't)dt\right|dx\\
 \leq&\int_{-\infty}^{-w_{0}}\int_{0}^{\infty}|\mu f_S(\mu(x+t))-\mu'f_S(\mu'(x+t))|\lambda f_T(\lambda t)dtdx\\
 & +\int_{-\infty}^{-w_{0}}\int_{0}^{\infty}|f_T(\lambda t)-f_T(\lambda't)|\mu'f_S(\mu'(x+t))dtdx\\
\leq&|\mu-\mu'|\int_{-\infty}^{-w_{0}}\int_{0}^{\infty}L_{S}(x+t)\lambda f_T(\lambda t)dtdx\\
 & +|\lambda-\lambda'|\int_{-\infty}^{-w_{0}}\int_{0}^{\infty}L_{T}(t)\mu'f_S(\mu'(x+t))dtdx\\
\leq& C(|\mu-\mu'|+|\lambda-\lambda'|).
\end{align*}
using the previous bounds for $L_{S}(x)$ and $L_{T}(t)$.
%Since $L_{S}$ and $L_{T}$ are both exponentially integrable as shown above. Finally, $\lambda(p)$ is Lipschitz as it is $C^{1}$ on $[\underline{p},\bar{p}]$ so there exists a uniform upper bound $\sup_{p\in[\underline{p},\bar{p}]}|\lambda'(p)|<\infty$.
\Halmos
\endproof

\begin{lemma}[Smoothness of the gradients]
There exists a constant $L\in(0,\infty)$ such that
\begin{align*}
&\|\nabla\ell(\mu,p)-\nabla\ell(\mu,p)\| \leq L\left(|\mu-\mu'|+|p-p'|\right),\\
%||\mathbb{E}[g(\mu,p,z)-g(\mu',p',z)|| & \leq L\left[|\mu-\mu'|+|p-p'|\right]\\
&\|g(\mu,p,z)-g(\mu,p,z')\| \leq Ld_{V}(z,z').
\end{align*}
\end{lemma}

\proof{Proof.}
% The first is a result of Assumption 1 that ensures Lipschitzness of
% the gradient (since $c$ and $\lambda$ are $\mathcal{C}^{2}$ on
% compact sets so have upper bounds on the Hessian), and uniform boundedness
% of $\mathbb{E}[W_{n}]$ and $\mathbb{E}[X_{n}]$ for all $n$ and
% for any $(\mu,p)$. 
Recall the characterization of $\nabla l$ and $g$ in \eqref{eq:gg1_gradient0} and \eqref{eq:gg1_gradient}.
For the first bound, we first note that under Assumption \ref{assu:function-assumption}, 
%can be derived based on Assumption 1, which gives uniform upper bounds for 
$c''$ and $\lambda''$ are uniformly bounded. 
%as well as Lipschitzness of $\mathbb{E}[W_{\infty}(\mu,p)]$ and $\mathbb{E}[X_{\infty}(\mu,p)]$.
Second, the Lipschitz continuity of $\mathbb{E}[W_{\infty}(\mu,p)]$ has been established in Lemma 4 of \citet{chen2023online}.
We next establish the Lipschitz continuity of
$\mathbb{E}[X_{\infty}(\mu, p)]$. For this, we leverage the result from \citet{rudolf2018perturbation}, which shows that Lipschitzness of the one-step
transition function and Wasserstein ergodicity implies the Lipschitzness of the stationary distribution. 
%Let $Z_{\infty}(\theta)=(W_{\infty}(\mu,p),X_{\infty}(\mu,p))$ for $\theta=(\mu,p)$. 
In particular, by Theorem 3.1 of \citet{rudolf2018perturbation},
\[
d_{V}(Z_{\infty}(\theta),Z_{\infty}(\theta'))\leq
\frac{K\Gamma}{(1-\rho)^2}\|\theta-\theta'\|
\]
% \[
% d_{V}(Z_{\infty}(\theta),Z_{\infty}(\theta'))\leq\left(\frac{\Gamma\cdot CK}{(1-\rho)}+\frac{\Gamma}{1-\rho}\right)\|\theta-\theta'\|,
% \]
where $\Gamma$ is the Lipschitz constant for the one-step transition kernel:
\[
\Gamma=\sup_{z}\sup_{\theta,\theta'}\frac{d_{V}(\delta_{z}P_{\theta},\delta P_{\theta'})}{V(z)\|\theta-\theta'\|},
\]
which was established in Lemma \ref{lm:smooth_P},
$\rho$ is the Lyapunov drift in Lemma \ref{lm:lyapunov} and the Wasserstein contraction rate in Lemma \ref{lm:wasserstein}, and $K$ is the constant term in
the Wasserstein contraction in Lemma \ref{lm:wasserstein}. Since $X\leq Ae^{\alpha_{1}W+\alpha_{2}X}$ for
some constant $A\in(0,\infty)$ large enough, we have
\begin{align*}
|\mathbb{E}[X_{\infty}(\theta)-X_{\infty}(\theta')]| & \leq Ad_{V}(Z_{\infty}(\theta),Z_{\infty}(\theta'))\\
 & \leq A \frac{K\Gamma}{(1-\rho)^2}\|\theta-\theta'\|.
 %\left(\frac{\Gamma\cdot CK}{(1-\rho)}+\frac{\Gamma}{1-\gamma}\right)||\theta-\theta'||
\end{align*}

For the second bound, since $g$ is linear in $z$ and there exists $L$ large enough so that 
\[\begin{split}
\|z-z'\|_{1}&\leq(|w|+|x|+|w'|+|x'|)1\{z\neq z'\}\\
&\leq L\left(e^{\alpha_{1}|w|+\alpha_{2}|x|}+e^{\alpha_{1}|w'|+\alpha_{2}|x'|}\right)1\{z\neq z'\},
\end{split}\]
we have
$\|g(\mu,p,z)-g(\mu,p,z')\|\leq Ld(z,z')$.
\Halmos
\endproof

We next verify Assumption \ref{ass:bound0}, i.e., bounds of the gradients.
%\paragraph{Uniform Upper bound on Gradient}
\begin{lemma}[Bounds of the gradients]
%The gradients are uniformly bounded, $\sup_{t}\mathbb{E}[||g(\mu,p,Z_{t})||_{2}]<\infty$ (Chen at al assumes $W_0=0$)
There exists $M\in(0,\infty)$, such that
$\|g(\mu,p,z)\|\leq MV(z)$
and
$\|\nabla\ell(\mu,p)\|\leq M$.
\end{lemma}

\proof{Proof.}
Since $e^x\leq 1+x$, under Assumption \ref{assu:function-assumption}, there exists $M>0$, such that
\[
\begin{split}
g_{p}(\mu,p,z) & =-\lambda(p)-p\lambda'(p)+h_{0}\lambda'(p)\left(w+x+\frac{1}{\mu}\right)\\
&\leq M\exp(\alpha_1 w + \alpha_2 x)
\end{split}
\]
and
\[
\begin{split}
g_{\mu}(\mu,p,Z_{t}) & =c'(\mu)-h_{0}\frac{\lambda'(p)}{\mu}\left(W_{t}+X_{t}+\frac{1}{\mu}\right)\\
&\leq M\exp(\alpha_1 w + \alpha_2 x)
\end{split}
\]

Next, since $\mathbb{E}[W_{\infty}(\mu,p)]<\infty$ and $\mathbb{E}[X_{\infty}(\mu,p)]<\infty$, under Assumption \ref{assu:function-assumption}, there exists $M>0$, such that
\[
\frac{\partial}{\partial p}\ell(\mu,p)  =-\lambda(p)-p\lambda'(p)+h_{0}\lambda'(p)\left(\mathbb{E}[W_{\infty}(\mu,p)]+\mathbb{E}[X_{\infty}(\mu,p)]+\frac{1}{\mu}\right)\leq M
\]
and
\[
\frac{\partial}{\partial\mu}\ell(\mu,p) =c'(\mu)-h_{0}\frac{\lambda'(p)}{\mu}\left(\mathbb{E}[W_{\infty}(\mu,p)]+\mathbb{E}[X_{\infty}(\mu,p)]+\frac{1}{\mu}\right)
\leq M.
\]

% By similar arguments in Lemma 1 in \citet{chen2023online}, note that
% for any feasible $(\mu,p)$, by coupling $W_{\infty}(\mu,p)\leq_{st}W_{\infty}(\underline{\mu},\underline{p})$
% and $W_{n}\leq_{st}W_{\infty}(\mu,p)$ by another coupling argument
% considering a workload process starting from an infinite past. As
% for $X$, let $\tau_{0,t}(\mu,p)=\sup\{i\leq t:W_{i}=0\}$ be the
% last emptying time before $n$ under $(\mu,p)$. $X_{n}$ can be expressed
% as the sum of interarrival epochs from $\tau_{0,t}(\mu,p)$ so we
% have

% \begin{align*}
% X_{t} & =\sum_{i=\tau_{0,t}(\mu,p)}^{t}\frac{T_{i}}{\lambda(p)}\leq_{st}\sum_{i=\tau_{0,t}(\mu,p)}^{t}\frac{T_{i}}{\lambda(p)}\\
%  & \leq_{st}\sum_{i=\tau_{0,t}(\underline{\mu},\underline{p})}^{n}\frac{T_{i}}{\lambda(\bar{p})}\\
%  & =\frac{\lambda(\underline{p})}{\lambda(\bar{p})}\sum_{i=\tau_{0,t}(\underline{\mu},\underline{p})}^{t}\frac{T_{i}}{\lambda(\underline{p})}\\
%  & \leq_{st}\frac{\lambda(\underline{p})}{\lambda(\bar{p})}X_{\infty}(\underline{\mu},\underline{p})
% \end{align*}
% This will imply that for all $t$, $\mathbb{E}[W_{t}]\leq\mathbb{E}[W_{\infty}(\underline{\mu},\underline{p})]$
% and $\mathbb{E}[X_{t}]\leq\mathbb{E}[\frac{\lambda(\underline{p})}{\lambda(\bar{p})}X_{\infty}(\underline{\mu},\underline{p})]$,
% which in turn implies that $\sup_{t}\mathbb{E}[||g(\mu,p,Z_{t})||_{2}]\leq\infty$.
\Halmos
\endproof

Lastly, since we can sample directly from $g$, no further approximation, $\hat g_t$, is needed, i.e., Assumption \ref{ass:error0} holds trivially. This concludes the proof of Theorem \ref{thm:newsvendor-conv}.

\section{Proofs of the results in Section \ref{sec:RL}}

\subsection{Proofs of Propositions \ref{prop:policy_gradient_ergo} and \ref{prop:policy_graident_stationary}}
We first present and prove some auxiliary lemmas.

Let $\tau_{cov}$ be the first time at which $s_t$ have visited all the states. Then we define define the cover time \citep{levin2017markov} as
\[
t_{cov}=\max_{s\in\mathcal{S}}\E_x[\tau_{cov}].
\] 
Note that from Theorem 11.2 in \citep{levin2017markov}
\[
t_{cov} \leq t_{hit}\sum_{k=1}^{|\mathcal{S}|-1}\frac{1}{k}.
\]
Similarly, we define $\hat t_{hit}$ and $\hat t_{cov}$ as the hitting time and cover time of the finite-state Markov chain $(s_t, \hat a_t)$ respectively.
\begin{lemma}
\label{lem:tcov}
Suppose the Markov chain $s_t$ has a finite hitting time $t_{hit}$. Then, the cover time of the Markov chain $(s_t,\hat a_t)$ satisfies
\[
\hat t_{cov} \leq (1+t_{hit})|\mathcal{A}|\sum_{k=1}^{|\mathcal{S}||\mathcal{A}|-1}\frac{1}{k}
\]
\end{lemma}
\proof{Proof.}
Define $\hat{\kappa}_{s,\hat{a}}=\min\{t\geq 0: s_t=s, \hat a_t=\hat a\}$.
Consider two arbitrary states $(s,\hat a)$ and $(s', \hat a')$ and we are interested in bounding $\E_{s,\hat a}[\hat{\kappa}_{s',\hat{a}'}]$. Let $\zeta_0=0$, and $\zeta_k,k\geq 1$ be the sequence of stopping time defined as
$\zeta_k=\inf\{t>\zeta_{k-1}: s_t=s'\}$. We also write $\Delta \zeta_k=\zeta_{k+1}-\zeta_{k}$. Since $\hat a_t$ are sampled uniformly at random from $\mathcal{A}$, independent of $s_t$, we have
\[
\hat{\kappa}_{s',\hat{a}'}\overset{d}{=}\sum_{k=1}^{N}\Delta \zeta_k,
\]
where $N$ is a Geometric random variable with probability Of success $1/|\mathcal{A}|$ and is independent of $\Delta\zeta_k$.
In addition, note that $\E_{s}[\Delta\zeta_1]\leq t_{hit}$ and $\E_{s'}[\Delta\zeta_k]\leq 1+t_{hit}$ for $k\geq 2$.
Then,
\[
\E_{s,\hat a}[\hat{\kappa}_{s',\hat{a}'}]\leq \E[N] (1+t_{hit})=(1+h_{hit})|\mathcal{A}|.
\]
Thus, $\hat t_{hit}\leq (1+t_{hit})|\mathcal{A}|$.
Then, the bound for the cover time follows from Theorem 11.2 in \citep{levin2017markov}.
\Halmos
\endproof

\begin{lemma} \label{lm:tau_K}
For a sequence of stopping times $\{\tau_k\}_{k\geq0}$ with $N(t)=\sup_k \{\tau_k\leq t\}$, if there exists $t_0>0$, such that $\mathbb{P}_{t}(N(t+t_0)>N(t))\leq \frac12$,
%$P_{t}(N(t+t_0)>N(t))\geq 1/2$,
then 
\[
\mathbb{P}_{\tau_0}(\tau_K-\tau_0>12Kt_0)\leq \exp(-K).
\]
\end{lemma}
\proof{Proof.}
Without loss of generality, we assume $\tau_0=0$.
Let $A_m=1_{N(m t_0)>N((m-1)t_0)}$. 
Then, 
\[\E_{(m-1)t_0}A_m=\mathbb{P}_{(m-1)t_0}(N(m t_0)>N((m-1)t_0))\geq 1/2.\] 
Next note that $N(12Kt_0)\geq \sum_{m=1}^{12K} A_m$ and $|A_m-\E_{(m-1)t_0}A_m|<1$. Then,
%Next, we have {\color{blue} Use Hoeffding here.}
\[\begin{split}
\mathbb{P}_{\tau_0}(\tau_K-\tau_0>12Kt_0)=&\mathbb{P}(N(12Kt_0)<K)\\
\leq& \mathbb{P}\left(\sum_{m=1}^{12K} A_m<K\right) \\
\leq& \mathbb{P}\left(\sum_{m=1}^{12K}(A_m - \E_{(m-1)t_0}A_m)< K - 12K\frac{1}{2}\right)\\
\leq& \exp\left(-\frac{(-5K)^2}{2\times 12K}\right) \mbox{by Azuma's inequality}\\
\leq &\exp(-K).
\end{split}\]
\Halmos
\endproof

\begin{lemma} \label{lm:W}
Given two matrices $A$ and $B$, if $\text{det}(A+\alpha B)=0$ for all $\alpha$, then there exist nonzero $W_0$ and $W_1$  such that that $AW_0=0$ and $AW_1+BW_0=0$. 
\end{lemma}
\proof{Proof.}
Let $A=U\Lambda V^T$ denote the singular value decomposition of $A$. Suppose the first $r$ entries of $\Lambda$ are zero.   Let $C=U^TBV$. We write
\[
\Lambda=\begin{bmatrix} 0 & 0\\ 0 &\Lambda_4\end{bmatrix},\quad C=\begin{bmatrix} C_1 &C_2\\ C_3 &C_4\end{bmatrix}
\]
where $\Lambda_4$ is invertible. Then, there exists $\alpha_0$ such that for all $\alpha\leq \alpha_0$, $[\Lambda_4+\alpha C_4]^{-1}$ exists and  its $L_2$ norm is smaller than a constant. Moreover,
\[
0=\text{det}(A+\alpha B)=\text{det}(\Lambda+\alpha C)=\text{det}(\Lambda_4+\alpha C_4)
\text{det}(\alpha C_1-\alpha^2C_2[\Lambda_4+\alpha C_4]^{-1} C_3). 
\]
This implies that
\[
\text{det}(C_1-\alpha C_2[\Lambda_4+\alpha C_4]^{-1} C_3)= 0,
\]
which further implies $\text{det}(C_1)=0$. Let $w_0$ be a nonzero $r$-dimensional vector such that $C_1 w_0=0$. Next let $w_1=-\Lambda_4^{-1}C_3 w_0$ which is a $n-r$ dimensional vector. Next, extend $w_0,w_1$ to an $n$-dimension vector $\bar w_0 =[w^T_0, 0, \dots, 0]^T, \bar w_1 =[ 0, \dots, 0, w_1^T]^T$. Let $W_i=V\bar w_i,i=0$ and $1$, then 
\[
AW_0=U\Lambda V^T V\bar w_0=0,\quad AW_1=U\Lambda V^T V\bar w_1=U[0,(\Lambda_4 w_1)^T]^T
\]
\[
BW_0=UC V^T V\bar w_0=U[0,(C_3w_0)^T ]^T, 
\]
which leads to our conclusion.
\Halmos
\endproof

\proof{Proof of Proposition \ref{prop:policy_gradient_ergo}.}
Consider running two coupled $Z_t$ and $\tilde Z_t$ under $\pi^{\theta}$, where $\tilde Z_0\sim \mu^\theta$, 
and the transition is coupled so that 
\[
(\hat{a}_t,a_t,s_{t+1},s'_{t+1}, a'_{t+1})=(\tilde{\hat{a}}_t,\tilde{a}_t,\tilde{s}_{t+1},\tilde{s}'_{t+1}, \tilde{a}'_{t+1})
\]
if $s_t=\tilde{s}_t$.
First, there exists a stopping time $\tau_0:=\min\{t\geq 0: s_t=\tilde s_t\}$, since $s_t$ is ergodic under $\pi^\theta$. Note that under the coupling, $(s_t, \hat a_t, a_t)=(\tilde s_t, \tilde{\hat{a}}_t, \tilde a_t)$ for $t\geq \tau_0$. In addition,
%Let 
% \[
% \xi_{\epsilon}=\frac{|\log (\epsilon/(8M+4))|}{|\log (1/4)|} t_{mix}.
% \] Then, 
\[
\mathbb{P}\left(\tau_0\geq \frac{|\log (\epsilon/(8M+4))|}{|\log (1/4)|} t_{mix}\right)\leq \frac{\epsilon}{4(2M+1)}.
\]
Let $\Delta_t:=Q_t-\tilde Q_t$. Then, for $t\geq \tau_0$,
\[
\Delta_{t+1}(s_t,\hat{a}_t)
=(1-\alpha)\Delta_t(s_t,\hat{a}_t)+\gamma \alpha \Delta_t(s'_{t+1},a'_{t+1}).
\]
Consider a sequence of covering times, $\{\tau_k\}_{k\geq 1}$, where $\tau_k$ is the time it takes $(s_t, \hat{a}_t)$ to visit all the state-action pairs at least $k$ times:
\[
\tau_k=\min \left\{t>\tau_{k-1} \text{ s.t. for any } (s,a)\in \mathcal{S}\times\mathcal{A} \text{ there is a } u\in(\tau_{k-1}, t] \text{ s.t. } s_u=s, \hat{a}_u=a\right\}.
\]
%We also define $N(t)=\sup_k\{\tau_k \leq t\}$

We next use induction to show that 
\[
\|\Delta_t\|_\infty\leq (1-(1-\gamma )\alpha)^{k-1}\|\Delta_{\tau_0}\|_\infty,\quad \forall \tau_k< t\leq \tau_{k+1}.
\]
The claim holds trivially when $k=0$. Suppose it is true for $\tau_k$, then for $\tau_k<t<\tau_{k+1}$, we first note that if $(s,a)\neq (s_t,\hat{a}_t)$, $\Delta_{t+1}(s,a)= \Delta_{t}(s,a)$. If $(s,a)=(s_t,\hat{a}_t)$,
\[
\Delta_{t+1}(s,a)\leq (1-\alpha+\alpha\gamma )\|\Delta_t\|_\infty
%\leq (1-a+a\gamma )\|\Delta_{\tau_k}\|_\infty
\]
This indicates that $\|\Delta_{t+1}\|_\infty\leq \|\Delta_{t}\|_\infty$. Thus,
\[
\Delta_{t+1}(s,a)\leq (1-\alpha+\alpha\gamma )\|\Delta_t\|_\infty
\leq (1-\alpha+\alpha\gamma )\|\Delta_{\tau_k}\|_\infty.
\]
This further indicates that when all state-action pairs are visited at least once after $\tau_k$, 
We have 
\[\|\Delta_{\tau_{k+1}}\|_\infty \leq (1-\alpha+\alpha\gamma )\|\Delta_{\tau_k}\|_\infty.\] 

%First, notice that by considering if $|\Delta_{t}(s_t,a_t)|=\|\Delta\|_\infty$, it is easy to conclude that  
%\[
%\|\Delta_{t+1}\|_\infty\leq \|\Delta_{t}\|_\infty.
%\]
%So $\|\Delta_{t}\|_\infty\leq \|\Delta_{\tau_k}\|_{\infty}$. 
%Second, we note that $\Delta_{t+1}(a,s)\neq \Delta_{t}(a,s)$ if and only if $(a_t,s_t)=(a,s)$. In this case, we have 
%\[
%\Delta_{t+1}(a,s)\leq (1-a+a\gamma )\|\Delta_t\|_\infty\leq 
%(1-a+a\gamma )\|\Delta_{\tau_k}\|_\infty
%\]
%In other words, when all state action pairs are visisted at least once after $\tau_k$, the $l_\infty$ norm of $\Delta_t$ will be reduced to $\|\Delta_{\tau_{k+1}}\|_\infty \leq (1-a+a\gamma )\|\Delta_{\tau_k}\|_\infty$. 

Next, let $N(t)=\sup_k \{\tau_k\leq t\}$. 
Recall that $\hat t_{cov}$ is the cover time of the Markov chain $(s_t,\hat a_t)$.
%$\hat{\kappa}_{s,\hat{a}}=\min\{t\geq 0: s_t=a, \hat a_t=\hat a\}$ and
%the hitting time of $s_t=s,\hat{a}_t=\hat{a}$ as $\hat{\kappa}_{s,\hat{a}}$ 
%\[
%t'_{cov}=\max_{a,s,s',\hat{a}'}\E_{s,a}[\hat{\kappa}_{s',\hat{a}'}]. 
%\]
Since $\E_{t}[\tau_{N(t)+1}-t]\leq \hat t_{cov}$, 
\[
\hat t_{cov} \geq \sum_{k>2\hat t_{cov}} k \mathbb{P}_t(\tau_{N(t)+1}-t = k) \geq 2\hat t_{cov} \mathbb{P}_t(\tau_{N(t)+1}-t > 2\hat t_{cov}),
\]
which implies that 
\[
\mathbb{P}_{t}(\tau_{N(t)+1}-t>2\hat t_{cov})\leq \frac12
~~\mbox{ and }~~
\mathbb{P}_{t}(N(t+2\hat t_{cov})>N(t))\geq \frac12.
\]
%$\mathbb{P}_{\tau_k}(\tau_{k+1}-\tau_k>2t_{cov})\leq \frac12$.
Then, by Lemma \ref{lm:tau_K}, $\mathbb{P}((\tau_K-\tau_0)\geq 12K\hat t_{cov})\leq e^{-K}$. 
Since $\tilde d(Z,\tilde Z) \leq 1+2M$ and $\|\Delta_{\tau_0}\|_\infty\leq 2M$, for any
\[
\bar \eta_\epsilon \geq %\frac{|\log \epsilon|}{|\log (1/4)|} 
\frac{|\log (\epsilon/(8M+4))|}{|\log (1/4)|}t_{mix}\\
+\max\left\{\frac{|\log (\epsilon/(4M))|}{|\log (1-\alpha+\alpha\gamma)|}, 12|\log(\epsilon/(8M+4))|\right\}\hat t_{cov},
\]
we have
\[\begin{split}
d(Z_{\bar \eta_\epsilon}, \tilde Z_{\bar \eta_\epsilon})\leq& (2M+1)\mathbb{P}\left(\tau_0>\frac{|\log (\epsilon/(8M+4))|}{|\log (1/4)|} t_{mix}\right)\\
&+ (2M+1)\mathbb{P}(\tau_K-\tau_0\geq 12|\log(\epsilon/(8M+4))|\hat t_{cov})\\
&+2M(1-(1-\gamma )\alpha)^{\frac{|\log (\epsilon/(4M))|}{|\log (1-\alpha+\alpha\gamma)|}}\\
\leq& (2M+1)\frac{\epsilon}{4(2M+1)}+(2M+1)\frac{\epsilon}{4(2M+1)}+\frac{\epsilon}{4M}2M
\leq \epsilon.
\end{split}\]
Lastly, by Lemma \ref{lem:tcov}, $\hat t_{cov} \leq (1+t_{hit})|\mathcal{A}|\sum_{k=1}^{|\mathcal{S}||\mathcal{A}|-1}1/k$.
\Halmos
\endproof

\proof{Proof of Proposition \ref{prop:policy_graident_stationary}.}
The first claim is due to the fact that $x_t=(s_t,a_t)$ is a Markov chain with invariant measure $\nu^\theta$. 

For the second claim, to simplify the notation, we denote $x,y$ as two state-action pairs.
%=(s,a)$.
Let
\begin{equation}\label{eq:W}
W(y,x):=%\E_{\mu^\theta}[Q(x')|x]=
\frac{1}{\nu^{\theta}(x)}\sum_Q\mu^{\theta}(x,Q) Q(y), 
\end{equation}
which can be seen as an $|\mathcal{S}||\mathcal{A}|$-dimensional vector. 

For fixed $x$ and $y$, we consider a test function $G(Q,s,a)=Q(y)1_{(s,a)=x}$. Under the invariant measure, we should have 
\begin{equation}\label{eq:eq_part1}
\begin{split}
\E_{z_t\sim\mu^{\theta}}[G(Q_{t+1},s_{t+1},a_{t+1})]&=\E_{z_t\sim\mu^{\theta}}[G(Q_{t},s_{t},a_{t})]\\
&=\sum_{Q}\mu^{\theta}(x,Q) Q(y)=\nu^{\theta}(x)W(y,x).
\end{split}
\end{equation}
For clarity with notation, let $y = (s_{y}, a_{y})$ and let $\tilde{x} = (\tilde{s}, \tilde{a})$. Let $\hat{a}$ denote the random action sampled for the TD learning step (sampled uniformly at random from $\mathcal{A}$. We can expand the one-step expectation as
%the left-hand side of \eqref{eq:eq_part1} can be written as
\begin{equation}
\label{eq:eq_part2}
\begin{split}
&\E_{z_t\sim\nu^{\theta}}[ G(Q_{t+1},s_{t+1},a_{t+1})]\\
=&\sum_{\tilde x,Q}\mu^{\theta}(x_t=\tilde x,Q_t=Q)
\Big[(1_{\tilde{s} \neq s_{y}} + 1_{\tilde{s} = s_{y}} \mathbb{P}(\hat{a} \neq a_{y}))
\mathcal{P}^{\theta}(\tilde x,x) Q(y)\\
&+1_{\tilde{s}= s_{y}}\mathbb{P}(\hat{a} = a_{y})\mathcal{P}^{\theta}(y,x)((1-\alpha)Q(y)+\alpha r(y)+\alpha \gamma \E^\theta[Q(s'_{t+1},a'_{t+1})])\Big]\\
=&\sum_{\tilde x,Q}\mu^{\theta}(\tilde x,Q)\mathcal{P}^{\theta}(\tilde x, x) Q(y)\\
&-\alpha \frac{1}{|A|} \sum_{\tilde{x}:\tilde{s} = s_{y}} \sum_{Q}\mu^{\theta}(\tilde{x},Q) \mathcal{P}^{\theta}(\tilde{x},x) \Big[Q(y)-r(y)-\gamma  \sum_{x'}\mathcal{P}^{\theta}(y,x')Q(x')\Big]\\
=&\sum_{\tilde x}\nu^{\theta}(\tilde x)W(y,\tilde x)\mathcal{P}^{\theta}(\tilde x, x)\\
&-\alpha \frac{1}{|A|}\sum_{\tilde{x}:\tilde{s} = s_{y}} \nu^{\theta}(\tilde{x}) \mathcal{P}^{\theta}(\tilde{x},x) \Big[W(y,\tilde{x})-r(y)-\gamma \sum_{x'} \mathcal{P}^{\theta}(y,x')W(x',y)\Big]
\end{split}
\end{equation}

% \begin{equation}\label{eq:eq_part2}
% \begin{split}
% &\E_{z_t\sim\nu^{\theta}}[ G(Q_{t+1},s_{t+1},a_{t+1})]\\
% =&\sum_{\tilde x,Q}\mu^{\theta}(x_t=\tilde x,Q_t=Q)\Big[1_{\tilde x\neq y}\mathcal{P}^{\theta}(\tilde x,x) Q(y)\\
% &+1_{\tilde x=y}\mathcal{P}^{\theta}(y,x)((1-\alpha)Q(y)+\alpha r(y)+\alpha \gamma \E^\theta[Q(s'_{t+1},a'_{t+1})])\Big]\\
% =&\sum_{\tilde x,Q}\mu^{\theta}(\tilde x,Q)\mathcal{P}^{\theta}(\tilde x, x) Q(y)\\
% &-\alpha  \sum_{Q}\mu^{\theta}(y,Q) \mathcal{P}^{\theta}(y,x) \Big[Q(y)-r(y)-\gamma  \sum_{x'}\mathcal{P}^{\theta}(y,x')Q(x')\Big]\\
% =&\sum_{\tilde x}\nu^{\theta}(\tilde x)W(y,\tilde x)\mathcal{P}^{\theta}(\tilde x, x)\\
% &-\alpha \mathcal{P}^{\theta}(y,x)\nu^{\theta}(y) \Big[W(y,y)-r(y)-\gamma \sum_{x'} \mathcal{P}^{\theta}(y,x')W(x',y)\Big]
% \end{split}
% \end{equation}

% {\color{blue}
% \[\begin{split}
% &\E_{z_t\sim\nu^{\theta}}[ G(Q_{t+1},s_{t+1},a_{t+1})]\\
% =&\sum_{\tilde x}\nu^{\theta}(\tilde x)W(y,\tilde x)\mathcal{P}^{\theta}(\tilde x, x)\\
% &-\alpha\epsilon \epsilon \mathcal{P}^{\theta}(y,x)\nu_s^{\theta}(y) \Big[W(y,y)-r(y)-\gamma \sum_{x'} \mathcal{P}^{\theta}(y,x')W(x',y)\Big],
% \end{split}\]
% where $\epsilon=1/|\mathcal{A}|$ and $\nu_s^{\theta}(y)=\sum_{a\in\mathcal{A}}\nu^{\theta}(y_1,a)$.
% }

Putting \eqref{eq:eq_part1} and \eqref{eq:eq_part2} together, we have 

\begin{equation}
\label{tmp:TDcond}
\begin{split}
&\nu^{\theta}(x)W(y,x)\\
=&\sum_{\tilde x}\nu^{\theta}(\tilde x)W(y,\tilde x)\mathcal{P}^{\theta}(\tilde x, x)\\
&-\alpha \frac{1}{|A|}\sum_{\tilde{x}:\tilde{s} = s_{y}} \nu^{\theta}(\tilde{x}) \mathcal{P}^{\theta}(\tilde{x},x) \Big[W(y,\tilde{x})-r(y)-\gamma \sum_{x'} \mathcal{P}^{\theta}(y,x')W(x',y)\Big]
\end{split}
\end{equation}
%
% \begin{equation}
% \label{tmp:TDcond}
% \begin{split}
% \nu^{\theta}(x)W(y,x)=&\sum_{\tilde x}\nu^{\theta}(\tilde x)W(y,\tilde x)\mathcal{P}^{\theta}(\tilde x, x)\\
% &-\alpha \mathcal{P}^{\theta}(y,x)\nu_s^{\theta}(y) [W(y,y)-r(y)-\gamma \sum_{x'} \mathcal{P}^{\theta}(y,x')W(x',y)].
% \end{split}
% \end{equation}
% {\color{blue}
% \[
% \begin{split}
% \nu^{\theta}(x)W(y,x)=&\sum_{\tilde x}\nu^{\theta}(\tilde x)W(y,\tilde x)\mathcal{P}^{\theta}(\tilde x, x)\\
% &-\alpha \mathcal{P}^{\theta}(y,x)\nu_s^{\theta}(y) [W(y,y)-r(y)-\gamma \sum_{x'} \mathcal{P}^{\theta}(y,x')W(x',y)].
% \end{split}
% \]
% (I think the rest of the proof will go through with this new system of equations. It is easy to verify that $W(y,x)=Q^{\theta}(y)$ is still a solution. We just need to double-check the uniqueness of the solution.)}
%which is a system of linear equations. 
By enumerating all possible $x,y$ combinations, we have a system of $|\mathcal{S}\times \mathcal{A}|^2$ linear equations. Then, it suffices to show this linear equation system has a unique solution $W$ and verify that $W(y,x)=Q^{\theta}(y)$ is a solution. 

To show the solution is unique, we note that the system of linear equations can be written as 
\[
AW-\alpha BW=\alpha b, 
\]
where $A,B$ are $|\mathcal{S}\times \mathcal{A}|^2\times |\mathcal{S}\times \mathcal{A}|^2$ matrices, and $b$ is an $|\mathcal{S}\times \mathcal{A}|^2$-dimensional vector. For such a system to have multiple solutions, $f(\alpha):=\text{det}(A-\alpha B)$ has to be zero. Note that since $A-\alpha B$ is linear in $\alpha$, $f(\alpha)$ is a polynomial function of $\alpha$. This implies that $f(\alpha)$ either has finitely many roots or $f(\alpha)\equiv 0$. 
%We will rule out the second case. Because other wise by linear algebra result, there is a nonzero $W$ so that 
If $f(\alpha)\equiv 0$, then there exist nonzero $W_0$ and $W_1$ such that
$AW_0=0$ and $AW_1+BW_0=0$ (see Lemma \ref{lm:W}).  $AW_0=0$ implies that for all $x,y$,
\[
\nu^{\theta}(x)W_0(y,x)-\sum_{\tilde x}\nu^{\theta}(\tilde x)W_0(y,\tilde x)\mathcal{P}^{\theta}(\tilde x, x)=0, 
\]
which indicates that 
$W_0(y,x)=W_0(y,\tilde x)$ for all $\tilde x$. We write $W_0(y):=W_0(y,x)$. 
We plug this into the second equation $AW_1+BW_0=0$, and obtain for all $x,y$

\[\begin{split}
&\nu^{\theta}(x)W_1(y,x)-\sum_{\tilde x}\nu^{\theta}(\tilde x)W_1(y,\tilde x)\mathcal{P}^{\theta}(\tilde x, x)\\
&+\frac{1}{|A|}\sum_{\tilde{x}:\tilde{s} = s_{y}} \nu^{\theta}(\tilde{x}) \mathcal{P}^{\theta}(\tilde{x},x) \Big[W_{0}(y)-\gamma \sum_{x'} \mathcal{P}^{\theta}(y,x')W_{0}(x')\Big] = 0
\end{split}\]

% \[\begin{split}
% &\nu^{\theta}(x)W_1(y,x)-\sum_{\tilde x}\nu^{\theta}(\tilde x)W_1(y,\tilde x)\mathcal{P}^{\theta}(\tilde x, x)\\
% &+\mathcal{P}^{\theta}(y,x)\nu^{\theta}(y) [W_0(y)-\gamma \sum_{x'}\mathcal{P}^{\theta}(y,x') W_0(x')]=0.
% \end{split}\]
For each fixed $y$, we sum the equations above for all $x$ and obtain
\[
\frac{1}{|A|}\nu^{\theta}_{s}(s_{y})[W_0(y)-\gamma \sum_{x'}\mathcal{P}^{\theta}(y,x') W_0(x')]=0.
\]

% \[
% \nu^{\theta}(y) [W_0(y)-\gamma \sum_{x'}\mathcal{P}^{\theta}(y,x') W_0(x')]=0.
% \]
This gives
\[
W_0(y)=\gamma \sum_{x'} \mathcal{P}^{\theta}(y,x')W_0(x').
\]
Let $y=\arg\max_x W_0(x)$, then we have $W_0(y)\equiv 0$, which is a contradiction. 

We next verify that $W(y,x)=Q^\theta(y)$ is a solution. Since
\[
\nu^{\theta}(x)=\sum_{\tilde x}\nu^{\theta}(\tilde x)\mathcal{P}^{\theta}(\tilde x, x) \mbox{ and }
Q^\theta(y)=r(y)+\gamma \sum_{x'} \mathcal{P}^{\theta}(y,x')Q^\theta(x'),
\]
we have
\begin{equation}
\begin{split}
\nu^{\theta}(x)Q^{\theta}(y)=&\sum_{\tilde x}\nu^{\theta}(\tilde x)Q^{\theta}(y)\mathcal{P}^{\theta}(\tilde x, x)\\
&-\alpha \frac{1}{|A|}\sum_{\tilde{x}:\tilde{s} = s_{y}} \nu^{\theta}(\tilde{x}) \mathcal{P}^{\theta}(\tilde{x},x) \Big[Q^{\theta}(y)-r(y)-\gamma \sum_{x'} \mathcal{P}^{\theta}(y,x')Q^{\theta}(x')\Big]
\end{split}
\end{equation}

% \[\begin{split}
% \nu^\theta(x) Q^\theta(y)=&\sum_{\tilde x}\nu^\theta(\tilde x)Q^\theta(y)\mathcal{P}^{\theta}(\tilde x, x)\\
% &-\alpha \mathcal{P}^\theta(y,x)\nu^\theta(y)\Big[Q^\theta(y)-r(y)-\gamma \sum_{x'} \mathcal{P}^\theta(y,x')Q^\theta(x')\Big].
% \end{split}\]

Lastly, note that
\begin{align*}
\E_{\mu^\theta}[Q(s_t, a_t) \nabla_{\theta} \pi^\theta(a_t|s_t)]
&=\sum_{s,a}\nu^\theta(s,a)\E_{\mu^\theta}[Q(s_t, a_t)|s_t=s,a_t=a] \nabla_\theta \pi^\theta(a|s)\\
&=\sum_{s,a}\nu^\theta(s,a)Q^\theta(s,a) \nabla_{\theta} \pi^\theta(a|s).
\end{align*}
\Halmos
\endproof

\subsection{Proof of Theorem~\ref{thm:pg-convergence}} \label{sec:proof-pg-convergence}

We proceed to verify that the assumptions required in Theorem~\ref{thm:main1} hold for the actor-critic update in Algorithm \ref{alg:TD}.

First, it is worth noting that under Assumption~\ref{ass:pg-bounded}, if $Q_{0}$ is initialized to be $\|Q_{0}\|_{\infty} \leq \frac{M}{1-\gamma}$, then $\|Q_{t}\|_{\infty} \leq \frac{M}{1-\gamma}$ almost surely for all $t$, under any policy $\pi^\theta$.

\paragraph{Lyapunov Function.}

%To verify the first inequality for Assumption~\ref{ass:ergodicity0}, 
%We consider a Lyapunov function $V(z)$ 
For the Markov chain $Z_{t} = (s_{t},a_{t},Q_{t})$. we denote $\widetilde{\mathcal{P}}_{\theta}$ as its transition kernal.
%For clarity, we refer to the transition kernel of this Markov chain as $\widetilde{\mathcal{P}}_{\theta}$.
Since there are finitely many states and actions and $\|Q_t\|_{\infty}\leq\frac{M}{1-\gamma}$ almost surely, we can construct a Lyapunov function for which the drift inequality holds trivially:
\[
V(z)=1+\| Q \|_{\infty}
\]
The corresponding drift inequality is
\[
\widetilde{\mathcal{P}}_{\theta}V(z)\leq1+\frac{M}{1-\gamma}
\]
where $\rho=0$ and $K = 1+\frac{M}{1-\gamma}$.
We have thus verified the first part of Assumption \ref{ass:ergodicity0}. Similarly, Assumption  \ref{ass:bound1} also holds trivially.

\paragraph{Wasserstein Ergodicity.}
%To verify the second part of Assumption~\ref{ass:ergodicity0}, 
For $z=(s,a,Q)$, we consider the metric
%show that $Z_{t}=(s_{t},a_{t},Q_{t})$ satisfies Wasserstein ergodicity with the metric %$d(Z,Z')$ given by
\begin{equation}
\label{eq:policy_grad_metric}
\tilde d(z,\tilde z)=1_{(s,a)\neq(\tilde s,\tilde a)}+\|Q-\tilde Q\|_{\infty}.
\end{equation}
We next show that for any $\theta\in\Theta$ and any $z$ and $\tilde z$, 
\[ d(\delta_{z}\widetilde{\mathcal{P}}_{\theta}^{t},\delta_{\tilde z}\widetilde{\mathcal{P}}_{\theta}^{t})
\leq (8M+4)r^t
%\frac{8M+4}{1-\gamma}r^{t}\tilde d(z,\tilde z)
\]
where 
\begin{equation}
\label{eq:pg_ergodicity_constant}
r=\exp\left\{-\left(\frac{1}{\log(1/\gamma)}+\left(\frac{1}{\log(1/(1-\alpha+\alpha\gamma))}+12\right)\frac{(1-\gamma)\min_{s}\rho(s)+1}{(1-\gamma)\min_{s}\rho(s)}|\mathcal{A}|(1+\log(|\mathcal{S}||\mathcal{A}|))\right)^{-1}\right\},
%\exp\left(-\frac{1}{12\left(\frac{2}{\log(1/\gamma)}+\frac{|\mathcal{A}|}{1-\gamma}\min_{s}\frac{1}{\rho(s)}\log\frac{1}{1-\alpha+\alpha\gamma}\right)}\right),
\end{equation}
which verifies the second part of Assumption \ref{ass:ergodicity0}.

We first bound the mixing time and hitting time of $s_t$, which we denote as $t_{\text{mix}}$ and $t_{\text{hit}}$. Recall that $s_t$ is a finite-state Markov chain with the transition kernel
\[
\bar{\mathcal{P}}_{\theta}(s_{t+1}|s_{t}):=\gamma\sum_{a_{t}\in\mathcal{A}}P(s_{t+1}|s_{t}, a_{t})\pi^{\theta}(a_{t}|s_{t})+(1-\gamma)\rho(s_{t+1}).
\]
Note that $s_t$ satisfies a strong version of Doeblin's condition: for any states $s_0, s\in \mathcal{S}$, 
\[
\bar{\mathcal{P}}_{\theta}(s|s_0)\geq (1-\gamma)\rho(s).
\]
Let $\bar\nu_{\theta}$ denote the stationary distribution of $s_t$. Since $1-(1-\gamma)\sum_{s} \rho(s)=1-(1-\gamma)=\gamma$, by Theorem 16.2.4 in \cite{meyn2012markov}, we have
\[
||\delta_{(s)}\bar{\mathcal{P}}_{\theta}^{n}-\bar \nu_{\theta}||_{\text{TV}}\leq\gamma^{n},
\]
which is uniform across all policy parameters $\theta$.
As a result, the mixing time $t_{\text{mix}}$ of $s_t$ satifies $t_{\text{mix}}\leq\log (1/4)/\log (1/\gamma)$. 

For the hitting time, we have for any $s_0, s \in \mathcal{S}$,
\[
\E_{s_0}[\kappa_{s}]\leq \frac{1}{(1-\gamma)\rho(s)}.
\]
Thus, $t_{\text{hit}}\leq 1/((1-\gamma)\min_s \rho(s))$.

Given the bounds for $t_{\text{mix}}$ and $t_{\text{hit}}$, by Proposition~\ref{prop:policy_gradient_ergo}, we have that the $\epsilon$-mixing
time of $Z_{t}$ satisfies
\[\begin{split}
\bar \eta_\epsilon \leq&  
\frac{|\log (\epsilon/(8M+4))|}{|\log (1/4)|}\frac{\log(1/4)}{\log(1/\gamma)}
+\max\left\{\frac{|\log (\epsilon/(4M))|}{|\log (1-\alpha+\alpha\gamma)|}, 12|\log(\epsilon/(8M+4))|\right\}\\
&\times\left(1+\frac{1}{(1-\gamma)\min_{s}\rho(s)}\right)|\mathcal{A}|\sum_{k=1}^{|\mathcal{S}||\mathcal{A}|-1}\frac{1}{k}\\
\leq& \frac{|\log (\epsilon/(8M+4))|}{|\log(1/\gamma)|}
+\max\left\{\frac{|\log (\epsilon/(4M))|}{|\log (1-\alpha+\alpha\gamma)|}, 12|\log(\epsilon/(8M+4))|\right\}\\
&\times\frac{(1-\gamma)\min_{s}\rho(s)+1}{(1-\gamma)\min_{s}\rho(s)}|\mathcal{A}|(1+\log(|\mathcal{S}||\mathcal{A}|))\\
\leq& \left(\log\frac{1}{\epsilon} + \log(8M+4)\right)\\
&\times \left(\frac{1}{\log(1/\gamma)}+\left(\frac{1}{\log(1/(1-\alpha+\alpha\gamma))}+12\right)\frac{(1-\gamma)\min_{s}\rho(s)+1}{(1-\gamma)\min_{s}\rho(s)}|\mathcal{A}|(1+\log(|\mathcal{S}||\mathcal{A}|))\right)\\
=&\left(\log\frac{1}{\epsilon} + \log(8M+4)\right)\frac{1}{\log(1/r)}.
\end{split}\]
Thus, given $t$, we can achieve 
$d(Z_t, \tilde Z_t)\leq \epsilon_{t} \tilde d(z,\tilde z)$,
where $\epsilon_{t}=(8M+4)r^t$,
since $\bar{\eta}_{\epsilon_{t}}\leq t$.

\paragraph{Lipschitz Gradients and Transition Kernel.}

To verify Assumption~\ref{ass:lip0}, we first show that the transition kernel is Lipschitz according to the metric~\eqref{eq:policy_grad_metric}:
\[
\tilde d(\delta_{z}\mathcal{\widetilde{P}}_{\theta},\delta_{z}\mathcal{\widetilde{P}}_{\tilde \theta}) \leq
\left(1 + \gamma \|Q\|_{\infty} \right)R |\mathcal{A}| 
\| \theta - \tilde \theta \|.
\]
Under Assumption~\ref{ass:pg-score}, the action probabilities are Lipschitz:
\begin{align*}
&|\pi^{\theta}(a_{t+1}|s_{t+1})-\pi^{\tilde \theta}(a_{t+1}|s_{t+1})|\\ 
\leq&\left(\sup_{\theta'\in\Theta}\|\nabla_{\theta}\pi^{\theta'}(a_{t+1}|s_{t+1})\|\right)\|\theta-\tilde \theta\|\\
\leq&\left(\sup_{\theta'\in\Theta}\|\pi^{\theta'}(a_{t+1}|s_{t+1})\nabla_{\theta}\log\pi^{\theta'}(a_{t+1}|s_{t+1})\|\right)\|\theta-\tilde \theta\|\\
\leq& R\|\theta-\tilde \theta\|.
\end{align*}
%Let $\text{Lip}_{\infty}=\{f:|f(Q)-f(\tilde Q)|\leq\|Q-\tilde Q\|_{\infty}\}$
%denote the set of Lipschitz functions on $\mathbb{R}^{|\mathcal{S}||\mathcal{A}|}$.
%We can describe the Wasserstein distance between the one-step distributions $\delta_{z}\mathcal{\widetilde{P}}_{\theta}$ and $\delta_{z'}\mathcal{\widetilde{P}}_{\theta}$ as
Next note that
\begin{align*}
d(\delta_{z}\mathcal{\widetilde{P}}_{\theta},\delta_{z'}\mathcal{\widetilde{P}}_{\theta})  =&\|\delta_{z}\mathcal{P}_{\theta}-\delta_{z'}\mathcal{P}_{\theta}\|_{\text{TV}}
+ \E[\|Q_{1}-\tilde Q_1\|_{\infty}],
%+\sup_{f\in\text{Lip}_{\infty}}|\mathbb{E}f(Q_{1})-\mathbb{E}f(Q'_{1})|.
% =&\frac{1}{2}\sum_{(s_{1},a_{1})\in\mathcal{S}\times\mathcal{A}}\left|P(s_{1}|s_{0},a_{0})\pi^{\theta}(a_{1}|s_{1})-P(s_{1}|s_{0},a_{0})\pi^{\theta'}(a_{1}|s_{1})\right|\\
% & +\sup_{f\in\text{Lip}_{\infty}}|\mathbb{E}f(Q_{1})-\mathbb{E}f(Q'_{1})|
\end{align*}
where $Q_{1}$ and $\tilde Q_1$ are properly coupled.
We can bound the total variation term using the Lipschitzness of the policy, i.e.,
\begin{align*}
\|\delta_{z}\mathcal{P}_{\theta}-\delta_{z}\mathcal{P}_{\tilde \theta}\|_{\text{TV}} 
=&\frac{1}{2}\sum_{(s_{1},a_{1})\in\mathcal{S}\times\mathcal{A}}|P(s_{1}|s_{0},a_{0})\pi^{\theta}(a_{1}|s_{1})-P(s_{1}|s_{0},a_{0})\pi^{\tilde \theta}(a_{1}|s_{1})|\\
\leq&\frac{1}{2}|\mathcal{A}| R\|\theta-\tilde \theta\|.
\end{align*}
We next bound the difference for the Q-function. Note that starting
from the same Q-function $Q_{0}$, 
%the policy parameter $\theta$ only changes the update through the next state sampled to update the $Q$-function. %$(s_{1},a_{1})$.
%Note that 
$a_0'$, which is sampled uniformly at random from $\mathcal{A}$, can be coupled and are thus identical under $\pi^\theta$ and $\pi^{\tilde \theta}$. In addition, $s_1'$ which is sampled according to $P(\cdot|s_0, a_0')$ can also be coupled and are thus identical under $\pi^\theta$ and $\pi^{\tilde \theta}$. Then, the only thing that differs between $\pi^\theta$ and $\pi^{\tilde \theta}$ is the next action $a_1'$ in the Q-update, i.e., $a_1'\sim\pi^{\theta}(\cdot|s_1')$ versus $\tilde a_1'\sim\pi^{\tilde \theta}(\cdot|s_1')$. Under the coupling described above, we have
%Randomness in the action $\hat{a}$ used for the update and randomness in the cost $c(s_{t},\hat{a})$ will follow identical distributions under $\theta$ and $\theta'$ and can thus be coupled.
\begin{align*}
 \mathbb{E}[\|Q_{1}-\tilde Q_{1}\|_{\infty}]
 =& \alpha\gamma \mathbb{E}[|Q_0(s_1', a_1')-Q_0(s_1', \tilde a_1')|]\\
 \leq& \alpha\gamma\|Q_0\|_{\infty}\frac{1}{2}|\mathcal{A}|\cdot R\|\theta-\tilde \theta\|,
\end{align*} 
since $Q_0$ is bounded function of $(s,a)$.

% Since $Q_{1}(s,a)$
% is a bounded function of $(s,a)$, 
% we can treat
% $f(Q_{1})$ as a bounded function of $(s,a)$ as well
% and apply the total variation distance type of argument.
% Specifically, we have,
% \begin{align*}
%  & \sup_{f\in\text{Lip}_{\infty}}|\mathbb{E}f(Q_{1})-\mathbb{E}f(\tilde Q_{1})|\\
%  & \leq\sup_{f\in\text{Lip}_{\infty}}|\mathbb{E}f(Q_{t}+1_{s=s_{t},a=\hat{a}}\alpha(c(s_{t},\hat{a})-Q_{t}(s_{t},\hat{a})+\gamma Q_{t}(s_{t+1},a_{t+1}))\\
%  & -\mathbb{E}f(Q_{0}+1_{s=s_{0},a=a_0'}\alpha(c(s_{0},\hat{a})-Q_{t}(s_{t},\hat{a})+\gamma Q_{t}(s'_{t+1},a'_{t+1})|\\
%  & \leq\sup_{|g|\leq \gamma\| Q_{t} \|_{\infty}}|\mathbb{E}g(s_{t+1},a_{t+1})-\mathbb{E}g(s'_{t+1},a'_{t+1})|\\
%  & \leq\gamma \| Q_{t} \|_{\infty} \frac{1}{2}|\mathcal{A}|\cdot R||\theta-\theta'||
% \end{align*}

Finally, we show that $\nabla\ell(\theta)$ is also Lipschitz. 
%as a result of Wasserstein ergodicity and Lipschitzness of the transition kernel. 
Recall that $\mu^{\theta}$ denotes the stationary distribution of $Z_{t}$ under $\pi^\theta$. By Assumption \ref{ass:ergodicity0}, which we have already verified, and the Lipschitzness of $\delta_z\widetilde{P}_{\theta}$ verified above, we can apply Theorem 3.1 of \cite{rudolf2018perturbation}, which gives us
\[
d(\mu^{\theta}, \mu^{\tilde \theta})\leq \frac{R |\mathcal{A}|(8M+4)}{(1-r)(1-\gamma)} \| \theta - \tilde \theta \|
\]
% \[
% d(\mu^{\theta}, \mu^{\tilde \theta})\leq \frac{\Gamma K}{(1-r)} \| \theta - \tilde \theta \|
% \]
where $r$ is the geometric rate of Wasserstein ergodicity in \eqref{eq:pg_ergodicity_constant}. 
% and
% \[
% \Gamma = R |\mathcal{A}|, 
% \quad K = \frac{8M+4}{1-\gamma}
% \]
Since $Q_t(s_{t},a_{t})\nabla_{\theta}\log\pi^{\theta}(a_{t}|s_{t})$ is Lipschitz with respect to the metric $\tilde d(z,\tilde z)$, the above bound for $d(\mu^{\theta}, \mu^{\tilde \theta})$ implies that $\nabla \ell(\theta)$ is Lipschitz with Lipschitz constant 
\[
\frac{R |\mathcal{A}|(8M+4)}{(1-r)(1-\gamma)} = O\left(\frac{1}{(1-\gamma)^{3}}\right).
\]
Note that this matches the dependence on $\gamma$ for the Lipschitz constant of $\nabla \ell(\theta)$ in~\cite{zhang2020global}.

\paragraph{Bounded Gradients.}
For 
$g(\theta,z)=Q(s_{t},a_{t})\nabla_{\theta}\log\pi^{\theta}(a_{t}|s_{t})$, we have
\[
\|g(\theta,z)\|\leq R(1+\|Q\|_\infty).
\]
In addition, 
\[
\nabla \ell(\theta)\leq\frac{M}{1-\gamma}.
\]
We have thus verified Assumption~\ref{ass:bound0}.
\Halmos
\end{appendices}	
\end{document}